\documentclass[twoside,11pt]{article}

\usepackage{amsmath}
\usepackage{amsthm}
\let\amsthmproof\proof
\let\amsthmendproof\endproof
\usepackage[preprint]{jmlr2e}
\let\proof\amsthmproof
\let\endproof\amsthmendproof

\usepackage[utf8]{inputenc}
\usepackage[T1]{fontenc}
\usepackage{url}
\usepackage{booktabs}
\usepackage{tabularx}
\newcolumntype{C}{>{\centering\arraybackslash}X}
\usepackage{amsfonts}
\usepackage{nicefrac}
\usepackage{microtype}
\usepackage{graphicx}
\usepackage{bbding}
\usepackage{threeparttable}
\usepackage{comment}
\usepackage{bm}
\usepackage{amssymb}
\usepackage{tikz}
\usepackage{lastpage}
\usepackage{natbib}
\usepackage{placeins}

\jmlrheading{}{2026}{1-\pageref{LastPage}}{}{}{}{Ruizhi Deng, Guohao Shen, and Tangrui Xiong}
\ShortHeadings{Sharp Covering-Number Bounds for ReLU Networks}{Deng, Shen, and Xiong}
\makeatletter
\def\ps@jmlrtps{\let\@mkboth\@gobbletwo\def\@oddhead{}\def\@evenhead{}%
\def\@oddfoot{\hfill\small\rm\thepage\hfill}\def\@evenfoot{}}
\makeatother
\firstpageno{1}

\title{Sharp Covering-Number Bounds for Feedforward ReLU Networks via Parameter Symmetries}

\author{\name Ruizhi Deng \email ruizhi.deng@connect.polyu.hk \\
  \addr Department of Applied Mathematics\\
  The Hong Kong Polytechnic University\\
  Hung Hom, Kowloon, Hong Kong
   \AND
 \name Guohao Shen\thanks{\protect\hypertarget{Hfootnote.\arabic{Hfootnote}}{Corresponding author.}} \email guohao.shen@polyu.edu.hk \\
  \addr Department of Applied Mathematics\\
  The Hong Kong Polytechnic University\\
  Hung Hom, Kowloon, Hong Kong
  \AND
 \name Tangrui Xiong \email tangrui.xiong@connect.polyu.hk \\
  \addr Department of Applied Mathematics\\
  The Hong Kong Polytechnic University\\
  Hung Hom, Kowloon, Hong Kong
}

\editor{}

\newcommand{\R}{\mathbb R}

\DeclareMathOperator{\vol}{vol}

\begin{document}
\maketitle

\begin{abstract}
In neural networks, distinct parameter configurations can realize the same input--output mapping, indicating that nominal parameter counts overstate functional complexity. We characterize the parameter-space symmetries of ReLU and Leaky ReLU networks across fully connected, convolutional, and residual architectures. Specifically, we prove that every global quasianalytic parameter map preserving the network function is a composition of positive channel scaling and channel permutation. Quotienting out both symmetries yields uniform covering-number upper bounds for bounded-parameter networks: positive scaling reduces the exponent from the parameter count $S$ to $S-H$ (for $H$ scalable channels), while permutation symmetry divides the bound by the factorial count $G$ of channel permutations. Matching lower bounds show that $S-H$ is sharp for each fixed architecture. For one-hidden-layer fully connected networks and under generic identifiability, permutation symmetry also yields a factor $1/G$ in the asymptotic lower bound. We further extend these results to fully connected networks with continuous piecewise-linear activations. Finally, we apply these bounds to empirical risk minimization and capacity control beyond i.i.d.\ settings, and note implications for weight rebalancing and optimization. Numerical experiments on MLPs, CNNs, and ResNets illustrate the local effective dimension $S-H$ and symmetry alignment in model averaging.
\end{abstract}

\begin{keywords}
covering numbers, ReLU networks, parameter symmetries, functional dimension, identifiability
\end{keywords}

\section{Introduction}

With the development of deep learning, neural networks have grown increasingly large, making model scale a key factor in their empirical performance \citep{kaplan2020scaling}. Since a neural network represents an input--output function through a collection of weights and biases, scaling up the model directly translates into a large number of parameters, pushing modern networks into heavily overparameterized regimes. Yet, the ability of such overparameterized models to generalize has motivated renewed study of the relationship between model size and statistical complexity \citep{zhang2017rethinking,belkin2019reconciling}. A central factor complicating this relationship is parameter redundancy: distinct parameter configurations can realize the exact same input--output function. Consequently, the raw parameter count $S$ fails to capture the true complexity of the functions represented by a neural network. Accounting for this redundancy is therefore essential for accurately measuring functional complexity and deriving tight statistical learning guarantees.

Equivalences in the parameter space, often referred to as parameter symmetries, provide a concrete lens for characterizing such redundancy \citep{zhao2025symmetry}. In a feedforward ReLU or Leaky ReLU network, permuting the neurons/channels within a hidden layer along with their incoming and outgoing weights leaves the network function unchanged; similarly, multiplying a hidden neuron/channel's incoming parameters by a positive constant while dividing its outgoing weights by the same constant preserves the overall mapping. Prior geometric studies have provided valuable insights into how these symmetries govern parameter identifiability and the local rank of the realization map \citep{rolnick2020reverse,phuong2020functional,grigsby2023hidden,grigsby2025functional}. A natural next step for statistical learning theory is to connect these local structural properties to the \emph{global} metric capacity of a bounded-parameter function class $\mathcal F$ \citep{petersen2021topological}, as measured by the uniform covering number $\mathcal N(\mathcal F,\varepsilon,\|\cdot\|_\infty)$.

Recently, \citet{shen2024complexity} incorporated permutation symmetry into deriving covering-number upper bounds for fully connected networks, improving the bound by a factorial factor $G=\prod_{l=1}^L d_l!$ (for hidden-layer widths $d_1,\dots,d_L$), while retaining the ambient exponent $S$ in $\varepsilon^{-S}$. In a complementary direction, \citet{ou2026covering} established metric-entropy upper and lower bounds for wide, deep ReLU MLPs that match up to absolute constants, with explicit dependence on the width, the depth, and the weight bound, without exploiting symmetry reductions. These developments leave several fundamental questions open \citep{shen2024complexity,zhao2025symmetry}: first, whether continuous scaling symmetry can be jointly exploited with permutation symmetry across modern architectures to strictly reduce the global covering-number exponent below the ambient parameter count $S$; second, whether any other regular parameter symmetries exist that could further compress the function space; and third, whether matching lower bounds can be established to certify that the resulting symmetry-reduced capacity bounds are sharp.

In this work, we address these questions for ReLU and Leaky ReLU networks within a unified framework encompassing fully connected (MLP), convolutional (CNN), and residual (ResNet) architectures. First, we prove that positive scaling and permutation of hidden neurons/channels exhaust all function-preserving quasianalytic maps of the whole parameter space. Second, by jointly quotienting out both symmetries, we show that positive scaling eliminates one degree of freedom per scalable neuron/channel, reducing the covering-number exponent from the ambient parameter count $S$ to $S-H$, where $H$ is the total number of scalable hidden neurons/channels, while permutation symmetry divides the bound by the total number $G$ of layerwise neuron/channel permutations. Specifically, for any fixed $L$-block architecture $\mathcal F$ with hidden widths of at least two, we establish the upper bound $\mathcal N(\mathcal F,\varepsilon,\|\cdot\|_\infty)\le(C_+/G)\varepsilon^{-(S-H)}$ together with a general nonasymptotic lower bound of order $\varepsilon^{-(S-H)}$ for all $0<\varepsilon\le 2B$. For fully connected networks with one hidden layer and, more generally, for generically identifiable architectures, we further obtain the two-sided asymptotic bound
\[
  \frac{C_-}{G}
  \le \liminf_{\varepsilon\downarrow0}\varepsilon^{S-H}\,\mathcal N(\mathcal F,\varepsilon,\|\cdot\|_\infty)
  \le \limsup_{\varepsilon\downarrow0}\varepsilon^{S-H}\,\mathcal N(\mathcal F,\varepsilon,\|\cdot\|_\infty)
  \le \frac{C_+}{G},
\]
where $C_-,C_+>0$ depend on the architecture and are independent of $\varepsilon$ (for a standard $L$-layer MLP with hidden widths $d_1,\dots,d_L$, we have $H=\sum_{l=1}^L d_l$ and $G=\prod_{l=1}^L d_l!$). A comparison with existing covering-number bounds is summarized in Table~\ref{tab:results}.

\begin{table}[t]
\centering
\small
\begin{threeparttable}
\caption{Comparison of $\varepsilon$-covering number bounds $\mathcal N(\mathcal F,\varepsilon,\|\cdot\|)$ for neural networks.}
\label{tab:results}
\begin{tabular}{llccc}
\toprule
Reference & Architectures & Norm & Upper bound & Lower bound\\
\midrule
\citet{bartlett2017spectral}\tnote{$\natural$} & MLP (no bias) & $L_2(P_n)$ & $\exp(c\,\varepsilon^{-2})$ & --\\
\citet{lin2019generalization}\tnote{$\natural$} & MLP, CNN (no bias) & $L_2(P_n)$ & $\exp(c\,\varepsilon^{-1/2})$ & --\\
\citet{bartlett2019nearly}\tnote{$\sharp$} & MLP, CNN, ResNet & $L_p(Q)$\tnote{$*$} & $\varepsilon^{-CSL\log S}$ & --\\
\citet{long2020generalization}\tnote{$\natural$} & MLP, CNN (no bias) & $\|\cdot\|_\infty$ & $\varepsilon^{-S}$ & --\\
\citet{shen2024complexity}\tnote{$\natural$} & MLP & $\|\cdot\|_\infty$ & $\varepsilon^{-S}/G$ & --\\
\citet{ou2026covering} & MLP & $L_p$\tnote{$\ddagger$} & $\varepsilon^{-CS}$ & $\varepsilon^{-cS}$\\
\midrule
\textbf{This paper} & \textbf{MLP, CNN, ResNet} & $\boldsymbol{\|\cdot\|_\infty}$ & $\boldsymbol{\varepsilon^{-(S-H)}/G}$
  & $\boldsymbol{\varepsilon^{-(S-H)}}$ \textbf{(}$\boldsymbol{/G}$\tnote{$\dagger$}\,\textbf{)}\\
\bottomrule
\end{tabular}
\begin{tablenotes}[para,flushleft]\footnotesize
\item Orders in $\varepsilon$ as $\varepsilon\downarrow 0$ for ReLU/Leaky ReLU networks (Theorem~\ref{thm:covering-number-cpwl} gives $\varepsilon^{-S}/G$ for general CPWL activations). $S$: parameter count; $L$: depth; $H$: scalable channels; $G$: channel permutations; $c,C>0$: constants. Parameter constraints: weight-matrix norms \citep{bartlett2017spectral,lin2019generalization}, unbounded \citep{bartlett2019nearly}, bounded distance from initialization \citep{long2020generalization}, and entrywise bounds otherwise. $^{\!\natural}$General Lipschitz activations (no scaling symmetry in general). $^{\!\sharp}$Piecewise-linear activations. $^{\!*}$For $1\le p<\infty$ and outputs in $[-1,1]$, converted from the pseudodimension bound $O(SL\log S)$ \citep[Theorem~2.6.7]{vanderVaart1996weak} ($S$ without weight sharing); the matching pseudodimension lower bound does not yield a covering-number lower bound as $\varepsilon\downarrow 0$. $^{\!\ddagger}$All $p\in[1,\infty]$ (including $\|\cdot\|_\infty$); lower bound requires width and depth $\ge 60$ (fixed-architecture $\varepsilon$-rates shown; their bounds also track explicit width, depth, and weight-bound factors). $^{\!\dagger}$Factor $1/G$ holds for one-hidden-layer MLPs (Theorem~\ref{thm:shallow-lower-bound}) and generically identifiable architectures (Theorem~\ref{thm:sharp-covering-exponent}(b)).
\end{tablenotes}
\end{threeparttable}
\end{table}

Our main contributions are as follows:
\begin{enumerate}
  \item Within a unified framework encompassing fully connected, convolutional, and residual architectures with ReLU/Leaky ReLU activations, we prove that positive channel scaling and channel permutation are the \emph{only} functional equivalences induced by quasianalytic parameter maps. Hence no other quasianalytic map of the whole parameter space preserves the network function.

  \item We derive explicit uniform covering-number upper bounds of order $G^{-1}\varepsilon^{-(S-H)}$ across the unified architecture. Furthermore, for networks of arbitrary depth and hidden widths of at least two, we establish matching lower bounds showing that the reduced exponent $S-H$ is sharp for each fixed architecture, and that permutation symmetry yields a factor $1/G$ in the asymptotic lower bound for one-hidden-layer fully connected networks and generically identifiable architectures.

  \item We extend our covering-number analysis to networks with general continuous piecewise-linear (CPWL) activations, deriving permutation-reduced upper bounds of order $G^{-1}\varepsilon^{-S}$ and complementary lower bounds. We also show that restricting the CPWL activations to a single-breakpoint regime can achieve the reduced exponent $S-H$ while preserving optimal approximation rates.

  \item We apply our covering-number bounds to empirical risk minimization. We establish excess-risk upper bounds with finite-sample reductions in channel-sparse architectures, lower bounds showing that their rates in the sample size cannot be improved in general, and, for Gaussian regression, an oracle inequality for least-squares estimation together with an asymptotic lower bound showing that its dependence on $S-H$ cannot be improved. Because the bounds hold under the uniform norm, they also provide the capacity control needed beyond standard i.i.d.\ data, including adversarial training, dependent data, and covariate shift. We also clarify how scaling and permutation symmetries enable weight rebalancing in function approximation and explain loss-landscape degeneracy in optimization.
\end{enumerate}

The remainder of this paper is organized as follows. Section~\ref{subsec:related-work} reviews related work across parameter symmetries, network capacity, and statistical learning. Section~\ref{sec:settings} introduces the unified network architecture, formalizes functional equivalence, and establishes the completeness of positive scaling and permutation symmetries. Section~\ref{sec:main-results} presents our covering-number upper and lower bounds and extends the analysis to general continuous piecewise-linear (CPWL) activations. Section~\ref{sec:consequences} develops the implications of these results for statistical risk estimation, learning beyond standard i.i.d.\ settings, function approximation, and optimization. Section~\ref{sec:experiments} presents numerical illustrations of empirical local ranks and symmetry-aligned model averaging, and Section~\ref{sec:conclusion} concludes the paper. Experimental details and all technical proofs are deferred to the appendices.

\subsection{Related Work}
\label{subsec:related-work}

\noindent\textbf{Parameter symmetries, identifiability, and model reverse engineering.}
The study of functional equivalence in neural networks dates back to classical identifiability and reconstruction results for smooth activations such as $\tanh$, where parameters are determined by the input--output map up to simple discrete or affine symmetries \citep{sussmann1992uniqueness,fefferman1994reconstructing,vlacic2021affine}. For ReLU and piecewise-linear networks, identifiability becomes inherently geometric and architecture-dependent. A series of works have established conditions and constructed nonempty open sets of parameters on which a network's weights are locally or globally identifiable from its realized function up to permutation and positive scaling \citep{phuong2020functional,bonapellissier2023identifiability,stock2023embedding,grillo2026identifiable}. Closely related is the problem of \emph{model reverse engineering} (or parameter extraction), which seeks to algorithmically reconstruct an unknown ReLU network's weights and biases, up to permutation and positive scaling, by querying its input--output function and locating the bent hyperplanes where neurons switch activation states \citep{milli2019model,rolnick2020reverse,jagielski2020high,carlini2020cryptanalytic,canales2024polynomial}. From a complementary algebraic and differential perspective, \citet{godfrey2022symmetries} characterized intertwining symmetries induced by invertible linear maps between hidden layers, and \citet{grigsby2023hidden,grigsby2025functional} studied the local functional dimension (the pointwise rank of the parameter-to-function realization map), showing that it is upper bounded by $S-H$ and varies across the parameter space; \citet{grillo2026identifiable} showed that this bound is attained for fully connected ReLU networks whose input and hidden widths are at least two.

\medskip\noindent\textbf{Capacity measures and covering numbers of neural networks.}
Bounding the capacity of deep neural networks is a central theme in statistical learning theory \citetext{\citealp{anthony1999neural,zhou2002covering}}. One major line of work studies the Vapnik--Chervonenkis (VC) dimension and pseudodimension of piecewise-linear networks over unbounded parameter spaces, establishing upper bounds of order $O(SL\log S)$ and lower bounds of order $\Omega(SL\log(S/L))$ for networks of depth $L$ and parameter count $S$ \citep{harvey2017nearly,bartlett2019nearly}. A second line of work imposes spectral or matrix-norm constraints on networks without bias vectors (setting all layer biases to zero), deriving size-independent Rademacher complexity bounds \citep{neyshabur2015norm,golowich2018size}, PAC-Bayesian margin bounds \citep{neyshabur2017pac}, and covering-number bounds whose logarithm grows polynomially in $1/\varepsilon$ under the empirical $L_2(P_n)$ norm evaluated on $n$ sample points \citep{bartlett2017spectral,lin2019generalization}. A third line of work directly analyzes networks with uniformly bounded weights and biases under the global uniform norm $\|\cdot\|_\infty$, where standard discretization of the $S$-dimensional parameter cube yields covering-number upper bounds of order $\varepsilon^{-S}$ \citep{anthony1999neural,long2020generalization,kong2025posterior}. Within this bounded-parameter framework, \citet{shen2024complexity} exploited neuron permutation equivalence in fully connected networks to divide the $\varepsilon^{-S}$ upper bound by the factorial factor $G=\prod_{l=1}^L d_l!$, while \citet{ou2026covering} recently established two-sided metric-entropy bounds that match up to absolute constants for wide, deep fully connected ReLU networks, valid in $L^p$ for $p\in[1,\infty]$.

\medskip\noindent\textbf{Symmetries in optimization and model merging.}
Parameter symmetries also play a central role in optimization and model aggregation \citep{zhao2025symmetry}. Continuous scaling invariance keeps the loss constant along each scaling fiber and creates flat directions in the loss landscape \citep{fukumizu2019semiflat}. Under gradient flow, this invariance conserves the difference between the incoming and outgoing squared norms of each channel \citep{kunin2021neural,zhao2023symmetries}. Moreover, positive rescaling can change the largest Hessian eigenvalue at a smooth minimum without changing the realized function \citep{dinh2017sharp}. These properties have motivated scale-invariant optimizers \citep{neyshabur2015path,meng2019gsgd} and symmetry teleportation algorithms \citep{zhao2022teleportation}. They also explain how weight decay ($\ell_2$ regularization) promotes balance between the incoming and outgoing norms of scalable units \citep{neyshabur2015norm,parhi2021banach}. On the discrete side, permutation symmetry implies that independently trained networks often converge to permuted copies of a solution. This observation underlies the study of mode connectivity \citep{tatro2020optimizing,entezari2022role,pittorino2022deep} and has led to channel-alignment algorithms for model merging and federated averaging \citep{wang2020fedma,singh2020model,ainsworth2023git,jordan2023repair,boufalis2026}, as well as symmetry-equivariant architectures that process network weights directly \citep{navon2023equivariant,kalogeropoulos2024scale}.\looseness=1

\section{Parameter Symmetries of Feedforward Neural Networks}
\label{sec:settings}

To examine how parameter symmetries reduce the complexity bounds of neural networks, we first characterize functional equivalences in common feedforward architectures. We begin by illustrating these symmetries across standard network layers, and then formulate a unified architecture to establish both the sufficiency and completeness of scaling and permutation symmetries.

\subsection{Functional Equivalence and Layerwise Symmetries}

\begin{definition}[Functional equivalence]\label{def:functional-equivalence}
Two neural networks $f_1(\cdot;\theta_1)$ and $f_2(\cdot;\theta_2)$ are said to be \emph{functionally equivalent} on an input domain $\mathcal{X}$ if they realize the same input--output mapping across the entire domain, i.e.,
\[
 f_1(x;\theta_1)=f_2(x;\theta_2)\qquad\text{for all }x\in\mathcal{X},
\]
where $\theta_1$ and $\theta_2$ denote the respective parameter vectors of the two networks.
\end{definition}

Functional equivalence in Definition~\ref{def:functional-equivalence} can occur across distinct architectures ($f_1\neq f_2$), as studied in network pruning and model compression \citep{han2015learning,cheng2017survey,frankle2019lottery}, or within a fixed architecture ($f_1=f_2=f$) through parameter symmetries such as neuron permutations, sign flips, and weight scaling \citep{shen2024complexity}. In this paper, we focus on the latter setting and analyze the functional complexity of a fixed architecture $f$ under equivalences of the form $f(x;\theta_1)=f(x;\theta_2)$.

While permutation symmetry holds for any activation function, continuous weight-scaling symmetry requires the activation to be positively homogeneous and thus fails for general activations such as sigmoid or $\tanh$. Accordingly, we focus on networks with the coordinatewise piecewise-linear activation $\sigma(t)=\max\{t,\alpha t\}$ for a fixed slope $0\le\alpha<1$ (recovering the ReLU for $\alpha=0$ and the Leaky ReLU for $0<\alpha<1$), which satisfies $\sigma(ct)=c\sigma(t)$ for all $c>0$. Before introducing our unified framework, we illustrate how scaling and permutation equivalences arise in three basic building blocks: fully connected layers, convolutional layers, and residual blocks.

\paragraph{Fully connected layers (permutation and scaling symmetries).}
A fully connected network $f$ of depth $L$ is a composition of affine maps and activations. Given an input $x\in\mathbb{R}^{d_0}$, the network computes
\begin{equation}\label{eq:deep-fnn}
 h^{(0)}=x,\qquad h^{(l)}=\sigma\bigl(W^{(l)}h^{(l-1)}+b^{(l)}\bigr)
 \quad(1\le l\le L),\qquad f(x;\theta)=a^\top h^{(L)}+\beta,
\end{equation}
where $d_1,\ldots,d_L$ are the hidden-layer widths, $W^{(l)}\in\mathbb{R}^{d_l\times d_{l-1}}$ and $b^{(l)}\in\mathbb{R}^{d_l}$ denote the weight matrix and bias vector of layer $l$, and $a\in\mathbb{R}^{d_L}$ and $\beta\in\mathbb{R}$ produce the scalar output. Let $\theta$ collect all weights and biases. As shown in the following example, the hidden neurons of any layer admit both positive scaling and permutation symmetries.

\begin{example}[Scaling and permutation in fully connected layers]
\label{ex:two-neuron-symmetry}
Consider two shallow networks ($L=1$) of the form $f(x;\theta_k)=W_k^{(2)}\sigma\bigl(W_k^{(1)}x+b_k^{(1)}\bigr)+b_k^{(2)}$ ($k=1,2$) with $d_1$ hidden neurons:
\begin{itemize}
 \item \emph{Positive scaling:} Let $D$ be a $d_1\times d_1$ diagonal matrix with positive diagonal entries. If we rescale the incoming and outgoing parameters via
 \[
\bigl(W_2^{(1)},b_2^{(1)}\bigr)=\bigl(DW_1^{(1)},Db_1^{(1)}\bigr),\qquad
  W_2^{(2)}=W_1^{(2)}D^{-1},\qquad b_2^{(2)}=b_1^{(2)},
 \]
 then $f(\cdot;\theta_1)$ and $f(\cdot;\theta_2)$ are functionally equivalent because positive homogeneity gives $\sigma(Dz)=D\sigma(z)$, so $W_2^{(2)}\sigma(DW_1^{(1)}x+Db_1^{(1)})=W_1^{(2)}D^{-1}D\sigma(W_1^{(1)}x+b_1^{(1)})$. 
 
 \item \emph{Neuron permutation:} Let $P$ be a $d_1\times d_1$ permutation matrix. If
 \[
\bigl(W_2^{(1)},b_2^{(1)}\bigr)=\bigl(PW_1^{(1)},Pb_1^{(1)}\bigr),\qquad
  W_2^{(2)}=W_1^{(2)}P^\top,\qquad b_2^{(2)}=b_1^{(2)},
 \]
 then $f(\cdot;\theta_1)$ and $f(\cdot;\theta_2)$ are functionally equivalent since $\sigma(Pz)=P\sigma(z)$ and $P^\top P=I$. 
\end{itemize}
\end{example}

Example~\ref{ex:two-neuron-symmetry} gives sufficient conditions for parameter equivalence in fully connected layers. Viewing $D$ and $P$ as symmetry operators on the parameter space directly quantifies the resulting redundancy: $D$ introduces $d_1$ independent positive scaling factors (one per hidden neuron), while $P$ relabels the $d_1$ hidden neurons and induces a discrete redundancy of size $d_1!$. Although Example~\ref{ex:two-neuron-symmetry} is stated for $L=1$ shallow networks, both equivalences apply immediately to any pair of adjacent layers in a deep network~\eqref{eq:deep-fnn}.

\paragraph{Convolutional layers (channelwise symmetries).}

A convolutional layer maps an input feature map $h:\Omega\to\mathbb{R}^c$ on a spatial grid $\Omega$ to an output feature map on a grid $\Omega'$, where $c$ and $c'$ are the input and output channel counts. Over a spatial window $\Lambda$ (e.g., a $k_h\times k_w$ window in 2D), a convolution kernel $K$ consists of $c'$ filters of shape $|\Lambda|\times c$, represented at each offset $u\in\Lambda$ by a weight matrix $K(u)\in\mathbb{R}^{c'\times c}$ whose entry $K_{jk}(u)$ connects input channel $k$ at location $p+u$ to output channel $j$ at location $p$. With zero padding outside $\Omega$, the convolution computes
\[
 (K\star h)_j(p)=\sum_{u\in\Lambda:\ p+u\in\Omega}\ \sum_{k=1}^{c}K_{jk}(u)\,h_k(p+u),
 \qquad p\in\Omega',\ 1\le j\le c',
\]
where the input locations around $p$ form the local \emph{patch} at $p$. Adding a channelwise bias $b\in\mathbb{R}^{c'}$ across all spatial locations and applying $\sigma$ coordinatewise yields the layer $h\mapsto\sigma(K\star h+b)$. {As is standard, each convolution uses either \emph{same} padding, with $\Omega'=\Omega$ and $0\in\Lambda$, or \emph{valid} padding, with $\Omega'=\{p:p+\Lambda\subset\Omega\}$, and its window fits into the input grid, that is, $p+\Lambda\subset\Omega$ for some $p$.} All convolutions use unit stride, and the input grid and each window are rectangular boxes (Cartesian products of finite integer intervals). For matrices $A\in\mathbb{R}^{c'\times c'}$ and $B\in\mathbb{R}^{c\times c}$, we write $AKB$ for the transformed kernel with $(AKB)(u)=AK(u)B$ for all $u\in\Lambda$.

\begin{example}[Scaling and permutation in convolutional layers]
\label{ex:conv-symmetry}
Consider two two-layer convolutional blocks mapping $h:\Omega_0\to\mathbb{R}^{c_0}$ to $g(h;\theta_k)=K_k^{(2)}\star\sigma\bigl(K_k^{(1)}\star h+b_k^{(1)}\bigr)+b_k^{(2)}$ ($k=1,2$) with $c_1$ hidden channels:
\begin{itemize}
 \item \emph{Channelwise positive scaling:} Let $D$ be a $c_1\times c_1$ diagonal matrix with positive diagonal entries. If
 \[
  \bigl(K_2^{(1)},b_2^{(1)}\bigr)=\bigl(DK_1^{(1)},Db_1^{(1)}\bigr),\qquad
  K_2^{(2)}=K_1^{(2)}D^{-1},\qquad b_2^{(2)}=b_1^{(2)},
 \]
 then $g(\cdot;\theta_1)$ and $g(\cdot;\theta_2)$ are functionally equivalent, since convolution is bilinear and $\sigma$ acts coordinatewise with $\sigma(Dz)=D\sigma(z)$ at every spatial location.

 \item \emph{Channel permutation:} Let $P$ be a $c_1\times c_1$ permutation matrix. If
 \[
  \bigl(K_2^{(1)},b_2^{(1)}\bigr)=\bigl(PK_1^{(1)},Pb_1^{(1)}\bigr),\qquad
  K_2^{(2)}=K_1^{(2)}P^\top,\qquad b_2^{(2)}=b_1^{(2)},
 \]
 then $g(\cdot;\theta_1)$ and $g(\cdot;\theta_2)$ are functionally equivalent, since $\sigma(Pz)=P\sigma(z)$ and $P^\top P=I$ at every spatial location.
\end{itemize}
\end{example}

Example~\ref{ex:conv-symmetry} gives sufficient conditions for parameter equivalence in convolutional layers. Because the $c_1$ filters in $K_k^{(1)}$ and the bias $b_k^{(1)}$ are shared across all spatial locations in $\Omega_1$, individual spatial activations within a feature map cannot be rescaled or permuted independently; instead, $D$ and $P$ act \emph{channelwise} across the entire feature map. Consequently, $D$ introduces $c_1$ independent positive scaling factors (determined by the channel count $c_1$ rather than the total number of spatial activations $|\Omega_1|c_1$), while $P$ relabels the $c_1$ hidden channels as whole units and induces a discrete redundancy of size $c_1!$. Both symmetries extend directly to any pair of adjacent layers in a deep convolutional network.

\paragraph{Residual blocks (coupled symmetries across shortcuts).}

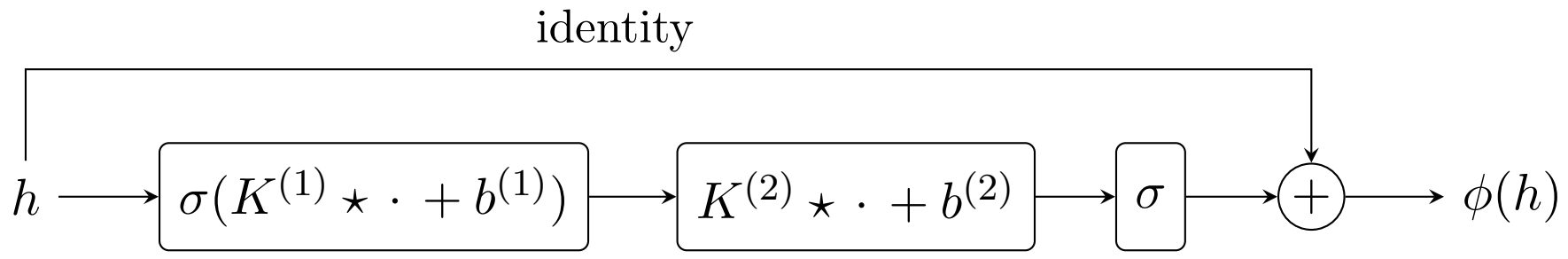
\begin{figure}[t]
\centering
\begin{tikzpicture}[>=stealth,x=1cm,y=1cm,
 box/.style={draw,rounded corners=2pt,minimum height=0.8cm,inner sep=4pt}]
 \node (in) at (0,0) {$h$};
 \node[box] (c1) at (2.6,0) {$\sigma(K^{(1)}\star\,\cdot\,+b^{(1)})$};
 \node[box] (c2) at (6.2,0) {$K^{(2)}\star\,\cdot\,+b^{(2)}$};
 \node[box] (act) at (8.4,0) {$\sigma$};
 \node[draw,circle,inner sep=1pt] (plus) at (9.6,0) {$+$};
 \node (out) at (11.1,0) {$\phi(h)$};
 \draw[->] (in) -- (c1);
 \draw[->] (c1) -- (c2);
 \draw[->] (c2) -- (act);
 \draw[->] (act) -- (plus);
 \draw[->] (plus) -- (out);
 \draw[->] (in.north) |- (4.4,0.95) -| (plus.north);
 \node[above] at (4.4,0.95) {\small identity};
\end{tikzpicture}
\caption{A residual block; the identity shortcut adds the input $h$ {after} the second activation.}
\label{fig:block}
\end{figure}

A residual block \citep{he2016} applies two convolutional layers to an input feature map $h:\Omega\to\mathbb{R}^{c_0}$ on a spatial grid $\Omega$ and adds the input back via an identity shortcut after the second activation, as in the ``ReLU before addition'' unit of \citet[Fig.~4(c)]{he2016identity} (Figure~\ref{fig:block}):
\[
 \phi(h;\theta)=h+\sigma\Bigl(K^{(2)}\star\sigma\bigl(K^{(1)}\star h+b^{(1)}\bigr)+b^{(2)}\Bigr).
\]
Here, the first (\emph{inner}) convolution maps $h$ to an intermediate feature map with $c_1$ inner channels, and the second maps back to the original grid $\Omega$ and channel count $c_0$.

\begin{example}[Scaling and permutation in residual blocks]
\label{ex:res-symmetry}
Consider two residual blocks $\phi(h;\theta_k)={h+\sigma\bigl(K_k^{(2)}\star\sigma(K_k^{(1)}\star h+b_k^{(1)})+b_k^{(2)}\bigr)}$ ($k=1,2$) mapping $h:\Omega\to\mathbb{R}^{c_0}$ through $c_1$ inner channels:
\begin{itemize}
 \item \emph{Inner-channel positive scaling:} Let $D$ be a $c_1\times c_1$ diagonal matrix with positive diagonal entries. If
 \[
  \bigl(K_2^{(1)},b_2^{(1)}\bigr)=\bigl(DK_1^{(1)},Db_1^{(1)}\bigr),\qquad
  K_2^{(2)}=K_1^{(2)}D^{-1},\qquad b_2^{(2)}=b_1^{(2)},
 \]
 then $\phi(\cdot;\theta_1)$ and $\phi(\cdot;\theta_2)$ are functionally equivalent, since $D$ commutes with the first activation $\sigma$ and cancels with $D^{-1}$ in $K_2^{(2)}$ before the shortcut addition.

 \item \emph{Inner-channel permutation:} Let $P$ be a $c_1\times c_1$ permutation matrix. If
 \[
  \bigl(K_2^{(1)},b_2^{(1)}\bigr)=\bigl(PK_1^{(1)},Pb_1^{(1)}\bigr),\qquad
  K_2^{(2)}=K_1^{(2)}P^\top,\qquad b_2^{(2)}=b_1^{(2)},
 \]
 then $\phi(\cdot;\theta_1)$ and $\phi(\cdot;\theta_2)$ are functionally equivalent, since $\sigma(Pz)=P\sigma(z)$ and $P^\top P=I$ before the shortcut addition.
\end{itemize}
\end{example}

Example~\ref{ex:res-symmetry} gives sufficient conditions for parameter equivalence in a residual block. Because the $c_1$ inner channels are internal to the residual branch, they behave like standard convolutional channels: $D$ introduces $c_1$ independent positive scaling factors, while $P$ induces a discrete redundancy of size $c_1!$. In contrast, when residual blocks are stacked in a deep network, the identity shortcut adds the input $h$ directly to the {activated output $\sigma\bigl(K^{(2)}\star(\cdots)+b^{(2)}\bigr)$ of the second convolution}. Consequently, the $c_0$ output channels cannot be rescaled or permuted independently of the input channels; any channelwise scaling or permutation applied to $h$ must be shared by the block output (with $K^{(1)}$, $K^{(2)}$, and $b^{(2)}$ adjusted accordingly). Thus, a residual block contributes independent scaling and permutation degrees of freedom only through its $c_1$ inner channels, while coupling the symmetries of its input and output channels.

\subsection{Unified Network Architecture and Characterization of Symmetries}

Building on the structural parallels above, we now formulate a unified network architecture encompassing fully connected, convolutional, and residual networks, and characterize both the sufficiency and completeness of scaling and permutation equivalences.

\begin{definition}[Unified block networks]\label{def:block-network}
Given an input domain $\mathcal{X}$, an $L$-block network maps an input $x\in\mathcal{X}$ to a scalar output via
\begin{equation}\label{eq:block-network}
 h_0=x,\qquad h_l=\phi_l(h_{l-1})\quad(1\le l\le L),\qquad
 f(x;\theta)=\frac{1}{|\Omega_L|}\sum_{p\in\Omega_L}a^\top h_L(p)+\beta,
\end{equation}
where $h_l:\Omega_l\to\mathbb{R}^{c_l}$ is a feature map with $c_l$ channels on a finite spatial grid $\Omega_l$ (with $h_l\in\mathbb{R}^{c_l}$ when $|\Omega_l|=1$), and each block $\phi_l$ takes one of three forms:
\begin{equation}\label{eq:blocks}
 \phi_l(h)=
 \begin{cases}
  \sigma_l\bigl(W_lh+b_l\bigr)
   & \text{(fully connected layer, $|\Omega_l|=1$)},\\[2pt]
  \sigma_l\bigl(K_l\star h+b_l\bigr)
   & \text{(convolutional layer)},\\[2pt]
    {h+\sigma_l\Bigl(K_l^{(2)}\star\sigma_l\bigl(K_l^{(1)}\star h+b_l^{(1)}\bigr)+b_l^{(2)}\Bigr)}
   & \text{(residual block)}.
 \end{cases}
\end{equation}
Here, $\sigma_l(t)=\max\{t,\alpha_l t\}$ is a coordinatewise activation with slope $0\le\alpha_l<1$, $W_lh$ is a linear map applied to the flattened input $h$, and $K_l$ (as well as $K_l^{(1)},K_l^{(2)}$) is a convolution kernel with spatial window $\Lambda_l$ (respectively, $\Lambda_l^{(1)},\Lambda_l^{(2)}$). In a residual block $l$, the input and output share the same grid $\Omega_l=\Omega_{l-1}$ and channel count $c_l=c_{l-1}$, while the first convolution is called the \emph{inner} layer and has $m_l$ output channels.
\end{definition}

Throughout the paper, we assume $c_0\ge1$ and that every hidden layer (including inner layers) has at least two channels. Let $\theta=(\theta_1,\ldots,\theta_L,a,\beta)\in\mathbb{R}^S$ collect all $S$ network parameters, where $\theta_l$ consists of $(W_l,b_l)$, $(K_l,b_l)$, or $(K_l^{(1)},b_l^{(1)},K_l^{(2)},b_l^{(2)})$ according to~\eqref{eq:blocks}, and $(a,\beta)\in\mathbb{R}^{c_L}\times\mathbb{R}$. For a parameter bound $B>0$, we define the bounded-parameter network class $\mathcal{F}=\{f(\cdot;\theta):\theta\in[-B,B]^S\}$ on $\mathcal{X}$. Varying the block types in~\eqref{eq:blocks} recovers multilayer perceptrons (MLPs), convolutional neural networks (CNNs), and residual networks (ResNets) as special cases. Specifically, when $|\Omega_l|=1$ for all $l$ and the network has no residual blocks, \eqref{eq:block-network} reduces to the fully connected network~\eqref{eq:deep-fnn} with widths $d_l=c_l$, activations $\sigma_l$, parameter vector $\theta=(W_1,b_1,\ldots,W_L,b_L,a,\beta)\in\mathbb{R}^S$, and parameter count $S=\sum_{l=1}^L(d_l d_{l-1}+d_l)+d_L+1$. Combining convolutional and fully connected layers yields standard CNNs, while following an initial convolutional layer with residual blocks over a common grid recovers the ResNet architecture of \citet{he2016} with the ``ReLU before addition'' unit of \citet[Fig.~4(c)]{he2016identity}, global average pooling, and a linear readout.

Next, we formalize how layerwise scaling and permutation jointly define functional equivalences across $\mathcal{F}$, and then show that under a natural regularity condition, they are the \emph{only} transformations that preserve the network function (completeness). To state the general equivalence, we call a hidden channel \emph{scalable} if it is not the output channel of a residual block, and a layer \emph{scalable} if its output channels are scalable.

\begin{proposition}[General equivalences by scaling and permutation]\label{prop:scaling-permutation}
For each block $l=0,\ldots,L$, let $P_l\in\mathbb{R}^{c_l\times c_l}$ be a permutation matrix and $D_l\in\mathbb{R}^{c_l\times c_l}$ be a positive diagonal scaling matrix, and for each residual block $l$, let $P_l^{(1)},D_l^{(1)}\in\mathbb{R}^{m_l\times m_l}$ be a permutation matrix and a positive diagonal matrix for its inner channels, subject to
\[
 P_0=D_0=I,\qquad (P_l,D_l)=(P_{l-1},D_{l-1})\ \text{for every residual block }l.
\]
Denote by $PD\cdot\theta$ the transformed parameter vector obtained by updating the parameters of each block $l=1,\ldots,L$ as
\[
 W_l'=(P_lD_l)W_l(P_{l-1}D_{l-1})^{-1}\quad\text{or}\quad K_l'=(P_lD_l)K_l(P_{l-1}D_{l-1})^{-1},\qquad b_l'=P_lD_lb_l
\]
for a fully connected or convolutional block, and
\[
 K_l'^{(1)}=(P_l^{(1)}D_l^{(1)})K_l^{(1)}(P_{l-1}D_{l-1})^{-1},\qquad b_l'^{(1)}=P_l^{(1)}D_l^{(1)}b_l^{(1)},
\]
\[
 K_l'^{(2)}=(P_lD_l)K_l^{(2)}(P_l^{(1)}D_l^{(1)})^{-1},\qquad b_l'^{(2)}=P_lD_lb_l^{(2)}
\]
for a residual block, together with the readout update $a'^\top=a^\top(P_LD_L)^{-1}$ and $\beta'=\beta$. Then $f(\cdot;\theta)$ and $f(\cdot;PD\cdot\theta)$ are functionally equivalent on $\mathcal{X}$, as each block output satisfies $h_l'=P_lD_lh_l$ for all $l=1,\ldots,L$.
\end{proposition}

Proposition~\ref{prop:scaling-permutation} unifies the layerwise examples from Section~\ref{sec:settings} into an explicit group action $PD\cdot\theta$ on $\mathbb{R}^S$. First, any transformation $P_{l-1}D_{l-1}$ from block $l-1$ is absorbed by $(P_{l-1}D_{l-1})^{-1}$ at block $l$, whereas an identity shortcut adds $h_{l-1}'=P_{l-1}D_{l-1}h_{l-1}$ directly to the residual branch and enforces the coupling $(P_l,D_l)=(P_{l-1},D_{l-1})$. This characterization underlies both function-preserving parameter manipulation \citep{dinh2017sharp,zhao2022teleportation,navon2023equivariant} and cross-shortcut channel alignment in model merging and federated aggregation \citep{wang2020fedma,singh2020model,ainsworth2023git}. Second, independent positive diagonal matrices arise at the scalable layers (the output channels of fully connected or convolutional blocks and the inner channels of residual blocks). Writing $\mathcal{L}_{\mathrm{res}}\subset\{1,\ldots,L\}$ for the index set of residual blocks, the total number of independent scaling degrees of freedom equals the number of scalable hidden channels:
\[
 H=\sum_{l\notin\mathcal{L}_{\mathrm{res}}}c_l+\sum_{l\in\mathcal{L}_{\mathrm{res}}}m_l.
\]
Accounting for these $H$ continuous scaling factors helps tighten norm-based complexity measures \citep{neyshabur2015path,dinh2017sharp,gonon2024path} and prevent scale mismatch during model interpolation \citep{jordan2023repair}. Third, each scalable layer with $c$ channels admits $c!$ relabelings, yielding
\[
 G=\prod_{l\notin\mathcal{L}_{\mathrm{res}}}c_l!\prod_{l\in\mathcal{L}_{\mathrm{res}}}m_l!
\]
channel permutations across the network (e.g., $G=\prod_{l=1}^L d_l!$ for an MLP) and replicating the loss landscape into $G$ equivalent basins \citep{entezari2022role,shen2024complexity}.

While Proposition~\ref{prop:scaling-permutation} shows that scaling and permutation are \emph{sufficient} to preserve the network function, we next ask whether other regular parameter transformations can induce additional equivalences \citep{phuong2020functional,zhao2025symmetry}.

\begin{proposition}[Completeness of scaling and permutation]
\label{prop:completeness}
Suppose that $\mathcal{X}$ contains an open ball, and let $T:\mathbb{R}^S\to\mathbb{R}^S$ be a quasianalytic map in the sense of \citet[Section~3]{bierstone2004resolution} satisfying $f(x;T(\theta))=f(x;\theta)$ for every $\theta\in\mathbb{R}^S$ and $x\in\mathcal{X}$. Then there exist fixed permutation matrices $P$ and quasianalytic positive diagonal matrices $D(\theta)$ as in Proposition~\ref{prop:scaling-permutation} such that $T(\theta)=PD(\theta)\cdot\theta$ for every $\theta\in\mathbb{R}^S$.
\end{proposition}

Proposition~\ref{prop:completeness} characterizes the completeness of scaling and permutation among global quasianalytic parameter transformations. Here, quasianalyticity is a smoothness condition generalizing real-analyticity: a smooth function is quasianalytic if it is uniquely determined by its Taylor series at any point, and thus cannot vanish on an open set without vanishing identically. Proposition~\ref{prop:completeness} shows that any quasianalytic map $T:\mathbb{R}^S\to\mathbb{R}^S$ preserving the network function for all $\theta\in\mathbb{R}^S$ must take the form of a fixed channel permutation $P$ composed with positive rescalings $D(\theta)$. This complements \citet{godfrey2022symmetries}, which characterized intertwining symmetries induced by invertible linear maps between adjacent layers, by analyzing general quasianalytic maps on the full parameter space $\mathbb{R}^S$ of the unified architecture. Proposition~\ref{prop:completeness} also provides a global counterpart to existing results on parameter identifiability \citep{phuong2020functional,bonapellissier2023identifiability,stock2023embedding,grillo2026identifiable}. At degenerate parameters $\theta$ (e.g., when outgoing weights vanish or a neuron remains inactive across $\mathcal{X}$), modifying the associated weights may leave the realized function unchanged; identifiability studies therefore investigate conditions on $\theta$ and $\mathcal{X}$ under which the equivalence class of a given network is generated solely by scaling and permutation. From a global perspective, Proposition~\ref{prop:completeness} examines transformations defined across the entire parameter space $\mathbb{R}^S$, showing that such parameter-specific redundancies at degenerate points cannot be extended to a global quasianalytic symmetry map. In the next section, we quantify how scaling and permutation equivalences reduce the covering numbers of $\mathcal{F}$.

\section{Sharp Covering-Number Bounds}
\label{sec:main-results}

Section~\ref{sec:settings} showed that scaling and permutation exhaust all functional equivalences in $\mathcal{F}$ induced by quasianalytic parameter maps. To quantify how these symmetries reduce the capacity of $\mathcal{F}$, we measure functional complexity via covering numbers under the uniform norm $\|g\|_\infty=\sup_{x\in\mathcal{X}}|g(x)|$.

\begin{definition}[Covering number]
\label{def:covering}
For a set $\mathcal{G}$ of bounded functions on a common domain $\mathcal{X}$ and a radius $\varepsilon>0$, the covering number $\mathcal{N}(\mathcal{G},\varepsilon,\|\cdot\|_\infty)$ is the smallest number of closed $\|\cdot\|_\infty$-balls of radius $\varepsilon$ with centers in $\mathcal{G}$ that cover $\mathcal{G}$.
\end{definition}

The uniform covering number $\mathcal{N}(\mathcal{G},\varepsilon,\|\cdot\|_\infty)$ is a distribution-free geometric measure that is independent of the observed samples. Because the uniform norm dominates the $L_p(Q)$ norm for any probability or empirical measure $Q$ on $\mathcal{X}$ and any $p\ge1$, bounds on $\mathcal{N}(\mathcal{G},\varepsilon,\|\cdot\|_\infty)$ directly control $L_p(Q)$ covering and bracketing numbers, providing statistical guarantees for empirical risk minimization and Bayesian posterior concentration in both standard i.i.d.\ and non-i.i.d.\ regimes \citep{oono2019approximation,schmidthieber2020nonparametric,jiao2023deep,shen2024complexity,kong2025posterior,ou2026covering}.

\subsection{Symmetry-Reduced Covering-Number Upper Bounds}
\label{subsec:upper-bounds}

We first establish a uniform covering-number upper bound for the bounded-parameter network class $\mathcal F=\{f(\cdot;\theta):\theta\in[-B,B]^S\}$ across the unified architecture of Definition~\ref{def:block-network}.

\begin{theorem}[Covering-number upper bound]
\label{thm:covering-upper-bound}
Let $\mathcal X\subset\mathbb R^{\Omega_0\times c_0}$ be any bounded domain. For any activation slopes $\alpha_1,\ldots,\alpha_L\in[0,1)$ and $B\ge1$, the uniform covering number of $\mathcal F$ satisfies, for all $0<\varepsilon\le2B$,
\begin{equation}\label{eq:detailed-upper-bound}
 \mathcal N(\mathcal F,\varepsilon,\|\cdot\|_\infty)
 \le\frac{\Bigl(12(L+L_{\mathrm{res}}+1)R_{\mathcal X}(2B)^{L+L_{\mathrm{res}}+2}c_L\bigl(\prod_{l=1}^L\kappa_l\bigr)\cdot\varepsilon^{-1}\Bigr)^{S-H}}
 {\prod_{l\notin\mathcal L_{\mathrm{res}}}c_l!\prod_{l\in\mathcal L_{\mathrm{res}}}m_l!}
 =:\frac{C_+}{G}\,\varepsilon^{-(S-H)},
\end{equation}
where $\kappa_l$ is the width factor of block $l\in\{1,\ldots,L\}$, given by
\[
 \kappa_l=
 \begin{cases}
  |\Omega_{l-1}|c_{l-1}\ (\text{or } d_{l-1}), & \text{if block $l$ is a fully connected layer},\\[4pt]
  |\Lambda_l|c_{l-1}, & \text{if block $l$ is a convolutional layer},\\[4pt]
  |\Lambda_l^{(2)}|m_l\cdot|\Lambda_l^{(1)}|c_{l-1}, & \text{if block $l$ is a residual block},
 \end{cases}
\]
with $|\Lambda_l|$ (and $|\Lambda_l^{(1)}|,|\Lambda_l^{(2)}|$) denoting the kernel window sizes and $m_l$ the inner channel count of residual block $l$. Here, $R_{\mathcal X}=\max\{1,\sup_{x\in\mathcal X}\|x\|_\infty\}$, $\mathcal L_{\mathrm{res}}\subset\{1,\ldots,L\}$ denotes the index set of residual blocks with $L_{\mathrm{res}}=|\mathcal L_{\mathrm{res}}|$, and $H=\sum_{l\notin\mathcal L_{\mathrm{res}}}c_l+\sum_{l\in\mathcal L_{\mathrm{res}}}m_l$ and $G=\prod_{l\notin\mathcal L_{\mathrm{res}}}c_l!\prod_{l\in\mathcal L_{\mathrm{res}}}m_l!$ are the total number of scalable channels and the number of channel permutations across the scalable layers, respectively.
\end{theorem}

Theorem~\ref{thm:covering-upper-bound} provides a unified covering-number upper bound under $\|\cdot\|_\infty$ for MLPs, CNNs, and ResNets. Whereas standard parameter-space covering arguments discretize all $S$ coordinates of $[-B,B]^S$ to yield bounds of order $\varepsilon^{-S}$, Theorem~\ref{thm:covering-upper-bound} exploits the scaling and permutation symmetries of Proposition~\ref{prop:scaling-permutation} to achieve a twofold reduction: positive scalings eliminate $H$ radial parameter directions, reducing the exponent from $S$ to $S-H$, while channel permutations partition the parameter space into equivalent regions, dividing the bound by $G$. The bound also reflects how each architectural module contributes to network capacity: convolutional layers reduce complexity at the channel level ($c_l$ and $c_l!$), while residual blocks---whose output channels are coupled across identity shortcuts---contribute $m_l$ scaling degrees of freedom and a divisor $m_l!$ through their inner channels.

Moreover, Theorem~\ref{thm:covering-upper-bound} applies directly to networks with layer-specific activation slopes $\alpha_l\in[0,1)$ and explicit bias vectors, which shift activation thresholds and support optimal approximation rates \citetext{\citealp{yarotsky2017,zhou2020universality,lu2021deep,shen2022approximation}}. This complements matrix-norm capacity bounds for bias-free networks \citep{neyshabur2015norm,bartlett2017spectral,neyshabur2017pac,golowich2018size,lin2019generalization} and path-norm bounds for networks with biases and skip connections \citep{gonon2024path}.

Previously, \citet{shen2024complexity} used permutation symmetry to bound the covering numbers of fully connected networks with biases. Within our framework, a standard $L$-layer MLP~\eqref{eq:deep-fnn} with widths $(d_0,d_1,\ldots,d_L)$ corresponds to setting $L_{\mathrm{res}}=0$, $|\Omega_l|=|\Lambda_l|=1$, and $c_l=d_l$ (so that $\kappa_l=d_{l-1}$, $H=\sum_{l=1}^L d_l$, and $G=\prod_{l=1}^L d_l!$), for which \eqref{eq:detailed-upper-bound} specializes to
\begin{equation}\label{eq:mlp-explicit-bound}
 \mathcal N(\mathcal F,\varepsilon,\|\cdot\|_\infty)
 \le\frac{\Bigl(12(L+1)R_{\mathcal X}(2B)^{L+2}\bigl(\prod_{l=0}^L d_l\bigr)\cdot\varepsilon^{-1}\Bigr)^{S-H}}
 {d_1!\times d_2!\times\cdots\times d_L!}.
\end{equation}
Compared with the $\varepsilon^{-S}/G$ bound in Theorem~3.5 of \citet{shen2024complexity}, quotienting out the $H$ positive scaling directions reduces the exponent in \eqref{eq:mlp-explicit-bound} from $S$ to $S-H$. For example, in a shallow network ($L=1$) of width $d_1=128$, permutation symmetry divides the bound by $128!\approx 3.86\times 10^{215}$ \citep{shen2024complexity}, while at $\varepsilon=10^{-2}$, the exponent reduction from $S$ to $S-d_1$ introduces an additional factor of $\varepsilon^{128}=10^{-256}$. More generally, the relative reduction of the exponent is $H/S\le1/\bar k$, where $\bar k$ denotes the average number of incoming weights and biases per scalable channel. It is therefore largest for shallow or narrow networks and for convolutions with small kernels and few input channels. In the shallow network above, $H/S=d_1/((d_0+2)d_1+1)\approx1/(d_0+2)$, which is close to one third for univariate inputs. For a deep MLP of constant width $W$, in contrast, $H/S\approx1/(W+1)$, and $\log G\approx LW\log W$ is small compared with the leading term $(S-H)\log(1/\varepsilon)\approx LW^2\log(1/\varepsilon)$ of the metric entropy, so both reductions become lower-order corrections in wide networks.

This combined reduction relies on the fact that continuous scaling and discrete permutation act compatibly on the parameter space: positive scalings project the $S$-dimensional parameter cube onto a finite union of $(S-H)$-dimensional coordinate faces, while channel permutations preserve this face structure and partition the resulting covering grid into orbits; retaining one center per orbit and counting the orbits via Burnside's lemma yields the divisor $G$. Because the two operations decouple, we can isolate and compare their individual and joint contributions in the following corollary.

\begin{corollary}[Upper bounds from permutation, scaling, and both]
\label{thm:covering-number-sp-deep}
Under the setting of Theorem~\ref{thm:covering-upper-bound}, for all $0<\varepsilon\le2B$,
\[
 \mathcal N(\mathcal F,\varepsilon,\|\cdot\|_\infty)\le
 \begin{cases}
  \dfrac{C_1}{G}\,\varepsilon^{-S}, & \text{(permutation only)},\\[10pt]
  C_2\,\varepsilon^{-(S-H)}, & \text{(scaling only)},\\[10pt]
  \dfrac{C_+}{G}\,\varepsilon^{-(S-H)}, & \text{(scaling and permutation)},
 \end{cases}
\]
where $H$ and $G$ are as in Theorem~\ref{thm:covering-upper-bound}, and $C_1,C_2,C_+>0$ are constants (defined in~\eqref{eq:main-prefactors} of Appendix~\ref{app:main}) satisfying $C_+=e\,C_2$ and $C_1^{1/S}=C_+^{1/(S-H)}=e^{1/(S-H)}C_2^{1/(S-H)}$.
\end{corollary}

In Corollary~\ref{thm:covering-number-sp-deep}, the first branch recovers the permutation-reduced bound of \citet{shen2024complexity} for MLPs and extends it to convolutional and residual networks, while the second branch isolates the exponent reduction from $S$ to $S-H$ yielded by positive scaling. Combining both symmetries in the third branch reflects the group action $PD\cdot\theta$: it divides the bound by $G$ and reduces the exponent by $H$, with $C_+=e\,C_2$ for the particular constants chosen here.

\subsection{Matching Lower Bounds and Generic Identifiability}
\label{subsec:lower-bounds}

Given the upper bound in Theorem~\ref{thm:covering-upper-bound}, we next ask whether its exponent is sharp and how channel permutations enter the lower bound. Theorem~\ref{thm:shallow-lower-bound} gives an asymptotic lower bound with the factor $1/G$ for fully connected networks with one hidden layer. For the unified architecture, Theorem~\ref{thm:sharp-covering-exponent} establishes the exponent $S-H$ in general, with the same factor in the asymptotic lower bound under the generic identifiability condition of Definition~\ref{def:generic-identifiability}.

\begin{theorem}[Permutation factor for one hidden layer]\label{thm:shallow-lower-bound}
Let $L=1$, so that $f(x;\theta)=\sum_{j=1}^{d_1}a_j\sigma(w_j^\top x+b_j)+\beta$ with $\sigma(t)=\max\{t,\alpha t\}$ and $0\le\alpha<1$, and let $\mathcal X\subset\mathbb R^{d_0}$ be the closure of a bounded connected open set. Then for every $B>0$,
\begin{equation}\label{eq:shallow-lower-bound}
\liminf_{\varepsilon\to0}\ \varepsilon^{S-H}\,\mathcal N(\mathcal F,\varepsilon,\|\cdot\|_\infty)
\ \ge\ \frac{C_-}{d_1!},
\end{equation}
where $S-H=(d_0+1)d_1+1$ and $C_->0$ is the architecture-dependent constant defined in \eqref{eq:area-volume} of Appendix~\ref{app:lower-bound-proof}.
\end{theorem}

In Theorem~\ref{thm:shallow-lower-bound}, up to positive scaling, almost every realized function corresponds to $d_1!$ distinct parameter representations, one for each ordering of its hidden neurons. This yields the factor $1/d_1!$ in~\eqref{eq:shallow-lower-bound}. For deeper networks, we formulate this uniqueness property as a general condition on the architecture.

\begin{definition}[Generic identifiability]\label{def:generic-identifiability}
A parameter $\theta\in\mathbb R^S$ is \emph{non-degenerate} if its partial derivatives $\{\partial f(\cdot;\theta)/\partial\theta_e\}_{e=1}^S$ attain the maximum possible rank $S-H$ on the subset of $\mathcal X$ where all pre-activations are nonzero. The architecture is \emph{generically identifiable on $\mathcal X$} if, for almost every non-degenerate $\theta$, any non-degenerate $\theta'$ satisfying $f(\cdot;\theta')=f(\cdot;\theta)$ on $\mathcal X$ is related to $\theta$ by a channel permutation $P$ and a positive scaling $D$, i.e., $\theta'=PD\cdot\theta$.
\end{definition}

Generic identifiability depends only on the architecture and the domain $\mathcal X$. The fully connected networks of Theorem~\ref{thm:shallow-lower-bound} are generically identifiable (Appendix~\ref{app:lower-bound-proof}). For deeper networks, identifiability up to scaling and permutation has been established at specific parameters \citep{phuong2020functional} and on nonempty open subsets \citep{grillo2026identifiable}; Definition~\ref{def:generic-identifiability} formulates the corresponding almost-everywhere property. An open question is which deep architectures satisfy Definition~\ref{def:generic-identifiability}, given that additional hidden symmetries can arise \citep{grigsby2023hidden}.

\begin{theorem}[Covering-number lower bounds for the unified architecture]
\label{thm:sharp-covering-exponent}
Suppose that the bounded domain $\mathcal X\subset\mathbb R^{\Omega_0\times c_0}$ contains a cube $[0,\delta]^{\Omega_0\times c_0}$ for some $0<\delta\le1$ (or an open ball of radius $\delta$ if $|\Omega_0|=1$ and the first block is a fully connected {layer}). Let $H$ and $G$ be as in Theorem~\ref{thm:covering-upper-bound}. Then for any fixed activation slopes $\alpha_1,\ldots,\alpha_L\in[0,1)$ and $B>0$, the following hold:

\textup{(a)} The uniform covering number of $\mathcal F$ satisfies, for all $0<\varepsilon\le2B$,
\begin{equation}\label{eq:main-result}
  \mathcal N(\mathcal F,\varepsilon,\|\cdot\|_\infty)
\ge\Bigl(\gamma_-\,\delta\min\{1,B\}^{{L+L_{\mathrm{res}}+2}}\cdot\varepsilon^{-1}\Bigr)^{S-H},
\end{equation}
where $\gamma_->0$ depends only on the architecture, the slopes $(\alpha_1,\ldots,\alpha_L)$, and $R_{\mathcal X}$.

\textup{(b)} If the architecture is generically identifiable on $\mathcal X$, then
\begin{equation}\label{eq:exact-G}
  \liminf_{\varepsilon\to0}\ \varepsilon^{S-H}\,\mathcal N(\mathcal F,\varepsilon,\|\cdot\|_\infty)
\ \ge\ \frac{C_-}{G},
\end{equation}
where $C_->0$ is the architecture-dependent constant defined in \eqref{eq:area-volume} of Appendix~\ref{app:lower-bound-proof}.
\end{theorem}

Parts~(a) and~(b) of Theorem~\ref{thm:sharp-covering-exponent} rest on complementary arguments: (a) packs a single local parameter cube to establish a nonasymptotic bound for all $0<\varepsilon\le2B$ with explicit dependence on $\delta$ and $B$, whereas (b) integrates over the full scaling-normalized parameter set: under generic identifiability, almost every realized function is counted $G$ times, dividing the resulting $(S-H)$-dimensional volume by $G$ and yielding the asymptotic lower bound~\eqref{eq:exact-G}. Moreover, the generic identifiability condition in part~(b) can be relaxed: if, for almost every non-degenerate $\theta$, the non-degenerate parameters realizing $f(\cdot;\theta)$ on $\mathcal X$ form at most $m$ orbits under positive scaling and channel permutation, the same proof yields $\liminf_{\varepsilon\to0}\varepsilon^{S-H}\mathcal N(\mathcal F,\varepsilon,\|\cdot\|_\infty)\ge C_-/(mG)$, since almost every function is then counted at most $mG$ times.

Combined with Theorem~\ref{thm:covering-upper-bound}, Theorems~\ref{thm:shallow-lower-bound} and~\ref{thm:sharp-covering-exponent} show that the exponent $S-H$ is sharp for each fixed architecture under the stated assumptions. Classical lower bounds focus on VC-dimension or pseudodimension over unbounded parameter spaces \citep{anthony1999neural,bartlett2019nearly}. For bounded parameters under $\|\cdot\|_\infty$, \citet{shen2024complexity} derived $G^{-1}\varepsilon^{-S}$ upper bounds without corresponding lower bounds, while \citet{ou2026covering} established metric-entropy lower bounds that match their upper bounds up to absolute constants for wide, deep ReLU MLPs (width and depth at least $60$) to characterize explicit width and depth dependence. For fully connected ReLU networks, the exponent $S-H$ is in line with the functional dimension established by \citet{grillo2026identifiable}. Complementing these results, Theorems~\ref{thm:shallow-lower-bound} and~\ref{thm:sharp-covering-exponent} apply to MLPs, CNNs, and ResNets of arbitrary depth and hidden widths of at least two, and provide a nonasymptotic lower bound with the exponent $S-H$ and, under the conditions above, an asymptotic lower bound with the divisor $1/G$.

\subsection{Continuous Piecewise-Linear Activated Networks}
\label{sec:cpwl-extension}

The exponent reduction from $S$ to $S-H$ in Sections~\ref{subsec:upper-bounds} and~\ref{subsec:lower-bounds} relies on the positive homogeneity of ReLU and Leaky ReLU activations, $\sigma_l(\lambda t)=\lambda\sigma_l(t)$ for $\lambda>0$. In practice and in approximation theory, neural networks are also equipped with general continuous piecewise-linear (CPWL) activations $\rho_l:\mathbb R\to\mathbb R$ with finitely many breakpoints \citep{goodfellow2013maxout,yarotsky2017,bartlett2019nearly,daubechies2022nonlinear}, such as $\rho(t)=\max\{-1,\min\{t,1\}\}$ or $\operatorname{ReLU6}(t)=\min\{\max\{t,0\},6\}$. Here, a point $t\in\mathbb R$ is a \emph{breakpoint} of $\rho_l$ if $\rho_l$ is not affine on any neighborhood of $t$. When $\rho_l$ has at least two breakpoints, rescaling a neuron's incoming weights changes the distance between its effective breakpoints; consequently, continuous scaling symmetry is lost in general, while layerwise neuron permutation symmetry remains preserved. To quantify the capacity of such networks, we consider the $L$-layer fully connected architecture~\eqref{eq:deep-fnn} with widths $\mathbf d=(d_0,d_1,\ldots,d_L)$ on a bounded domain $\mathcal X\subset\mathbb R^{d_0}$, and let $\mathcal F_{\boldsymbol\rho}$ denote the class of networks with fixed layerwise CPWL activations $\boldsymbol\rho=(\rho_1,\ldots,\rho_L)$ and parameters $\theta\in[-B,B]^S$.

\begin{theorem}[Covering-number upper bound for CPWL networks]
\label{thm:covering-number-cpwl}
Let $\boldsymbol\rho=(\rho_1,\ldots,\rho_L)$ be fixed continuous piecewise-linear activations with finitely many breakpoints, and let $J_{\rho_l}=\operatorname{Lip}(\rho_l)$ denote the Lipschitz constant of $\rho_l$. For every $B\ge1$ and $0<\varepsilon\le2B$, the uniform covering number of $\mathcal F_{\boldsymbol\rho}$ satisfies
\begin{equation}\label{eq:cpwl-covering}
 \mathcal N(\mathcal F_{\boldsymbol\rho},\varepsilon,\|\cdot\|_\infty)
 \le\frac{\Bigl(16(L+1)\bigl(R_{\mathcal X}+\rho_0\bigr)(2B)^{L+2}
 \bigl(\prod_{l=1}^L\bar J_{\rho_l}\bigr)\bigl(\prod_{l=0}^L d_l\bigr)\cdot\varepsilon^{-1}\Bigr)^S}
 {d_1!\times d_2!\times\cdots\times d_L!}
 =:\frac{1}{G}\Bigl(\frac{C_{\boldsymbol\rho}}{\varepsilon}\Bigr)^S,
\end{equation}
where $R_{\mathcal X}=\max\{1,\sup_{x\in\mathcal X}\|x\|_\infty\}$, $\rho_0=\max_{1\le l\le L}|\rho_l(0)|$, $\bar J_{\rho_l}=\max\{1,J_{\rho_l}\}$, $S=\sum_{l=1}^L(d_{l-1}+1)d_l+d_L+1$, and $G=\prod_{l=1}^L d_l!$ (with the expression for $C_{\boldsymbol\rho}$ for general $B>0$ given in Appendix~\ref{app:cpwl}).
\end{theorem}

For fully connected networks with CPWL activations, the upper bound in Theorem~\ref{thm:covering-number-cpwl} has exponent $S$ and includes the permutation factor $1/G$. With multiple breakpoints, the usual scaling symmetry can fail. For instance, under $\rho(t)=\max\{-1,\min\{t,1\}\}$ with breakpoints at $t=\pm 1$, multiplying a neuron's incoming weights and bias by $\lambda>0$ and dividing its outgoing weights by $\lambda$ can change both its active transition interval (from $[-1,1]$ to $[-\lambda^{-1},\lambda^{-1}]$) and its saturation level (from $\pm 1$ to $\pm\lambda^{-1}$). Moreover, whenever each $\rho_l$ is non-affine, two $\rho_l$-neurons in layer $l$ can represent one ReLU neuron on any bounded domain \citep[Proposition~1(b)]{yarotsky2017}. Embedding a ReLU network of widths $\lfloor d_l/2\rfloor$ into $\mathcal F_{\boldsymbol\rho}$ and applying Theorem~\ref{thm:sharp-covering-exponent}(a) yields a corresponding lower bound.

\begin{corollary}[Covering-number lower bound for CPWL networks]
\label{cor:cpwl-lower-bound}
Let $\boldsymbol\rho=(\rho_1,\ldots,\rho_L)$ be fixed non-affine continuous piecewise-linear activations with finitely many breakpoints, and suppose that the bounded domain $\mathcal X\subset\mathbb R^{d_0}$ contains an open ball. Assume $d_l\ge4$ for all $1\le l\le L$. Then there exists a threshold $B_{\boldsymbol\rho}=B_{\boldsymbol\rho}(\boldsymbol\rho,\mathbf d,R_{\mathcal X})>0$, given in~\eqref{eq:cpwl-embedding-bound}, such that for all $B\ge B_{\boldsymbol\rho}$ and $0<\varepsilon\le 1$,
\[
 \mathcal N(\mathcal F_{\boldsymbol\rho},\varepsilon,\|\cdot\|_\infty)
 \ge C_{\boldsymbol\rho,-}\,\varepsilon^{-\bigl(d_0\lfloor d_1/2\rfloor+\sum_{l=2}^L\lfloor d_{l-1}/2\rfloor\lfloor d_l/2\rfloor+\lfloor d_L/2\rfloor+1\bigr)}
 \ge C_{\boldsymbol\rho,-}\,\varepsilon^{-(S-H)/7},
\]
where $C_{\boldsymbol\rho,-}=C_{\boldsymbol\rho,-}(\mathbf d,\mathcal X)>0$ is given in Appendix~\ref{app:cpwl} (and the exponent $(S-H)/7$ tightens to $(S-H)/4$ when the widths $d_l$ are even).
\end{corollary}

In Theorem~\ref{thm:covering-number-cpwl}, scaling symmetry can fail because preactivations in $\mathcal F_{\boldsymbol\rho}$ can traverse multiple breakpoints of $\rho_l$, whose pairwise distances fix an intrinsic scale. Near any single breakpoint, $\rho_l$ consists of two linear pieces and is therefore locally positively homogeneous up to a shift. This observation suggests that if the preactivations in each layer remain within a neighborhood of a single breakpoint, local positive homogeneity---and hence the reduced complexity exponent $S-H$---should be recovered.

To formalize this single-breakpoint regime, for each layer $1\le l\le L$, let $I_l\subset\mathbb R$ be a closed interval of positive length at most $2$ centered at a breakpoint of $\rho_l$ and containing no other breakpoint of $\rho_l$. Writing $\mathbf I=(I_1,\ldots,I_L)$, we define $\mathcal F_{\boldsymbol\rho,\mathbf I}(B)\subset\mathcal F_{\boldsymbol\rho}$ as the subclass of networks with parameters in $[-B,B]^S$ whose layer-$l$ preactivations lie in $I_l$ for all $x\in\mathcal X$ (abbreviated as $\mathcal F_{\rho,I}(B)$ when $\rho_l=\rho$ and $I_l=I$ across all layers). A natural question is whether restricting preactivations to $I_l$ limits the expressive power of the network. The following corollary shows that $\mathcal F_{\boldsymbol\rho,\mathbf I}(B)$ simultaneously recovers the reduced covering-number exponent $S-H$ and retains optimal approximation rates over H\"older smooth function classes \citep{yarotsky2017,schmidthieber2020nonparametric,jiao2023deep,kong2025posterior}.

\begin{corollary}[Single-breakpoint regime: reduced exponent and H\"older approximation]
\label{cor:cpwl-local-range}
Suppose $d_l\ge2$ for all $1\le l\le L$. Then for every $B>0$ and $0<\varepsilon\le4B$,
\begin{equation}\label{eq:cpwl-local-cover}
 \mathcal N(\mathcal F_{\boldsymbol\rho,\mathbf I}(B),\varepsilon,\|\cdot\|_\infty)
 \le\frac{C_{\boldsymbol\rho,\mathbf I}}{G}\,\varepsilon^{-(S-H)},
\end{equation}
where $H=\sum_{l=1}^L d_l$, $G=\prod_{l=1}^L d_l!$, and $C_{\boldsymbol\rho,\mathbf I}>0$ is a constant depending on $(d_0,\ldots,d_L)$, $B$, $R_{\mathcal X}$, $\boldsymbol\rho$, and $\mathbf I$ (given in Appendix~\ref{app:cpwl}). Moreover, when a non-affine CPWL activation $\rho$ and a single-breakpoint interval $I$ are shared across layers, for any H\"older ball $\mathcal H^s_{d_0}(R_H)$ of smoothness $s>0$ and radius $R_H\ge1$ on a bounded input box $\mathcal X$ \citep[following the convention of][]{kong2025posterior}, there exists a parameter bound $B_*=B_*(\rho,I,R_H)>0$, independent of $\delta$, such that for every $0<\delta\le1$, a network architecture with depth $L=O(\log(2/\delta))$ and width $\max_l d_l=O(\delta^{-d_0/(2s)})$ satisfies
\[
 \sup_{f_0\in\mathcal H^s_{d_0}(R_H)}
 \inf_{f\in\mathcal F_{\rho,I}(B_*)}\|f-f_0\|_\infty\le\delta.
\]
\end{corollary}

In Corollary~\ref{cor:cpwl-local-range}, both properties follow from the local structure of $\rho$ on $I$: centering and rescaling the preactivations on $I_l$ reduces $\mathcal F_{\boldsymbol\rho,\mathbf I}(B)$ to the positively homogeneous setting of Theorem~\ref{thm:covering-upper-bound}, yielding~\eqref{eq:cpwl-local-cover}. For the approximation bound, because $\rho$ has a nonzero slope change at the midpoint of $I$, shifting and rescaling inputs into $I$ allows pairs of $\rho$-units to construct localized spike and product approximants while keeping all preactivations within $I$, and distributing the required scaling across $L=O(\log(2/\delta))$ layers keeps the parameter magnitudes bounded by $B_*$ independently of $\delta$ \citep{kong2025posterior}. Together, the covering-number bounds (\eqref{eq:detailed-upper-bound} and~\eqref{eq:cpwl-local-cover}) and this bounded-weight approximation guarantee provide the capacity and approximation control needed for the statistical applications in Section~\ref{sec:consequences}.

\section{Applications and Implications}
\label{sec:consequences}

The covering-number bounds and symmetry reductions in Sections~\ref{sec:settings} and~\ref{sec:main-results} have direct applications to statistical estimation and deep learning theory. First, they yield excess-risk bounds for empirical risk minimization whose rates in the sample size are optimal up to a logarithmic factor and provide finite-sample complexity reductions in channel-sparse architectures (Section~\ref{subsec:risk-bounds}). Second, because our bounds hold under the uniform norm $\|\cdot\|_\infty$, they apply directly to learning settings beyond standard independent and identically distributed (i.i.d.) data, including adversarial training, dependent data, and covariate shift (Section~\ref{subsec:sup-norm}). Third, the underlying scaling and permutation symmetries clarify weight rebalancing in network approximation and loss-landscape degeneracy in optimization (Sections~\ref{subsec:risk-bounds} and~\ref{subsec:geometry}).

\subsection{Statistical Risk Bounds and Complexity Reduction}
\label{subsec:risk-bounds}

In statistical learning, uniform covering numbers control the sample complexity of a hypothesis class through its metric entropy $\log\mathcal N(\mathcal F,\varepsilon,\|\cdot\|_\infty)$ \citetext{\citealp{vanderVaart1996weak}; \citealp{steinwart2008support}; \citealp{wainwright2019high}}. Taking the logarithm in Theorem~\ref{thm:covering-upper-bound} gives, for every $0<\varepsilon\le2B$,
\begin{equation}\label{eq:metric-entropy-bound}
  \log\mathcal N(\mathcal F,\varepsilon,\|\cdot\|_\infty)
  \le(S-H)\log\frac{1}{\varepsilon}+\log C_+-\log G.
\end{equation}
By Theorem~\ref{thm:sharp-covering-exponent}, the exponent $S-H$ is sharp for each fixed architecture, and under generic identifiability, permutation symmetry enters the asymptotic lower bound through the factor $1/G$. To apply~\eqref{eq:metric-entropy-bound} to statistical estimation, let $(X_i,Y_i)_{i=1}^n$ be i.i.d.\ copies of $(X,Y)\in\mathcal X\times\mathcal Y$ with $X\sim P_X$. For a loss function $\ell:\mathbb R\times\mathcal Y\to\mathbb R$, let $R(f)=\mathbb E[\ell(f(X),Y)]$ and $R_n(f)=\frac1n\sum_{i=1}^n\ell(f(X_i),Y_i)$ denote the population and empirical risks, and let $\widehat f\in\arg\min_{f\in\mathcal F}R_n(f)$ be an empirical risk minimizer over $\mathcal F$. Combining~\eqref{eq:metric-entropy-bound} with standard empirical process bounds \citetext{\citealp{bartlett2005local}; \citealp{boucheron2013concentration}; \citealp{wainwright2019high}} and standard minimax lower bounds yields the following excess-risk bounds.

\begin{proposition}[Statistical risk bounds]
\label{prop:risk-bounds}\label{cor:regression}\label{cor:rate}\label{cor:erm}
Let $B\ge1$. In parts~(a) and~(b), let $n\ge1$ and assume that on the range of $\mathcal F\times\mathcal Y$, the loss satisfies $0\le\ell(a,y)\le M_\ell$ and is $L_\ell$-Lipschitz in $a$ ($M_\ell,L_\ell>0$). The lower bounds in these two parts hold with a common absolute constant $c_->0$ whenever $\{0,1\}\subseteq\mathcal Y$.

\textup{(a) (General Lipschitz loss: slow rate).} With $\varepsilon_*=\min\{M_\ell/L_\ell,2B\}$ and $C_\ell=48\max\{M_\ell,\frac{\sqrt\pi}{2}L_\ell\varepsilon_*\}$,
\[
  \mathbb E\bigl[R(\widehat f)\bigr]-\min_{f\in\mathcal F}R(f)
  \le\frac{C_\ell}{\sqrt n}
    \Bigl(\sqrt{(S-H)\log\frac1{\varepsilon_*}+\log C_+-\log G}
          +\sqrt{S-H}\Bigr),
\]
and for the absolute loss, the worst-case excess risk over distributions of $(X,Y)$ satisfies $\inf_{\widetilde f}\sup\bigl\{\mathbb E[R(\widetilde f)]-\min_{f\in\mathcal F}R(f)\bigr\}\ge c_-\,n^{-1/2}$ over all measurable estimators $\widetilde f$.

\textup{(b) (Bernstein condition: fast rate).} If a minimizer $f^\star\in\arg\min_{f\in\mathcal F}R(f)$ satisfies $\operatorname{Var}(\ell(f(X),Y)-\ell(f^\star(X),Y))\le v\bigl(R(f)-\min_{g\in\mathcal F}R(g)\bigr)$ for all $f\in\mathcal F$ and some $v>0$, then for all $n\ge 1/(2BL_\ell)$,
\[
  \mathbb E\bigl[R(\widehat f)\bigr]-\min_{f\in\mathcal F}R(f)
  \le\frac{3}{n}+\frac{2(v+2M_\ell/3)}{n}
    \Bigl((S-H)\log(L_\ell n)+\log C_+-\log G+1\Bigr),
\]
and for the squared loss, the worst-case excess risk over such distributions satisfies $\inf_{\widetilde f}\sup\bigl\{\mathbb E[R(\widetilde f)]-\min_{f\in\mathcal F}R(f)\bigr\}\ge c_-\,n^{-1}$ over all measurable estimators $\widetilde f$.

\textup{(c) (Dependence on $S-H$.)} Suppose that $\mathcal X$ satisfies the domain condition of Theorem~\ref{thm:sharp-covering-exponent} and that $P_X$ has a density that is positive almost everywhere on $\mathcal X$. Consider the regression model $Y=f(X;\theta)+\xi$ with $\theta\in[-B,B]^S$ and noise $\xi\sim\mathcal N(0,\varsigma^2)$ independent of $X$. Under the squared loss, the excess risk of a function $f$ is $\|f-f(\cdot;\theta)\|_{L^2(P_X)}^2$, and
\[
  \liminf_{n\to\infty}\ n\,\inf_{\widetilde f}\sup_{\theta\in[-B,B]^S}\mathbb E_\theta\bigl\|\widetilde f-f(\cdot;\theta)\bigr\|_{L^2(P_X)}^2\ \ge\ \varsigma^2\,(S-H).
\]
\end{proposition}

The upper bounds of parts~(a) and~(b) of Proposition~\ref{prop:risk-bounds} also hold for the CPWL class $\mathcal F_{\boldsymbol\rho,\mathbf I}(B)$ of Corollary~\ref{cor:cpwl-local-range} (with $C_{\boldsymbol\rho,\mathbf I}$ replacing $C_+$). The lower bounds in parts~(a) and~(b) show that the rates $n^{-1/2}$ and $n^{-1}$ of the upper bounds in the sample size cannot be improved in general, up to a logarithmic factor in part~(b). Part~(c) shows that, for regression with Gaussian noise, the factor $S-H$ cannot be improved as $n\to\infty$ (in Bayesian neural networks, the metric-entropy bound~\eqref{eq:metric-entropy-bound} similarly governs posterior contraction rates via the testing framework of \citet{ghosal2000convergence,kong2025posterior}). Although quotienting out symmetries does not alter the asymptotic $n$-rate, it yields a finite-sample reduction whenever the architecture is channel-sparse or locally connected: positive scaling removes the fraction $H/S$ of the exponent, which is quantified after~\eqref{eq:mlp-explicit-bound} in Section~\ref{subsec:upper-bounds}. In addition, $-\log G=\Theta(\sum_l c_l\log c_l)$ offsets part of the width dependence of $\log C_+$.

Part~(b) requires a bounded loss and therefore does not cover regression with Gaussian noise. The following corollary treats this case. It also accounts for the approximation error when the regression function lies outside $\mathcal F$.

\begin{corollary}[Oracle inequality for least-squares regression]
\label{cor:oracle-regression}
Let $B\ge1$ and $M\ge1$. Consider the regression model $Y=f_0(X)+\xi$ with $\|f_0\|_\infty\le M$ and noise $\xi\sim\mathcal N(0,\varsigma^2)$ independent of $X$. Let $\widehat f$ be a least-squares estimator over $\mathcal F_M:=\{f\in\mathcal F:\|f\|_\infty\le M\}$. Then for all $n\ge3\varsigma$,
\[
  \mathbb E\bigl\|\widehat f-f_0\bigr\|_{L^2(P_X)}^2
  \le4\inf_{f\in\mathcal F_M}\|f-f_0\|_{L^2(P_X)}^2
  +\frac{72\max\{\varsigma,M\}^2}{n}\Bigl((S-H)\log\frac{n}{\varsigma}+\log C_+-\log G+6\Bigr).
\]
\end{corollary}

Corollary~\ref{cor:oracle-regression} follows from Theorem~\ref{thm:covering-upper-bound} and the oracle inequality of \citet[Lemma~4]{schmidthieber2020nonparametric}. The corollary has two consequences.

First, it complements the lower bound of Proposition~\ref{prop:risk-bounds}(c). Suppose that $f_0\in\mathcal F$ and $M\ge\sup_{f\in\mathcal F}\|f\|_\infty$. Then the approximation error vanishes, and the risk of the least-squares estimator is at most of order $(S-H)\log n/n$. Under the assumptions of Proposition~\ref{prop:risk-bounds}(c), the minimax risk over $\mathcal F$ has an asymptotic lower bound of order $\varsigma^2(S-H)/n$. For each fixed architecture, the least-squares upper bound matches this rate up to a logarithmic factor.

Second, Corollary~\ref{cor:oracle-regression} yields estimation rates over H\"older balls with a parameter bound that does not grow with $n$. \citet[Theorem~1]{kong2025posterior} use positive scaling in their construction to approximate functions in a fixed H\"older ball by fully connected ReLU networks whose weights and biases are bounded independently of the target accuracy. Combining this result with Corollary~\ref{cor:oracle-regression} gives $\mathbb E\|\widehat f-f_0\|_{L^2(P_X)}^2=O\bigl([(\log^3 n)/n]^{2s/(2s+d_0)}\bigr)$ for $f_0$ in a H\"older ball of smoothness $s$, using networks with uniformly bounded parameters (Appendix~\ref{app:sec4}). This rate is minimax-optimal up to logarithmic factors. It is known for sparse bounded-weight networks \citep{schmidthieber2020nonparametric} and for dense networks with growing weights \citep{jiao2023deep}, and \citet{ou2026covering} remove the logarithmic factors for Lipschitz functions. We state it as an illustration of Corollary~\ref{cor:oracle-regression} and not as a new rate.

\subsection{Uniform-Norm Guarantees Beyond Standard i.i.d.\ Learning}
\label{subsec:sup-norm}

Our covering-number bounds hold under the uniform norm $\|f\|_\infty=\sup_{x\in\mathcal X}|f(x)|$. They therefore also control the capacity of $\mathcal F$ in adversarial training, under dependent data, and under distribution shift. In these settings, capacity bounds tailored to the empirical $L_2(P_n)$ norm or a single fixed $L_p(Q)$ measure \citep{bartlett2017spectral,lin2019generalization,bartlett2019nearly} do not directly transfer, whereas uniform-norm bounds control worst-case perturbations and families of distributions simultaneously.

Specifically, for any probability measure $Q$ on $\mathcal X$ and $1\le p\le\infty$, let $\mathcal N(\mathcal F,\varepsilon,L_p(Q))$ be the covering number under $\|f\|_{L_p(Q)}=(\int_{\mathcal X}|f|^p\,dQ)^{1/p}$, and let $\mathcal N_{[\,]}(\mathcal F,2\varepsilon,L_p(Q))$ be the $L_p(Q)$-bracketing number, defined as the minimum number of function pairs $[\underline f_j,\overline f_j]$ such that every $f\in\mathcal F$ satisfies $\underline f_j(x)\le f(x)\le\overline f_j(x)$ on $\mathcal X$ with $\|\overline f_j-\underline f_j\|_{L_p(Q)}\le 2\varepsilon$. Moreover, for an adversarial radius $\rho\ge0$, define the adversarial loss class $\mathcal L_\rho=\bigl\{\ell_\rho(f,x,y):=\sup_{\|x'-x\|\le\rho,\,x'\in\mathcal X}\ell(f(x'),y):f\in\mathcal F\bigr\}$. Because a uniform $\varepsilon$-ball around $f_j$ forms a pointwise bracket $[f_j-\varepsilon,f_j+\varepsilon]$ of width $2\varepsilon$ and satisfies $|\ell_\rho(f,x,y)-\ell_\rho(f_j,x,y)|\le L_\ell\|f-f_j\|_\infty\le L_\ell\varepsilon$, we have
\begin{equation}\label{eq:sup-decomposition}
  \max\Bigl\{
    \mathcal N\bigl(\mathcal L_\rho,L_\ell\varepsilon,\|\cdot\|_\infty\bigr),\ 
    \sup_Q\,\mathcal N_{[\,]}\bigl(\mathcal F,2\varepsilon,L_p(Q)\bigr),\ 
    \sup_Q\,\mathcal N\bigl(\mathcal F,\varepsilon,L_p(Q)\bigr)
  \Bigr\}
  \le \mathcal N(\mathcal F,\varepsilon,\|\cdot\|_\infty).
\end{equation}
Combined with Theorem~\ref{thm:covering-upper-bound}, the three quantities in~\eqref{eq:sup-decomposition} serve three different problems.

In \emph{adversarial training} ($\rho>0$), the risk $R_\rho(f)=\mathbb E[\ell_\rho(f,X,Y)]$ evaluates the worst-case loss over the perturbation ball $\{\|x'-X\|\le\rho\}$ \citep{madry2018towards,schmidt2018adversarially}. Empirical $L_2(P_n)$ covers do not control this risk, because two functions that agree at the sample points may differ inside these balls \citep{yin2019rademacher}. The first bound in~\eqref{eq:sup-decomposition} covers the adversarial loss class $\mathcal L_\rho$. Hence the upper bound of Proposition~\ref{prop:risk-bounds}(a) holds for adversarial empirical risk minimization, with $R_\rho$ in place of $R$.

For \emph{dependent data and maximum likelihood estimation}, uniform laws of large numbers for $\beta$-mixing sequences and Hellinger rates for nonparametric maximum likelihood estimators are stated in terms of the bracketing entropy $\log\mathcal N_{[\,]}(\mathcal F,2\varepsilon,L_p(Q))$ \citep{yu1994rates,mohri2008rademacher,vanderVaart1996weak,vandegeer2000empirical}. Ordinary $L_p(Q)$ covering entropy does not bound this quantity in general. The second bound in~\eqref{eq:sup-decomposition} provides it with the reduced dimension $S-H$ and the factorial term $-\log G$.

Under \emph{covariate shift and distributionally robust learning}, the target is the risk $R_Q(f)=\mathbb E_Q[\ell(f(X),Y)]$ under a shifted test distribution $Q\ne P$ \citep{shimodaira2000improving,sugiyama2007covariate}, or the worst-case risk $\sup_{Q\in\mathcal Q}R_Q(f)$ over a family $\mathcal Q$ of distributions \citep{ben2013robust,duchi2021learning}. Excess-risk bounds for these problems require covering numbers that hold simultaneously for all $Q\in\mathcal Q$. The third bound in~\eqref{eq:sup-decomposition} gives the bound $(C_+/G)\varepsilon^{-(S-H)}$ uniformly over all probability measures $Q$ on $\mathcal X$, including shifted test distributions and measures supported on low-dimensional sets.

\subsection{Implications for Optimization}
\label{subsec:geometry}

The symmetries also describe the redundancy that parameter-space optimization faces. By Proposition~\ref{prop:scaling-permutation}, the realization map $\theta\mapsto f(\cdot;\theta)$ is constant along each symmetry orbit
\begin{equation}\label{eq:orbit-invariance}
  \mathcal O(\theta)
  =\bigl\{P D\cdot\theta : P\in\Gamma,\ D\in\mathcal D_+\bigr\},
  \qquad\text{where}\quad |\Gamma|=G,\quad \dim(\mathcal D_+)=H,
\end{equation}
with $\Gamma$ denoting the group of channel permutations and $\mathcal D_+$ the group of positive channel scalings. The empirical risk $R_n(f(\cdot;\theta))$ is therefore constant along $\mathcal O(\theta)$. At an interior stationary point $\theta^\star$ where the empirical risk is twice differentiable, the Hessian vanishes on the tangent space to the positive scaling orbit. If the orbit has dimension $H$ at that point, the Hessian has rank at most $S-H$. For a bias-free ReLU network with one hidden layer, \citet[Theorem~4]{dinh2017sharp} show that unrestricted positive rescaling can make the largest eigenvalue of a nonzero Hessian at a smooth minimum arbitrarily large. An $\ell_2$ penalty favors smaller parameter norms along a scaling orbit. If the penalty attains its minimum on the full orbit, the parameters multiplied by each independent channel scale have the same Euclidean norm as those divided by it \citep{neyshabur2015norm,parhi2021banach}. The permutation group $\Gamma$ in turn replicates every minimum into up to $G$ copies $\{P\cdot\theta^\star:P\in\Gamma\}$. Linear interpolation between such copies can encounter a loss barrier, which motivates channel alignment before interpolation or model averaging \citep{wang2020fedma,singh2020model,entezari2022role,ainsworth2023git,jordan2023repair,boufalis2026}.

\section{Numerical Illustrations}
\label{sec:experiments}

In this section, we present numerical experiments that illustrate the parameter symmetries, effective-dimension reductions, and optimization implications established in Sections~\ref{sec:settings}--\ref{sec:consequences}.\footnote{The supplementary archive contains the code and measurements for both experiments.} Section~\ref{subsec:exp-rank} examines the local functional dimension underlying the covering-number exponent $S-H$ by evaluating the numerical rank of the parameter Jacobian across MLP, CNN, and ResNet architectures of varying depths and widths. Section~\ref{subsec:exp-alignment} then examines symmetry-aligned parameter interpolation and multi-model averaging in trained ReLU networks to illustrate the optimization consequences of Section~\ref{subsec:geometry} and the complementary roles of channel permutations and positive scalings.

\subsection{Empirical Local Rank}
\label{subsec:exp-rank}

The lower-bound proof uses a parameter at which $S-H$ directions give independent changes in the network output. We examine this local condition numerically by evaluating the parameter Jacobian at finitely many inputs for each of nine MLP, CNN, and ResNet architectures. The discrete permutation count $G$, by contrast, reflects global identifications between parameter points and cannot be read from this local rank. We examine small, medium, and large networks with ReLU, Leaky ReLU ($\alpha=0.1$), and HardTanh ($\rho(t)=\max\{-1,\min\{t,1\}\}$) activations, nonzero biases, and scalar outputs, using input dimension $d_0=4$ for MLPs and $8\times 8\times 1$ inputs with $3\times 3$ convolutions for CNNs and ResNets. Across five random seeds for each configuration, we calibrate the hidden layers on $2S$ uniform inputs in $[-1,1]^{\Omega_0\times c_0}$ so that the activation breakpoints are active, then fix the parameters and evaluate the numerical rank of $J_n=\bigl(\partial f(x_i;\theta)/\partial\theta_e\bigr)_{i,e}$ on $n=8S$ inputs. For HardTanh, we test a two-breakpoint regime crossing $\pm 1$ (\emph{Two}) and a single-breakpoint regime crossing only $+1$, comparing standard zero padding (\emph{One, zero}) against padding at the local scaling center (\emph{One, center}) for convolutional hidden layers (MLPs require no padding). Full calibration and numerical details are given in Appendix~\ref{app:exp-rank}.

\begin{table}[t]
\centering
\small
\renewcommand{\arraystretch}{1.15}
\setlength{\tabcolsep}{2pt}
\caption{Mean numerical rank of the parameter Jacobian over five calibrated parameter draws. ResNet entries list the first convolutional width followed by residual-block inner widths; HardTanh column labels refer to the regimes defined in the text, and dashes indicate that padding does not apply to MLPs.}
\label{tab:jacobian-rank}
\begin{tabularx}{\textwidth}{@{}lCCCCCCC@{}}
\toprule
Architecture & \multicolumn{2}{c}{Dimension} & \multicolumn{2}{c}{Homogeneous} & \multicolumn{3}{c}{HardTanh} \\
\cmidrule(lr){2-3} \cmidrule(lr){4-5} \cmidrule(l){6-8}
(hidden widths) & $S$ & $S-H$ & ReLU & Leaky ReLU & Two & One, zero & One, center \\
\midrule
MLP $(4,4)$           & $45$   & $37$   & $37$   & $37$   & $45$   & $37$     & \textemdash \\
MLP $(16,16,8)$       & $497$  & $457$  & $457$  & $457$  & $497$  & $457$    & \textemdash \\
MLP $(32,32,32,16)$   & $2817$ & $2705$ & $2705$ & $2705$ & $2817$ & $2705$   & \textemdash \\
\midrule
CNN $(2,2)$           & $61$   & $57$   & $57$   & $57$   & $61$   & $59$     & $57$ \\
CNN $(4,8,4)$         & $633$  & $617$  & $617$  & $617$  & $633$  & $625.4$  & $617$ \\
CNN $(8,12,16,8)$     & $3869$ & $3825$ & $3825$ & $3825$ & $3869$ & $3848.6$ & $3825$ \\
\midrule
ResNet $(2;\,2)$      & $99$   & $95$   & $95$   & $95$   & $99$   & $98.4$   & $95$ \\
ResNet $(4;\,4,4)$    & $637$  & $625$  & $625$  & $625$  & $637$  & $634.8$  & $625$ \\
ResNet $(8;\,8,12,8)$ & $4173$ & $4137$ & $4137$ & $4137$ & $4173$ & $4168.4$ & $4137$ \\
\bottomrule
\end{tabularx}
\end{table}

In Table~\ref{tab:jacobian-rank}, the numerical rank equals $S-H$ for every calibrated ReLU and Leaky ReLU network across all five seeds, consistent with positive channel scaling accounting for the continuous parameter redundancy. In the ResNets, only the first convolutional and inner channels---rather than all hidden channels---are subtracted from $S$, reflecting the coupling of block-output channels induced by identity shortcuts.

The HardTanh columns illustrate the role of activation scale discussed in Section~\ref{sec:cpwl-extension}. When preactivations cross both breakpoints ($\pm 1$), scaling invariance is lost and the empirical rank equals $S$ in every run. When evaluation inputs cross only the breakpoint at $+1$, the rank returns to $S-H$ in every MLP, consistent with Corollary~\ref{cor:cpwl-local-range}. In convolutions, this single-breakpoint reduction also depends on boundary padding. Zero padding leaves the rank above $S-H$ due to boundary--interior offset mismatch under local scaling about $1$, whereas center padding recovers $S-H$. Additional singular-value spectra and scaling-orbit perturbation diagnostics are provided in Appendix~\ref{app:exp-rank} (Figure~\ref{fig:geometry-witnesses}).

\begin{figure}[t]
  \centering
  \includegraphics[width=0.90\textwidth]{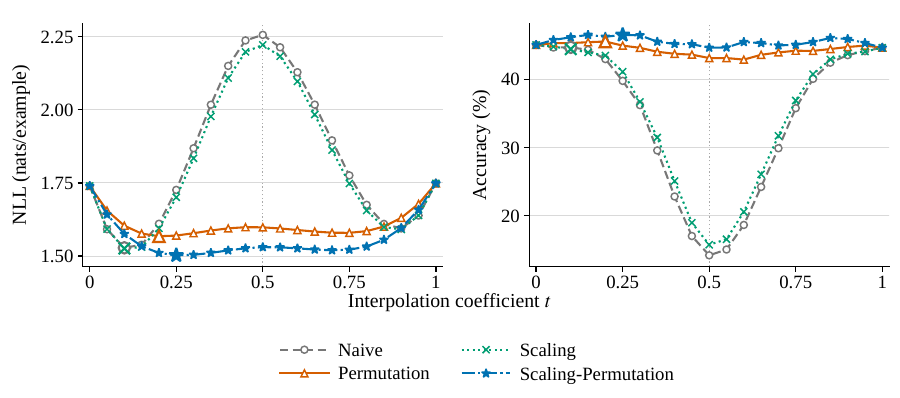}
  \caption{Validation NLL (left) and accuracy (right) along the interpolation path between two CIFAR-10 CNNs (seed~$0$, Epoch~$40$). In both panels, large markers indicate the weight $t$ minimizing validation NLL for each scheme.}
  \label{fig:cnn-interpolation}
\end{figure}

\subsection{Symmetry Alignment and Model Averaging}
\label{subsec:exp-alignment}

The local Jacobian rank in Section~\ref{subsec:exp-rank} reflects the continuous scaling directions $H$, but does not capture the discrete permutation symmetry in parameter space. To study how permutation and positive scaling symmetries act jointly in trained ReLU networks, we turn to parameter alignment and model averaging, as in linear mode connectivity \citep{entezari2022role} and model merging \citep{wang2020fedma,singh2020model,ainsworth2023git,jordan2023repair,boufalis2026}. Because independently trained networks generally differ in their channel orderings and scales, directly averaging their weights often yields high loss. Aligning their parameters through function-preserving permutations and positive scalings before averaging lets us examine how each symmetry affects the loss surface along linear interpolations and at the equal-weight average.

\paragraph{Experimental setup.}
We consider two image classification settings: ReLU MLPs on MNIST \citep{lecun1998gradient} and ReLU CNNs on CIFAR-10 \citep{krizhevsky2009learning}. On MNIST, we randomly partition a $10{,}000$-image training pool into $K\in\{50,100,200\}$ disjoint subsets of $200$, $100$, and $50$ images, and train one MLP independently on each subset (client). On CIFAR-10, we train $K\in\{2,5,10\}$ CNNs independently on disjoint stratified random subsets of $1{,}000$ images each ($100$ per class), so the pooled training set contains $2{,}000$, $5{,}000$, and $10{,}000$ images. We compare four averaging schemes: \emph{Naive} (direct averaging without alignment), \emph{Scaling} (S; positive scaling only), \emph{Permutation} (P; channel permutation only), and \emph{Scaling--Permutation} (SP; joint permutation and positive scaling). Let $\theta^{(1)},\ldots,\theta^{(K)}$ denote the trained parameters of the $K$ clients. For each aligned scheme, we form $\widetilde{\theta}^{(i)}=g_i\cdot\theta^{(i)}$ using a symmetry transformation $g_i=P^{(i)}D^{(i)}$ of channelwise permutations $P^{(i)}$ and positive scalings $D^{(i)}$ (with $P^{(i)}=I$ for S and $D^{(i)}=I$ for P). We choose $g_1,\ldots,g_K$ via alternating updates \citep[extending][]{ainsworth2023git} to minimize $\frac{1}{K}\sum_{i=1}^K\|\widetilde{\theta}^{(i)}-\bar{\theta}\|_2^2$, where $\bar{\theta}=\frac{1}{K}\sum_{i=1}^K\widetilde{\theta}^{(i)}$. We evaluate all models on the standard $10{,}000$-image test sets using negative log-likelihood (NLL) over five random seeds, alongside two reference baselines: \emph{Client mean} (mean individual client NLL) and \emph{Centralized} (training on the pooled data). Accuracy curves for multi-model averaging, numerical tables, and full training and solver details are given in Appendix~\ref{app:exp-alignment}.

\paragraph{Two-model interpolation.}
For $K=2$ CIFAR-10 CNNs, Figure~\ref{fig:cnn-interpolation} shows validation NLL and accuracy along the linear path $(1-t)\widetilde{\theta}^{(1)}+t\widetilde{\theta}^{(2)}$ for $t\in\{0,0.05,\ldots,1\}$. Both \emph{Naive} and pure \emph{Scaling} show a clear loss barrier and accuracy drop near the midpoint $t=0.5$, whereas P and SP have zero measured barrier on this $21$-point grid. Channel permutation accounts for the main improvement, while SP further outperforms P near the equal-weight midpoint in both NLL and accuracy, reflecting the additional benefit of accounting for positive scaling symmetry beyond permutation alone. The five-seed test summary in Table~\ref{tab:sym-interp} exhibits the same pattern, with SP achieving lower midpoint test NLL than P across all five seeds.

\begin{figure}[t]
  \centering
  \includegraphics[width=0.96\textwidth]{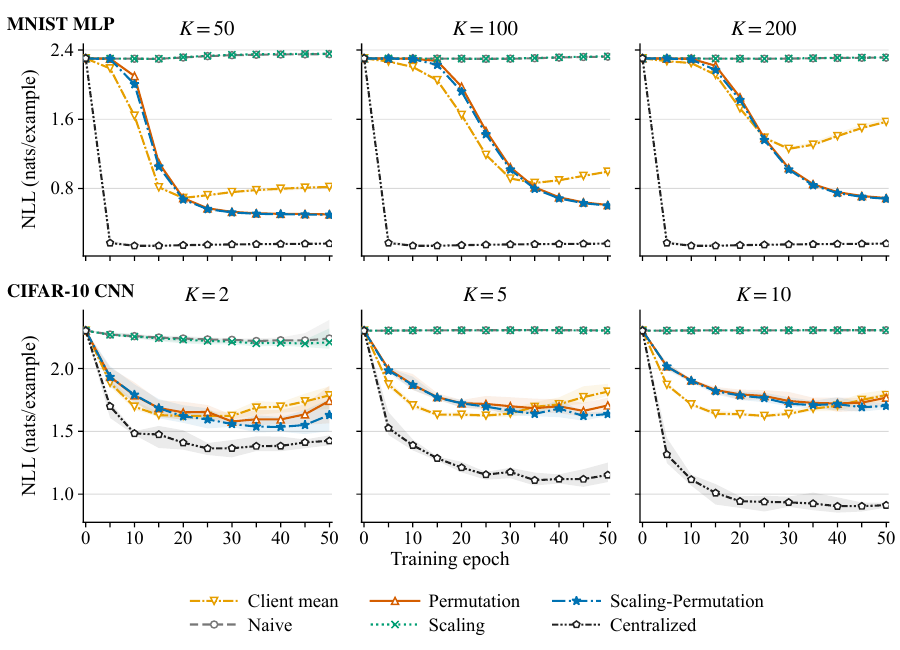}
  \caption{Model averaging on MNIST (top) and CIFAR-10 (bottom). Lines and shaded bands show the mean and range of test NLL across five seeds.}
  \label{fig:multi-nll}
\end{figure}

\paragraph{Multi-model averaging.}
Figure~\ref{fig:multi-nll} shows test NLL for parameter averaging across $K$ independently trained networks (see Appendix~\ref{app:exp-alignment} for the corresponding accuracy curves and Table~\ref{tab:sym-averaging}). Across all settings, \emph{Naive} and pure \emph{Scaling} yield high test NLL, remaining near chance ($\mathrm{NLL}\approx 2.30$) in most cases. On the MNIST MLPs (top row), both P and SP merge up to $K=200$ clients trained on $50$ images each, achieving lower test NLL than \emph{Client mean} while remaining above \emph{Centralized} training, with a small difference between P and SP across all $K$. On the CIFAR-10 CNNs (bottom row), SP shows a clearer improvement over P across $K\in\{2,5,10\}$, especially in later epochs ($30$--$50$). Overall, joint permutation--scaling alignment improves upon permutation or scaling alone across the examined settings.

\section{Conclusion}
\label{sec:conclusion}

This paper established uniform covering-number bounds with a sharp exponent for ReLU and Leaky ReLU networks across a unified architecture encompassing MLPs, CNNs, and ResNets. We proved that positive channel scalings and channel permutations exhaust all global quasianalytic parameter equivalences on a continuous domain, and showed that quotienting out both symmetries reduces the upper-bound exponent from $S$ to $S-H$ while dividing the bound by the factorial permutation count $G$. Matching lower bounds show that the exponent $S-H$ is sharp for each fixed architecture, and that the $G$-fold permutation symmetry yields a factor $1/G$ in the asymptotic lower bound for one-hidden-layer fully connected networks and, more generally, under generic identifiability. We further extended this analysis to fully connected networks with general CPWL activations, obtaining permutation-reduced upper bounds with exponent $S$ in general and $S-H$ in a single-breakpoint regime. We applied our covering-number bounds to statistical risk estimation in i.i.d.\ settings and to capacity control beyond i.i.d.\ data, and discussed how the underlying symmetries govern weight rebalancing in approximation and flat directions in optimization. Numerical experiments further illustrate these findings, both locally through empirical Jacobian ranks across architectures and activation regimes and through symmetry-aligned interpolation and model averaging in trained ReLU networks. Two directions remain for future work. First, establishing explicit architectural conditions under which deeper networks satisfy generic identifiability would extend the asymptotic $1/G$ lower bound beyond one hidden layer. Second, extending the analysis from global equivalence on a continuous domain $\mathcal X$ to equivalence on a finite sample $\{x_1,\ldots,x_n\}$ could capture data-dependent redundancies and tighten empirical $L_2(P_n)$ covering numbers in overparameterized regimes.

\bibliography{ref.bib}

\clearpage
\appendix

Appendix~\ref{app:exp-details} provides the experimental details and supplementary results for Section~\ref{sec:experiments}. Appendix~\ref{app:sec2} proves Propositions~\ref{prop:scaling-permutation} and~\ref{prop:completeness}, Appendices~\ref{app:main} and~\ref{app:lower-bound-proof} prove the upper bounds (Theorem~\ref{thm:covering-upper-bound} and Corollary~\ref{thm:covering-number-sp-deep}) and the lower bounds (Theorems~\ref{thm:shallow-lower-bound} and~\ref{thm:sharp-covering-exponent}), Appendix~\ref{app:cpwl} proves the results on continuous piecewise-linear activations (Section~\ref{sec:cpwl-extension}), and Appendix~\ref{app:sec4} proves Proposition~\ref{prop:risk-bounds} and Corollary~\ref{cor:oracle-regression}.

\section{Experimental Details and Supplementary Results}
\label{app:exp-details}

This appendix provides the architectural, optimization, and numerical specifications for Section~\ref{sec:experiments}, together with supplementary diagnostics, accuracy curves, and summary tables.

\subsection{Empirical Local Rank and Scaling-Orbit Diagnostics}
\label{app:exp-rank}

\paragraph{Experimental setup and rank computation.}
In Table~\ref{tab:jacobian-rank}, MLPs take inputs in $[-1,1]^4$, whereas CNNs and ResNets take single-channel $8\times 8$ inputs in $[-1,1]^{8\times 8\times 1}$, use $3\times 3$ convolutions, and apply spatial average pooling before the scalar linear output. Each ResNet begins with a convolutional layer followed by residual blocks as in Figure~\ref{fig:block}. All input coordinates are drawn independently and uniformly from $[-1,1]$. Across five random seeds, all activation regimes for a given architecture share the same input sample, output weights, and initial hidden weight directions. Weights are initialized as independent centered Gaussians with variance inversely proportional to the number of incoming weights per unit, and the output bias has standard deviation $0.1$.

Because uncalibrated random weights can leave some hidden units purely linear or inactive across the sampled inputs, we calibrate each hidden layer sequentially on an initial sample of $2S$ inputs so that every channel crosses its intended activation breakpoints. Specifically, for ReLU and Leaky ReLU, we subtract each channel's sample mean so that its preactivations take both positive and negative values. For two-breakpoint HardTanh, we center and rescale each channel so that its preactivations cross both $-1$ and $+1$. For single-breakpoint HardTanh, we shift and rescale each channel into a narrow band around $+1$ so that every channel crosses $+1$ while remaining strictly above $-1$ throughout the evaluation sample. For the convolutional and residual networks, the two single-breakpoint columns in Table~\ref{tab:jacobian-rank} share identical network parameters and differ only in how hidden feature maps are padded at the spatial boundary: standard zero padding versus center padding ($1$ after each HardTanh layer and $1+k$ after $k$ residual additions).

After calibration, we hold all parameters fixed and evaluate the $n\times S$ Jacobian $J_n$ on $n=8S$ inputs (the $2S$ calibration inputs together with $6S$ additional uniform inputs), on which the activation breakpoint ranges are verified. We compute the singular values $\sigma_1 \ge \cdots \ge \sigma_S \ge 0$ of $J_n$ and define its numerical rank as the number of singular values exceeding $10^{-10}\sigma_1$. Changing the relative cutoff to $10^{-12}$ leaves all $210$ runs unchanged, while raising it to $10^{-8}$ alters only three.

\begin{figure}[!htb]
  \centering
  \includegraphics[width=0.96\linewidth]{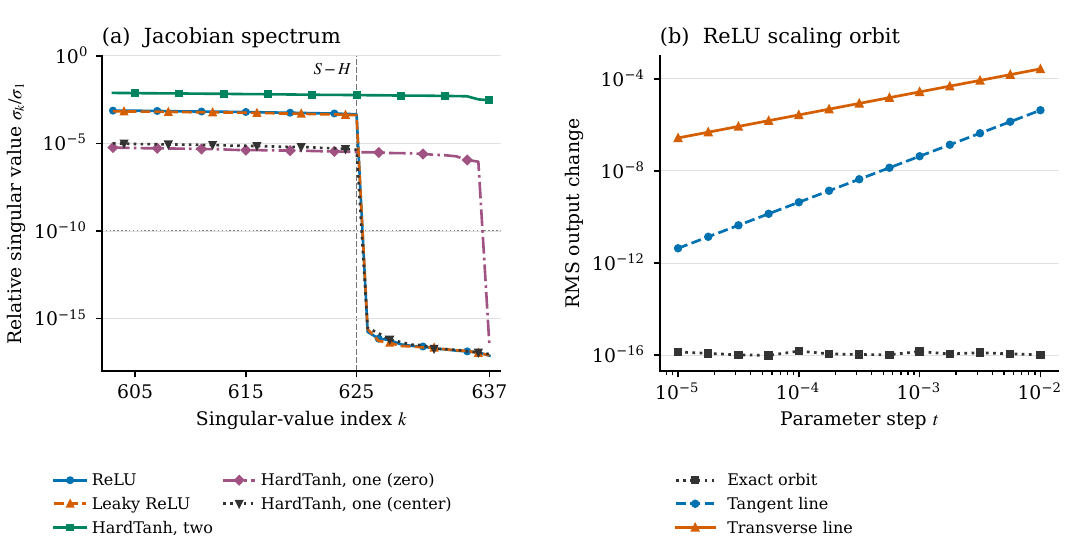}
  \caption{Local geometry diagnostics for the medium ResNet ($S=637$, $S-H=625$, seed~$0$) evaluated on $n=5{,}096$ inputs. \emph{(a)}~Tail of the normalized Jacobian singular values $\sigma_k/\sigma_1$; the vertical line marks $S-H$ and the horizontal line marks the cutoff $10^{-10}$, with the two single-breakpoint HardTanh curves sharing identical parameters and inputs. \emph{(b)}~Root-mean-square (RMS) output change on the same inputs along three unit-speed parameter paths starting from the ReLU parameter vector.}
  \label{fig:geometry-witnesses}
\end{figure}

\paragraph{Singular-value spectra and scaling-orbit diagnostic.}
Figure~\ref{fig:geometry-witnesses} shows two local diagnostics for the medium ResNet (seed~$0$, $n=5{,}096$). In panel~(a), the normalized singular values $\sigma_k/\sigma_1$ drop abruptly to machine precision immediately after $S-H=625$ for ReLU, Leaky ReLU, and center-padded single-breakpoint HardTanh, whereas two-breakpoint and zero-padded single-breakpoint HardTanh remain well above the cutoff beyond $625$.

Panel~(b) compares an exact positive scaling orbit against straight-line parameter perturbations starting from the same ReLU parameter vector with unit initial speed over step sizes $t\in[10^{-5},10^{-2}]$. Along the scaling orbit, the RMS output change stays at numerical precision ($\sim 10^{-16}$). Along the straight line tangent to the orbit at $t=0$, the first derivative vanishes and the curvature of the orbit produces a quadratic output change (fitted log--log slope $2.00$), whereas the straight line orthogonal to the scaling directions produces a linear change (slope $1.00$).

\subsection{Symmetry Alignment and Model Averaging}
\label{app:exp-alignment}

\begin{figure}[tbp]
  \centering
  \includegraphics[width=0.96\linewidth]{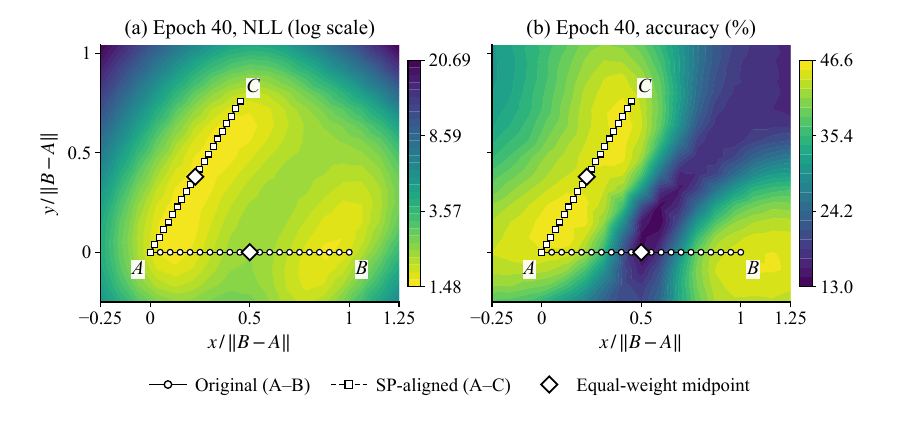}
  \caption{CIFAR-10 validation NLL (left) and accuracy (right) landscapes ($K=2$, seed~$0$) in the affine plane spanned by $\theta_A=\theta^{(1)}$, $\theta_B=\theta^{(2)}$, and $\theta_C=g_1^{-1}g_2\cdot\theta^{(2)}$ (labeled $A,B,C$), preserving Euclidean distance proportions normalized by $\|\theta_B-\theta_A\|_2$, at Epoch~$40$. Circles, squares, and diamonds mark $A\text{--}B$, $A\text{--}C$, and their midpoints.}
  \label{fig:sp-euclidean-landscape}
\end{figure}

\paragraph{Experimental setup and evaluation.}
On CIFAR-10, we use a fixed stratified split with $1{,}000$ images ($100$ per class) for learning-rate scheduling, $4{,}000$ images ($400$ per class) for validation, and ten disjoint subsets of $1{,}000$ images each ($100$ per class) for the clients; for $K\in\{2,5,10\}$, we use the first $K$ clients (pooled sizes $2{,}000$, $5{,}000$, and $10{,}000$). The CNN consists of three $3\times 3$ convolutional layers (stride~$1$, padding~$1$) with $32$, $64$, and $128$ channels, each followed by ReLU; the first two layers end with $2\times 2$ max pooling, and the third is followed by spatial average pooling to a $4\times 4$ grid and a linear output layer. All layers include bias terms, giving $113{,}738$ parameters in total. Because channel permutations commute with spatial pooling and max pooling is positively homogeneous, the same channelwise symmetries apply to these multi-class CNNs. Input images are scaled to $[-1,1]$; during training, we apply standard data augmentation (random cropping with $4$-pixel zero padding and random horizontal flips) and train for $50$ epochs with AdamW \citep{loshchilov2019decoupled} (initial learning rate $0.003$, weight decay $10^{-4}$, batch size $128$, with learning-rate decay guided by the $1{,}000$-image scheduling set).

On MNIST, we use a fixed split with $10{,}000$ images for validation and $10{,}000$ images as the training pool. For each random seed, the training pool is randomly partitioned into $K\in\{50,100,200\}$ equal disjoint subsets of $200$, $100$, and $50$ images. The ReLU MLP has layer widths $(784,1024,512,10)$ with bias terms ($1{,}333{,}770$ parameters in total) and pixel values scaled to $[0,1]$. Each model is trained for $50$ epochs using SGD (learning rate $0.05$, momentum $0.9$, no weight decay, batch size $128$).

\begin{figure}[!htb]
\centering
\includegraphics[width=0.94\linewidth]{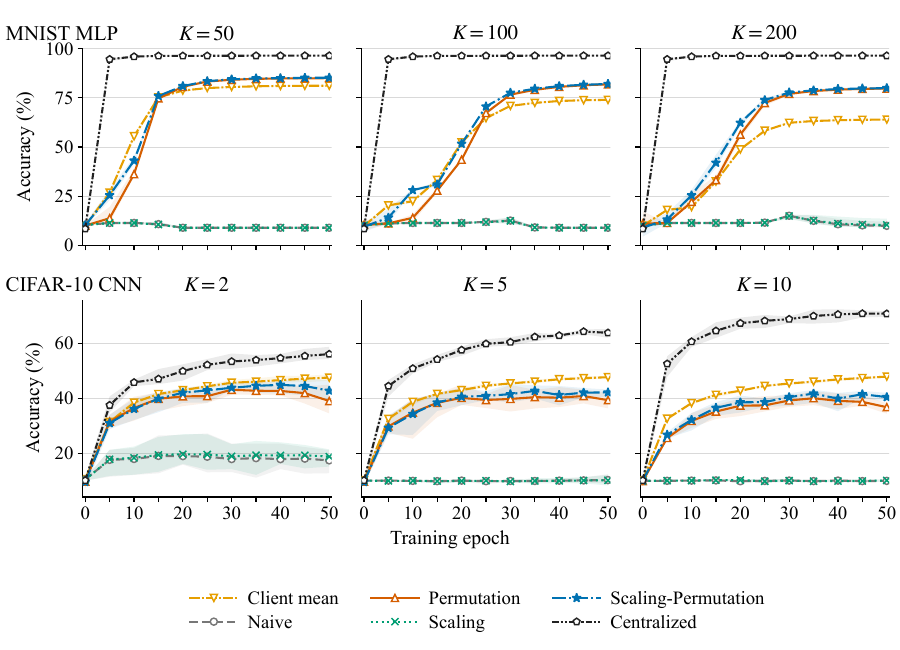}
\caption{Test accuracy for multi-client averaging on MNIST MLPs (top; $K\in\{50,100,200\}$) and CIFAR-10 CNNs (bottom; $K\in\{2,5,10\}$) across epochs $0,5,\ldots,50$ (mean and min--max range over five seeds), complementing Figure~\ref{fig:multi-nll}.}
\label{fig:multi-acc}
\end{figure}

All models are evaluated on the standard $10{,}000$-image test sets across five random seeds ($0$--$4$) using classification accuracy and the average cross-entropy loss (NLL) of the ten output logits $z_{i,1},\ldots,z_{i,10}$,
\[
  L = \frac{1}{n}\sum_{i=1}^n \Bigl[\log\sum_{c=1}^{10}\exp(z_{i,c}) - z_{i,y_i}\Bigr].
\]
Within each seed, all four averaging schemes are evaluated on the same client checkpoints at epochs $0,5,\ldots,50$ (where Epoch~$0$ is the initialization), and shaded bands in Figures~\ref{fig:multi-nll} and~\ref{fig:multi-acc} show the minimum--maximum range across seeds. Whenever checkpoints are selected on the validation set, we choose the epoch minimizing validation NLL (breaking ties by higher validation accuracy, then earlier epoch). In Table~\ref{tab:sym-interp}, all four schemes are compared at a common epoch per seed chosen by equal-weight SP validation NLL (epochs $40,35,35,45,35$ across seeds $0$--$4$). In Table~\ref{tab:sym-averaging}, each entry reports the validation-selected value, where the epoch is chosen separately for each scheme, seed, and $K$ (for \emph{Client mean}, each client selects its own epoch before averaging); the value in parentheses is evaluated at the synchronized final epoch (Epoch~$50$).

\paragraph{Alignment solver.}
We minimize the mean squared parameter distance of Section~\ref{subsec:exp-alignment} (over all weights and biases, without using labels) by block coordinate descent under the channelwise action of Proposition~\ref{prop:scaling-permutation}, iteratively aligning each client to the mean of the other $K-1$ clients. Layerwise permutation updates are solved as linear assignment problems \citep{ainsworth2023git,crouse2016implementing}, and single-channel scaling updates minimize the resulting one-dimensional objective over $d\in[1/32,32]$ via its quartic first-order condition. The \emph{Scaling} scheme initializes from the identity permutation, whereas \emph{SP} initializes from the \emph{Permutation} solution and alternates permutation and scaling updates until a stopping tolerance or iteration limit is reached; full implementation details are provided in the supplementary code.

\paragraph{Two-model landscapes and interpolation.}
For $K=2$ CIFAR-10 CNNs with alignment maps $g_1,g_2$, set $\theta_A=\theta^{(1)}$, $\theta_B=\theta^{(2)}$, and $\theta_C=g_1^{-1}g_2\cdot\theta^{(2)}$, so that the segment $\theta_A\text{--}\theta_C$ realizes the same functions as the jointly aligned segment. Figure~\ref{fig:sp-euclidean-landscape} plots the validation landscape in the affine plane spanned by $(\theta_A,\theta_B,\theta_C)$ with Euclidean distance proportions \citep[cf.][]{benzing2022random}. Although $\theta_B$ and $\theta_C$ realize the same function up to numerical precision, linear interpolation along $\theta_A\text{--}\theta_B$ crosses a high-loss barrier, whereas $\theta_A\text{--}\theta_C$ stays in a low-loss region. This improvement is not governed by Euclidean distance alone: with $\theta_A$ fixed, P places the aligned endpoint slightly closer to $\theta_A$ than SP, yet yields higher test NLL at the midpoint.

Along the $21$-point grid $T=\{0,0.05,\ldots,1\}$, let $L(t)$ denote test NLL, let $t_\star\in T$ minimize validation NLL, and define the empirical barrier
\[
  B_T = \max_{t\in T}\bigl\{L(t) - (1-t)L(0) - tL(1)\bigr\} \ge 0.
\]
Table~\ref{tab:sym-interp} summarizes the five-seed test results; SP achieves lower midpoint test NLL than P in all five seeds, and the same advantage persists when the common epoch is instead selected by P validation NLL or fixed at Epoch~$50$ (cf.\ Table~\ref{tab:sym-averaging}).

\begin{table}[tbp]
\centering
\small
\renewcommand{\arraystretch}{1.15}
\caption{Five-seed mean test performance for CIFAR-10 two-client interpolation at the common epoch selected by equal-weight SP validation NLL (parentheses denote sample SD).}
\label{tab:sym-interp}
\begin{tabularx}{0.92\textwidth}{@{}lCCCCC@{}}
\toprule
 & Barrier & \multicolumn{2}{c}{Midpoint ($t=0.5$)} & \multicolumn{2}{c}{Selected weight ($t_\star$)} \\
\cmidrule(lr){3-4} \cmidrule(l){5-6}
Scheme & $B_T$ & NLL & Acc.\ (\%) & NLL & Acc.\ (\%) \\
\midrule
Naive & $0.52\ (0.03)$ & $2.23\ (0.02)$ & $17.27$ & $1.52$ & $45.67$ \\
S     & $0.50\ (0.02)$ & $2.21\ (0.02)$ & $18.41$ & $1.52$ & $45.84$ \\
P     & $0.00\ (0.00)$ & $1.57\ (0.03)$ & $43.57$ & $1.51$ & $46.56$ \\
SP    & $0.00\ (0.00)$ & $1.51\ (0.03)$ & $45.61$ & $1.48$ & $47.12$ \\
\bottomrule
\end{tabularx}
\end{table}

\paragraph{Multi-model accuracy and summary table.}
Figure~\ref{fig:multi-acc} shows the test accuracy trajectories corresponding to Figure~\ref{fig:multi-nll}, and Table~\ref{tab:sym-averaging} reports five-seed mean test NLL and accuracy under both validation-based checkpoint selection and synchronized Epoch~$50$. Under both selection protocols the five-seed mean test NLL is lower for SP than for P on CIFAR-10. Relative to \emph{Client mean}, the comparison depends on the selection protocol: on CIFAR-10 at $K=5$, SP is lower in test NLL by $0.18$ at Epoch~$50$, whereas under independent checkpoint selection it is higher by $0.02$. At Epoch~$50$ on CIFAR-10, the mean maximum predicted probability on misclassified test examples also decreases from $0.61$--$0.62$ (\emph{Client mean}) to $0.43$, $0.29$, and $0.28$ (SP) for $K=2,5,10$.

\begin{table}[!htb]
\centering
\small
\renewcommand{\arraystretch}{1.05}
\setlength{\tabcolsep}{3pt}
\caption{Five-seed mean test performance of the multi-client averaging schemes: \emph{(a)}~test NLL (nats/example); \emph{(b)}~test accuracy (\%). Each entry reports the validation-selected value (epoch chosen per scheme, seed, and $K$ by validation NLL; \emph{Client mean} selects each client's epoch individually before averaging), with the synchronized Epoch-$50$ value in parentheses.}
\label{tab:sym-averaging}

\emph{(a)}~Test NLL (nats/example)\\[2pt]
\begin{tabularx}{\textwidth}{@{}lCCCCCC@{}}
\toprule
& \multicolumn{3}{c}{CIFAR-10} & \multicolumn{3}{c}{MNIST} \\
\cmidrule(lr){2-4} \cmidrule(l){5-7}
Scheme & $K=2$ & $K=5$ & $K=10$ & $K=50$ & $K=100$ & $K=200$ \\
\midrule
Client mean & $1.59$~($1.79$) & $1.59$~($1.82$) & $1.59$~($1.79$) & $0.68$~($0.82$) & $0.86$~($0.99$) & $1.22$~($1.57$) \\
S           & $2.18$~($2.21$) & $2.30$~($2.30$) & $2.30$~($2.31$) & $2.30$~($2.36$) & $2.30$~($2.33$) & $2.30$~($2.31$) \\
P           & $1.56$~($1.74$) & $1.65$~($1.71$) & $1.70$~($1.77$) & $0.50$~($0.50$) & $0.61$~($0.61$) & $0.69$~($0.69$) \\
SP          & $1.51$~($1.63$) & $1.61$~($1.64$) & $1.68$~($1.70$) & $0.50$~($0.50$) & $0.60$~($0.60$) & $0.68$~($0.68$) \\
Centralized & $1.36$~($1.42$) & $1.09$~($1.15$) & $0.88$~($0.91$) & $0.14$~($0.16$) & $0.14$~($0.16$) & $0.14$~($0.16$) \\
\bottomrule
\end{tabularx}

\smallskip
\emph{(b)}~Test accuracy (\%)\\[2pt]
\begin{tabularx}{\textwidth}{@{}lCCCCCC@{}}
\toprule
& \multicolumn{3}{c}{CIFAR-10} & \multicolumn{3}{c}{MNIST} \\
\cmidrule(lr){2-4} \cmidrule(l){5-7}
Scheme & $K=2$ & $K=5$ & $K=10$ & $K=50$ & $K=100$ & $K=200$ \\
\midrule
Client mean & $44.00$~($47.54$) & $44.48$~($47.68$) & $44.63$~($47.90$) & $79.24$~($81.18$) & $72.73$~($73.98$) & $62.86$~($63.95$) \\
S           & $19.28$~($18.75$) & $10.17$~($10.13$) &  $9.95$~($10.08$) & $11.07$~($8.92$) & $11.80$~($8.92$) & $11.46$~($10.35$) \\
P           & $43.74$~($39.02$) & $41.74$~($39.32$) & $40.35$~($36.76$) & $85.02$~($85.02$) & $81.87$~($81.87$) & $79.75$~($79.75$) \\
SP          & $45.61$~($42.77$) & $43.07$~($42.14$) & $41.54$~($40.42$) & $85.28$~($85.25$) & $82.05$~($82.05$) & $79.96$~($79.96$) \\
Centralized & $53.04$~($56.09$) & $63.47$~($63.85$) & $70.95$~($70.80$) & $96.15$~($96.45$) & $96.15$~($96.45$) & $96.15$~($96.45$) \\
\bottomrule
\end{tabularx}
\end{table}

\clearpage

\section{Functional Equivalence}
\label{app:sec2}

The proofs in Appendices~\ref{app:sec2}--\ref{app:lower-bound-proof} proceed layer by layer, where a layer consists of one affine map followed by one activation. A fully connected or convolutional block is one layer. A residual block consists of two layers, its \emph{inner layer} and its \emph{outer layer}, and its shortcut adds the input of the block to the output of the outer layer. We number the layers by $l=1,\ldots,\widetilde L$, where $\widetilde L:=L+L_{\mathrm{res}}$, and write $\mathcal R$ for the set of outer layers. Layer $l$ outputs a feature map $h_l$ with $c_l$ channels on the grid $\Omega_l$. It uses the activation $\sigma_l$ of its block, the matrix $A_l$ of its linear map on flattened feature maps, and the bias $b_l\in\R^{c_l}$, which is repeated over positions as $\bar b_l(p,j)=b_{l,j}$. We omit the subscript of $\sigma_l$ when the layer is clear from context. The preactivation and the output of layer $l$ are
\begin{equation}\label{eq:preactivation}
 z_l=A_lh_{l-1}+\bar b_l,\qquad
 h_l=\sigma_l(z_l)+\mathbf 1_{\{l\in\mathcal R\}}h_{l-2},\qquad h_0=x,
\end{equation}
and the network output is $f(x;\theta)=|\Omega_{\widetilde L}|^{-1}\sum_{p\in\Omega_{\widetilde L}}a^\top h_{\widetilde L}(p)+\beta$. We call each coordinate $v=(l,p,j)$ of $z_l$ a \emph{neuron} of layer $l$ and write $z_v$ for its preactivation.

\begin{proof}[Proof of Proposition~\ref{prop:scaling-permutation}]
The proof is an induction over the layers. For each layer $l$, let $Q_l$ be the matrix that $PD\cdot\theta$ assigns to the output channels of layer $l$. Thus $Q_0=I$, $Q_l=P_k^{(1)}D_k^{(1)}$ if $l$ is the inner layer of a residual block $k$, and $Q_l=P_kD_k$ if $l$ is any other layer of a block $k$. The coupling $(P_k,D_k)=(P_{k-1},D_{k-1})$ of a residual block $k$ gives $Q_l=Q_{l-2}$ for every outer layer $l\in\mathcal R$. In this notation, the updates of Proposition~\ref{prop:scaling-permutation} read
\[
 A_l'=Q_lA_lQ_{l-1}^{-1},\qquad \bar b_l'=Q_l\bar b_l,\qquad a'^\top=a^\top Q_{\widetilde L}^{-1},\qquad\beta'=\beta,
\]
where $Q_l$ acts on the channels at every position. We show that $z_l'=Q_lz_l$ and $h_l'=Q_lh_l$ for every layer $l$.

For $l=0$, we have $h_0'=x=Q_0h_0$. Suppose that the claim holds for all layers before $l$. Then
\[
 z_l'=A_l'h_{l-1}'+\bar b_l'
 =Q_lA_lQ_{l-1}^{-1}Q_{l-1}h_{l-1}+Q_l\bar b_l=Q_lz_l .
\]
The matrix $Q_l$ is the product of a permutation matrix and a positive diagonal matrix, and $\sigma_l$ acts coordinatewise with $\sigma_l(\lambda t)=\lambda\sigma_l(t)$ for $\lambda>0$. Therefore $\sigma_l(Q_lz)=Q_l\sigma_l(z)$, and with $Q_l=Q_{l-2}$ for $l\in\mathcal R$ we obtain
\[
 h_l'=\sigma_l(Q_lz_l)+\mathbf 1_{\{l\in\mathcal R\}}Q_{l-2}h_{l-2}
 =Q_l\bigl(\sigma_l(z_l)+\mathbf 1_{\{l\in\mathcal R\}}h_{l-2}\bigr)=Q_lh_l .
\]
This proves the claim. At the last layer, $a'^\top h_{\widetilde L}'(p)=a^\top Q_{\widetilde L}^{-1}Q_{\widetilde L}h_{\widetilde L}(p)=a^\top h_{\widetilde L}(p)$ at every position $p\in\Omega_{\widetilde L}$. Averaging over $p$ and adding $\beta'=\beta$ gives $f(x;PD\cdot\theta)=f(x;\theta)$.
\end{proof}

\begin{proof}[Proof of Proposition~\ref{prop:completeness}]
On every region of $(\theta,x)$ with a constant activation pattern, $f$ is a polynomial whose coefficients are sums of products of parameters along paths through the network. Step~1 shows that some power of $T$ fixes every such path product, and that every map with this property is a positive scaling. The remaining steps identify the channel permutation contained in $T$ by comparing the inputs at which the networks with parameters $\theta$ and $T(\theta)$ are not differentiable. Step~2 removes a positive scaling from $T$ and obtains a map $\widehat T$ that is linear in the biases. Steps~3 and~4 show that, up to a channel permutation and a positive scaling, the bias map of $\widehat T$ is triangular with respect to the layers and its kernel maps are linear. Step~5 shows layer by layer that the normalized parameter equals $\theta$. Hence $T(\theta)$ is a positive scaling of $P\cdot\theta$ for a fixed channel permutation $P$, and Step~1, applied to the map $\theta\mapsto P^{-1}\cdot T(\theta)$, completes the proof.

\medskip
\noindent\textbf{Preliminaries.}
Let $\mathcal{C}^{\rm qa}$ denote the quasianalytic class of \citet[Section~3]{bierstone2004resolution} that contains the components of $T$. We use the following properties of $\mathcal{C}^{\rm qa}$. The class contains all polynomials. It is closed under composition, differentiation, products, reciprocals of nonvanishing functions, and division by a coordinate: if $g\in\mathcal{C}^{\rm qa}$ vanishes on $\{u_i=0\}$, then $g/u_i\in\mathcal{C}^{\rm qa}$. It satisfies the identity principle: a function in $\mathcal{C}^{\rm qa}$ that vanishes on a nonempty open subset of a connected domain vanishes identically. We say that a property holds for \emph{generic} $\theta$ if it holds outside the zero set of a nonzero function in $\mathcal{C}^{\rm qa}$. By the identity principle, such a zero set is closed and nowhere dense.

\medskip
\noindent\textbf{Notation.}
Write $\theta=(w,b)$, where $w$ collects the entries of the weight matrices and kernels and the readout weights $a_j$, and $b$ collects the biases $b_{l,j}$ and $\beta$. Neurons are indexed as in~\eqref{eq:preactivation}. A \emph{path} $\pi$ starts at an input coordinate $x_a$ or at a bias, runs through the network along weights and shortcuts, and ends at the readout. Its \emph{path product} $w_\pi(\theta)$ is the product of the parameters along $\pi$. This product includes the bias if $\pi$ starts at a bias, and it includes the readout factor $a_j/|\Omega_{\widetilde L}|$. Shortcuts contribute the factor $1$, and the path of $\beta$ has the product $\beta$. We write $\Pi(a)$ for the set of paths that start at $x_a$ and $\Pi_{\rm b}$ for the set of paths that start at a bias. We use three elementary facts. Two distinct paths in $\Pi(a)$, or in $\Pi_{\rm b}$, pass through different sets of neurons. Two distinct paths in $\Pi(a)$ have different path products, because the weights along such a path determine its neurons one after another. By the padding convention of Section~\ref{sec:settings}, every neuron and every weight lies on a path that starts at an input coordinate.

An \emph{activation pattern} $\omega$ assigns to every neuron $v$ of a layer $l$ a slope $\omega_v\in\{1,\alpha_l\}$, and $\omega_\pi:=\prod_{v\in\pi}\omega_v$ denotes the product of the slopes of the neurons on a path $\pi$. We call $\omega$ \emph{realized} if there is a nonempty open set of points $(\theta,x)$, with $x$ in the interior of $\mathcal X$, on which every neuron $v$ has a nonzero preactivation, positive if $\omega_v=1$ and negative otherwise. On such a set, $f$ equals the polynomial
\[
 F_\omega(\theta,x)=\sum_{a}\Bigl(\sum_{\pi\in\Pi(a)} \omega_\pi w_\pi(\theta)\Bigr)x_a
   +\sum_{\pi\in\Pi_{\rm b}}\omega_\pi w_\pi(\theta),
\]
and by continuity $f=F_\omega$ also holds on the closure of this set.

\medskip
\noindent\textbf{Step 1 (A power of $T$ is a positive scaling).}
\emph{(1a) $T$ permutes the polynomial pieces.}
Let $\mathfrak{P}$ be the finite set of the polynomials $F_\omega$ of all realized patterns $\omega$. Let $\omega$ be realized on the open set $O$. The closed sets on which $f=F_{\omega'}$, with $\omega'$ realized, are finitely many and cover all points $(\theta,x)$ with $x$ in the interior of $\mathcal X$. By the Baire category theorem, $O$ has a nonempty open subset that the map $(\theta,x)\mapsto(T(\theta),x)$ sends into one of these sets. On this subset, $F_{\omega'}(T(\theta),x)=f(x;T(\theta))=f(x;\theta)=F_\omega(\theta,x)$, and the identity principle extends this equality to all $(\theta,x)$. Hence the pullback $F\mapsto F\circ T$, where $(F\circ T)(\theta,x):=F(T(\theta),x)$, maps the finite set $\mathfrak P$ onto a set that contains $\mathfrak P$. It is therefore a permutation of $\mathfrak P$. Let $N\ge1$ be the order of this permutation, so that $F_\omega(T^N(\theta),x)=F_\omega(\theta,x)$ for every realized $\omega$.

\emph{(1b) $T^N$ fixes every path product.}
To isolate the individual path products in the coefficients of $F_\omega$, we show that the slope products $\omega_\pi$ are linearly independent. Let $\Pi$ be $\Pi(a)$ or $\Pi_{\rm b}$, and suppose that numbers $c_\pi$, not all zero, satisfy $\sum_{\pi\in\Pi}c_\pi\omega_\pi=0$ for every realized pattern $\omega$. Choose $\pi_0$ with $c_{\pi_0}\ne0$ such that the set $\mathcal U$ of neurons on $\pi_0$ is maximal with respect to inclusion. This set contains at most one neuron of each layer.

We first construct a point $(\theta_0,x_0)$ at which the neurons in $\mathcal U$ have zero preactivation and all other neurons have nonzero preactivation. Let $x_0$ be an interior point of $\mathcal X$ with pairwise distinct nonzero coordinates, and choose the layers in increasing order. In each layer, one channel without a neuron in $\mathcal U$ receives a generic positive row and a large bias, so that its neurons are active and its outputs at distinct positions differ. By the padding convention of Section~\ref{sec:settings}, any two positions of a layer share an offset, so the patches that they read at $x_0$ differ. If $\mathcal U$ contains a neuron $(l,p,j)$, then channel $j$ of layer $l$ receives a generic row, for which the preactivations at distinct positions differ, and the bias for which the preactivation at $p$ vanishes. All other channels receive zero rows and nonzero biases.

At $(\theta_0,x_0)$, the preactivation of each neuron in $\mathcal U$ depends on the bias of its own channel with coefficient $1$, and it does not depend on the biases of later layers. Varying these biases near $\theta_0$ therefore realizes all $2^{|\mathcal U|}$ sign patterns of the neurons in $\mathcal U$, while all other neurons keep their signs. The alternating sum of the relation $\sum_{\pi\in\Pi}c_\pi\omega_\pi=0$ over these $2^{|\mathcal U|}$ patterns removes every path that misses a neuron of $\mathcal U$. Distinct paths in $\Pi$ pass through different sets of neurons, so by the maximality of $\mathcal U$ only $\pi_0$ remains. Its contribution is $c_{\pi_0}$ times the product of the factors $1-\alpha_l$ over the layers $l$ of the neurons in $\mathcal U$, which is nonzero. This contradiction proves the linear independence.

We now compare the coefficients of $x_a$ and the constant terms in $F_\omega(T^N(\theta),x)=F_\omega(\theta,x)$ for all realized $\omega$. By the linear independence,
\begin{equation}\label{complete:eq:fixed-products}
 w_\pi(T^N(\theta))=w_\pi(\theta)\quad\text{for every path }\pi.
\end{equation}
Moreover, let $\mathcal{V}_a:=\mathrm{span}\{w_\pi:\pi\in\Pi(a)\}$ and $\mathcal{V}_b:=\mathrm{span}\{w_\pi:\pi\in\Pi_{\rm b}\}$. By the linear independence, the coefficients of $x_a$ in the polynomials of $\mathfrak{P}$ span $\mathcal V_a$, and their constant terms span $\mathcal V_b$. Since the pullback by $T$ permutes $\mathfrak{P}$, we obtain
\begin{equation}\label{complete:eq:pullback}
 \phi\longmapsto\phi\circ T\quad\text{is a linear bijection of }\mathcal{V}_a
 \text{ and of }\mathcal{V}_b .
\end{equation}
In particular, $w_\pi(T(w,b))$ is a polynomial in $w$ alone for every $\pi\in\Pi(a)$, and it is a polynomial in $(w,b)$ that is linear in $b$ for every $\pi\in\Pi_{\rm b}$.

\emph{(1c) Maps that fix all path products are positive scalings.}
Let $T'$ be a map with components in $\mathcal C^{\rm qa}$ such that $f(x;T'(\theta))=f(x;\theta)$ and $w_\pi(T'(\theta))=w_\pi(\theta)$ for all $\theta$, all $x\in\mathcal X$, and all paths $\pi$. We show that $T'(\theta)=D(\theta)\cdot\theta$, where $D(\theta)$ consists of positive diagonal matrices as in Proposition~\ref{prop:scaling-permutation} with entries in $\mathcal C^{\rm qa}$. By~\eqref{complete:eq:fixed-products}, this applies to $T'=T^N$. Write $\theta'=(w',b')=T'(\theta)$. The path of $\beta$ gives $\beta'=\beta$.

We first define the scaling factors. A \emph{tail} of a channel is a path $\gamma$ from a neuron of this channel to the readout, and $w_\gamma(w)$ denotes the product of the weights along $\gamma$. Let $\gamma$ be a tail of a channel $(l,j)$. The path that starts at the bias $b_{l,j}$ and follows $\gamma$ gives $b_{l,j}'\,w_\gamma(w')=b_{l,j}\,w_\gamma(w)$. We extend $\gamma$ backward to a path $\pi$ from an input coordinate. The product $w_\pi(w')=w_\pi(w)$ is nonzero for generic $w$, so $w_\gamma(w')\ne0$ on a dense subset of the hyperplane $\{b_{l,j}=0\}$. Hence $b_{l,j}'$ vanishes on this hyperplane. Division by the coordinate $b_{l,j}$ gives a function $\mu_{l,j}:=b_{l,j}'/b_{l,j}\in\mathcal C^{\rm qa}$, and by continuity
\begin{equation}\label{complete:eq:tails}
 \mu_{l,j}\,w_\gamma(w')=w_\gamma(w)\qquad\text{for every tail $\gamma$ of the channel }(l,j).
\end{equation}

Next, let $e$ be a weight from a channel $(l-1,i)$ to a neuron of a channel $(l,j)$, and let $\gamma$ be a tail of this neuron. Then $e$ followed by $\gamma$ is a tail of $(l-1,i)$, and \eqref{complete:eq:tails} gives $\mu_{l-1,i}\,\theta_e'\,w_\gamma(w')=\theta_e\,w_\gamma(w)=\mu_{l,j}\,\theta_e\,w_\gamma(w')$. Since $w_\gamma(w')\ne0$ for generic $\theta$, we obtain $\mu_{l-1,i}\,\theta_e'=\mu_{l,j}\,\theta_e$, where $\mu_{0,i}:=1$ for the input channels. In the same way, a shortcut followed by a tail gives $\mu_{l-2,j}=\mu_{l,j}$ for every outer layer $l\in\mathcal R$, and the readout gives $\mu_{\widetilde L,j}\,a_j'=a_j$. Together with $b_{l,j}'=\mu_{l,j}b_{l,j}$, these relations state that $\theta'=D(\theta)\cdot\theta$ with the diagonal entries $\mu_{l,j}$, provided that all $\mu_{l,j}$ are positive.

We now show that $\mu_{l,j}$ has no zero if $l<\widetilde L$. Every layer has at least two channels, so the channel $(l,j)$ has two tails $\gamma_1$ and $\gamma_2$ without a common weight. Suppose that $\mu_{l,j}(\theta_0)=0$. By~\eqref{complete:eq:tails}, the Taylor series of $\mu_{l,j}$ at $\theta_0$ has no constant term and divides the Taylor series of both $w_{\gamma_1}$ and $w_{\gamma_2}$. Up to invertible factors, the Taylor series of $w_{\gamma_r}$ at $\theta_0$ is the product of the variables $\theta_e-\theta_{0,e}$ over the weights $e$ of $\gamma_r$ with $\theta_{0,e}=0$. These two products have no common variable, so they have no common divisor without constant term in the factorial ring of formal power series. This contradiction shows that $\mu_{l,j}$ has no zero and hence a constant sign.

It remains to determine the signs. Let $\mathcal O$ be the nonempty open set of points $(\theta,x)$, with $x$ in the interior of $\mathcal X$, at which all preactivations are positive. As in (1a), every nonempty open subset of $\mathcal O$ contains a nonempty open set $O$ and a realized pattern $\omega'$ such that $f(x;\theta')=F_{\omega'}(\theta',x)$ and every neuron $v$ with $\omega'_v=1$ has a nonnegative preactivation at $(\theta',x)$ for all $(\theta,x)\in O$. Since $T'$ fixes the path products and $f=F_\omega$ with $\omega\equiv1$ on $\mathcal O$, the coefficients of $x_a$ in $f(x;\theta')=f(x;\theta)$ give $\sum_{\pi\in\Pi(a)}(1-\omega'_\pi)w_\pi(\theta)=0$ on $O$. The path products of distinct paths in $\Pi(a)$ are distinct monomials, so $\omega'_\pi=1$ for all these paths. Every neuron lies on such a path, so $\omega'\equiv1$. Thus all preactivations $z'_v$ of the network with parameter $\theta'$ are nonnegative on $O$. Let $D_l:=\operatorname{diag}(\mu_{l,j})$ act on the channels of layer $l$. If $h_{l-1}'=D_{l-1}h_{l-1}$, then the relations above give $z_l'=D_lz_l$. Since $z_l>0$ and $z_l'\ge0$ on $O$, we get $\mu_{l,j}\ge0$, and positive homogeneity gives $h_l'=D_lh_l$. By induction over the layers, $\mu_{l,j}\ge0$ on $O$ for all channels. For $l<\widetilde L$, this shows that $\mu_{l,j}>0$ everywhere. For $l=\widetilde L$, the relation $\mu_{\widetilde L,j}\,a_j'=a_j$ shows that $\mu_{\widetilde L,j}$ has no zero on the two connected sets $\{a_j>0\}$ and $\{a_j<0\}$. Both sets contain parameters of points of $\mathcal O$, so $\mu_{\widetilde L,j}$ is positive on both, and $\mu_{\widetilde L,j}\ge0$ everywhere. If $\mu_{\widetilde L,j}(\theta_0)=0$, then $\theta_0$ is a minimum of $\mu_{\widetilde L,j}$, and differentiating $\mu_{\widetilde L,j}\,a_j'=a_j$ with respect to $a_j$ at $\theta_0$ gives $0=1$. Hence all $\mu_{l,j}$ are positive, and $T'$ is a positive scaling.

\medskip
\noindent\textbf{Step 2 (Reduction to a map that is linear in the biases).}
We show that $T$ differs by a positive scaling from the map
\[
 \widehat{T}(w,b):=\bigl(\widehat{w}(w),\,\mathcal{M}(w)b\bigr),\qquad
 \widehat{w}(w):=T_w(w,0),\quad \mathcal{M}(w):=D_b T_b(w,0),
\]
where $T=(T_w,T_b)$ denotes the splitting of $T$ into weights and biases. More precisely, $T(\theta)$ is a positive scaling of $\widehat T(\theta)$ for generic $\theta$, the map $\widehat T$ satisfies $f(x;\widehat{T}(\theta))=f(x;\theta)$ for all $\theta$, and $\det\mathcal M(w)\ne0$ for every $w$.

We first construct the scaling. By~\eqref{complete:eq:pullback}, $w_\pi(T(w,b))$ does not depend on $b$ for every path $\pi$ that starts at an input coordinate, so $w_\pi(T_w(w,b))=w_\pi(\widehat w(w))$. For generic $w$, all entries of $\widehat w(w)$ are nonzero. Indeed, every weight lies on such a path $\pi$, and $w_\pi\circ T$ does not vanish identically by~\eqref{complete:eq:pullback}. Fix such a $w$, and for every weight $e$ let $\rho_e$ be the ratio of the entries of $T_w(w,b)$ and $\widehat w(w)$ at $e$. The product of the ratios $\rho_e$ along every path from an input coordinate to the readout equals $1$. For a channel $(l,j)$, let $\lambda_{l,j}(w,b)$ be the product of the ratios $\rho_e$ along a path from an input coordinate to a neuron of this channel. This product does not depend on the path, because the path continues to the readout along a tail and the product over the complete path is $1$. By induction over the layers, it does not depend on the position of the neuron either, because any two positions of a layer share an offset and therefore read the previous layer through a common weight. Hence $\rho_e=\lambda_{l,j}/\lambda_{l-1,i}$ for every weight $e$ from a channel $(l-1,i)$ to a channel $(l,j)$, with $\lambda_{0,i}:=1$, the ratio at $a_j$ is $\lambda_{\widetilde L,j}^{-1}$, and $\lambda_{l,j}=\lambda_{l-2,j}$ for $l\in\mathcal R$. Thus $T_w(w,b)$ is the scaling of $\widehat w(w)$ with the factors $\lambda_{l,j}(w,b)$. These factors are continuous and nonzero in $b$ and equal $1$ at $b=0$, so they are positive.

Now let $\gamma$ be a tail of a channel $(l,j)$, so that $w_\gamma(T_w(w,b))=\lambda_{l,j}(w,b)^{-1}w_\gamma(\widehat w(w))$. By~\eqref{complete:eq:pullback}, the product $T_{b_{l,j}}(w,b)\,w_\gamma(T_w(w,b))$ of the path that starts at $b_{l,j}$ and follows $\gamma$ is linear in $b$. Since $w_\gamma(\widehat w(w))\ne0$, the function $\lambda_{l,j}(w,b)^{-1}T_{b_{l,j}}(w,b)$ is linear in $b$, and so is $T_\beta(w,b)$. Differentiation at $b=0$, where $\lambda_{l,j}=1$, identifies these linear functions with the entries of $\mathcal M(w)b$. Therefore $T(w,b)$ is the scaling of $\widehat T(w,b)$ with the positive factors $\lambda_{l,j}(w,b)$, for generic $w$ and all $b$. By Proposition~\ref{prop:scaling-permutation}, $f(x;\widehat T(\theta))=f(x;T(\theta))=f(x;\theta)$ for these $\theta$, and by continuity for all $\theta$.

To see that $\mathcal M(w)$ is invertible, we apply Step~1 to $\widehat T$. Some power $\widehat T^N$ is a positive scaling with factors $\mu_{l,j}(w,b)>0$. Its bias component is $\mathcal{M}(w_{N-1})\cdots\mathcal{M}(w_0)b$, where $w_0=w$ and $w_{n+1}=\widehat w(w_n)$, and it multiplies each bias $b_{l,j}$ by $\mu_{l,j}(w,b)$. Differentiating at $b=0$ shows that $\mathcal{M}(w_{N-1})\cdots\mathcal{M}(w_0)$ is a diagonal matrix with positive entries. Hence $\det\mathcal{M}(w)\ne 0$ for every $w$.

Steps~3--5 use the identity $f(x;\widehat T(\theta))=f(x;\theta)$ at parameters for which $\theta$ and $\widehat T(\theta)$ both satisfy finitely many conditions of the following type: a nonzero polynomial $q$ in the path products does not vanish. By Step~1, the pullback by $\widehat T^N$ fixes every polynomial in the path products, so $q\circ\widehat T$ does not vanish identically. Hence $\theta$ and $\widehat T(\theta)$ satisfy these conditions simultaneously for generic $\theta$. We assume this throughout Steps~3--5 without further mention.

\medskip
\noindent\textbf{Step 3 (The bias map is block lower triangular).}
We order the biases by layers, with $\beta$ last, and write $\mathbf e_{s,j}$ for the coordinate direction of the bias $b_{s,j}$. We show that, for every layer $s$, the vectors $\mathcal M(w)\mathbf e_{s,j}$ have no component in the layers before $s$, and that the diagonal block of layer $s$ is
\[
 \mathcal{M}_{s,s}(w)=P_s\operatorname{diag}\bigl(\tau_{s,1}(w),\dots,\tau_{s,c_s}(w)\bigr)
\]
with a permutation matrix $P_s$ that does not depend on $w$ and functions $\tau_{s,j}$ without zeros.

We first consider $\beta$. Adding a constant to $\beta$ adds this constant to $f$ and does not change the inputs at which $f(\cdot;\theta)$ is not differentiable. The same must hold for $\widehat T(\theta)$, so $\mathcal M(w)$ maps the direction of $\beta$ to itself with the factor $1$.

The proof for the layers is a downward induction on $s=\widetilde L,\dots,1$. Suppose that the claim holds for all layers after $s$. We let the biases of the layers after $s$ tend to $+\infty$ at separated scales $b_{s+1}\ll b_{s+2}\ll\cdots\ll b_{\widetilde L}$. Then all neurons of these layers are active for the parameter $\theta$. By the induction hypothesis, the diagonal blocks of $\mathcal M(w)$ for these layers are permutation matrices with nonzero factors. The corresponding biases of $\widehat T(\theta)$ therefore diverge as well, and every neuron of a layer $l>s$ has a constant slope, $1$ or $\alpha_l$, for the parameter $\widehat T(\theta)$. For both parameters, the layers after $s$ thus act as fixed affine maps. Up to terms that depend only on the layers before $s$ or that are linear in the biases of the layers $s,\ldots,\widetilde L$, the function $f(x;\theta)$ then equals
\begin{equation}\label{complete:eq:cut}
 \sum_{p\in\Omega_s}\sum_{j=1}^{c_s}\mathsf{r}_{s,p,j}(w)\,h_{s,j}(p),
\end{equation}
where $\mathsf{r}_{s,p,j}(w)=\partial f/\partial h_{s,j}(p)$ is a polynomial in the weights of the layers after $s$.

We now examine \eqref{complete:eq:cut} along lines $b_0+t\,d$ in the space of the biases, where the direction $d$ is supported on the layers $1,\ldots,s$. First, let $d=\mathbf{e}_{s,j}$. Then the layers before $s$ do not change along the line, and only the $|\Omega_s|$ neurons $(s,p,j)$, $p\in\Omega_s$, change sign. Hence the derivative of $f$ along the line jumps only by linear combinations of the polynomials $\mathsf{r}_{s,p,j}(w)$, $p\in\Omega_s$, of the single channel $j$. Next, let $d$ have a nonzero entry $d_{s',i}$ in a layer $s'<s$, and let $s'$ be the first such layer. Consider a simple zero $(x,b_0)$ of a neuron $(s',q,i)$. Moving the bias $b_{0,s,j}$ across the threshold of a neuron $(s,p,j)$ changes the slope of this neuron by $1-\alpha_s$, and it changes the jump of the derivative of $f$ across $\{z_{(s',q,i)}=0\}$ by a nonzero multiple of $\mathsf{r}_{s,p,j}(w)$. This holds for every channel $j=1,\dots,c_s$. Thus every direction with a nonzero entry before layer $s$ produces jumps that involve all $c_s\ge 2$ channels of layer $s$. In contrast, a direction $d=\sum_{j}d_{s,j}\mathbf{e}_{s,j}$ within layer $s$ produces $|\Omega_s|\cdot\#\{j:d_{s,j}\ne 0\}$ jumps, and this number attains its minimum $|\Omega_s|$ exactly on the coordinate axes $\R\mathbf{e}_{s,j}$. Since $f(x;\widehat T(\theta))=f(x;\theta)$, the matrix $\mathcal{M}(w)$ respects these properties. Modulo the biases of the layers after $s$ and $\beta$, it therefore maps $\mathrm{span}\{\mathbf{e}_{s,j}\}_{j=1}^{c_s}$ to itself and permutes the axes $\R\mathbf{e}_{s,j}$. This proves the claim for layer $s$. The factors $\tau_{s,j}(w)$ have no zeros because $\det\mathcal{M}(w)\ne 0$. By continuity, the permutation $P_s$ and the signs $\epsilon_{s,j}:=\operatorname{sign}\tau_{s,j}(w)\in\{\pm1\}$ do not depend on $w$.

\medskip
\noindent\textbf{Step 4 (The kernel maps are linear).}
We now determine the weight map $\widehat{w}(w)$. Write $\mathcal{M}_{s,s}(w)=P_s D_s(w) E_s$ with $D_s(w)=\operatorname{diag}(|\tau_{s,j}(w)|)$ and $E_s=\operatorname{diag}(\epsilon_{s,j})$. We apply the inverse of the channel permutation and positive scaling $P_sD_s(w)$ to the channels of every layer $s$ of $\widehat T(\theta)$, and we adjust the readout accordingly. This yields a normalized parameter $\widetilde\theta=\widetilde\theta(w,b)$ that realizes the same function. Its biases satisfy
\begin{equation}\label{complete:eq:normalized-bias}
 \widetilde{b}_{s,i}=\epsilon_{s,i}b_{s,i} + (\text{a linear function of the biases of the layers before }s),
 \qquad \epsilon_{s,i}\in\{-1,1\},
\end{equation}
and for every outer layer $s\in\mathcal{R}$ its shortcut multiplies $\widetilde h_{s-2}$ by the matrix $\widetilde{I}_s(w)=(P_s D_s(w))^{-1}P_{s-2}D_{s-2}(w)$, which is a permutation matrix with positive factors.

We show by downward induction on $l=\widetilde L,\dots,1$ that each kernel $\widetilde{K}_l$ of $\widetilde\theta$ depends linearly on $K_l$ alone. Step~5 then gives invertibility of the kernel map. Let $\widetilde{K}_l(u)_{j,i}$ be the entry that connects the channel $(l-1,i)$ to a neuron $v$ of the channel $(l,j)$, and let $\gamma$ be a tail of $v$ that traverses every later kernel. For $l>1$, take the coefficient of $b_{l-1,i}$ in the path product $\widetilde b_{l-1,i}\,\widetilde K_l(u)_{j,i}\,w_\gamma(\widetilde w)$ using \eqref{complete:eq:normalized-bias} and \eqref{complete:eq:pullback}. For $l=1$, use $\mathcal V_a$ for paths from the input coordinate $x_a$. In either case, $\widetilde K_l(u)_{j,i}\,w_\gamma(\widetilde w)$ is a linear combination of weight products from the same source, each of degree at most one in each kernel from layer $l$ onward and in the readout weights. At the final bias $b_{\widetilde L,j}$, the same comparison gives $\widetilde a_j$ as a nonzero multiple of $a_j$. By the induction hypothesis, $w_{\gamma}(\widetilde{w})$ is a product of nonzero linear forms, one in the kernel of each layer after $l$ and one in the readout weights. Since $\widetilde{K}_l(u)_{j,i}$ is continuous, the polynomial vanishes on the zero set of each of these forms and is therefore divisible by it. The forms involve disjoint sets of variables, so the polynomial is divisible by their product. The quotient is therefore affine in $K_l$ alone.

Unless $l$ is a residual inner layer, every path from the source uses $K_l$ once. For an inner layer, a path either uses both $K_l$ and $K_{l+1}$ once or bypasses both. Setting $K_{l+1}=0$ in the polynomial identity makes the chosen tail vanish by the induction hypothesis. The sum of the shortcut terms is then zero and, being independent of $K_{l+1}$, vanishes identically. This proves homogeneity of degree one in $K_l$. The source and the first factor of the tail fix the input and output channels $i$ and $j$, respectively. Hence
\begin{equation}\label{complete:eq:kernel-map}
 \widetilde{K}_l(u)_{j,i}=\sum_{u'\in\Lambda_l}C_{l,ji}(u,u')\,K_l(u')_{j,i},
 \qquad C_{l,ji}(u,u')\in\R.
\end{equation}
The same division along a shortcut shows that each matrix $\widetilde{I}_s$ is constant, and that $\widetilde{z}_s$ is an affine function of $(K_s,b_s)$ when the weights and biases of the layers before $s$ are fixed.

\medskip
\noindent\textbf{Step 5 (The normalized parameter equals $\theta$).}
We now compare the zero sets of the neurons of $\theta$ and $\widetilde\theta$ layer by layer and show that $\widetilde{\theta}=\theta$.

We prove $\widetilde{h}_s=h_s$, $\widetilde{A}_s=A_s$, and $\widetilde{b}_s=b_s$ by forward induction on $s=1,\dots,\widetilde L$. Suppose that $\widetilde{h}_{s'}=h_{s'}$ for all $s'<s$. As in Step~3, we let the biases of the layers after $s$ tend to $+\infty$, so that $f$ reduces to~\eqref{complete:eq:cut}, and we choose the biases of the layers before $s$ so large that all neurons before layer $s$ are active on an open ball of inputs $x$. On this ball, the coordinates of $h_{s-1}(x)$ are linearly independent polynomials in $(x,w)$. This follows by induction over the layers. The coordinate $(p,j)$ of $h_{s-1}$ is the sum of the bias $b_{s-1,j}$, of the products of the kernel entries $K_{s-1}(u)_{jk}$ with the coordinates $h_{s-2,k}(p+u)$, and of the shortcut term $h_{s-3,j}(p)$ if $s-1\in\mathcal R$. Distinct coordinates do not share such a product, and every coordinate contains at least one, because every position reads the input under the padding convention of Section~\ref{sec:settings}.

First, let $s$ be the layer of a fully connected or convolutional block. Only channel $i$ of $\widetilde{z}_s$ contains $b_{s,i}$, with the coefficient $\epsilon_{s,i}$ by \eqref{complete:eq:normalized-bias}, and each position $p\in\Omega_s$ reads the translated patch $p+\Lambda_s$ through \eqref{complete:eq:kernel-map}. The functions~\eqref{complete:eq:cut} of $\theta$ and $\widetilde{\theta}$ have the same zero hyperplanes $\{z_{s,v}=0\}$ and the same gradient jumps across them. This pairs each neuron $(s,p,i)$ of $\widetilde\theta$ with a neuron $(s,\psi(p),i)$ of $\theta$ for a permutation $\psi$ of $\Omega_s$, and the coefficients of $b_{s,i}$ give $\widetilde z_{s,(p,i)}=\epsilon_{s,i}z_{s,(\psi(p),i)}$. Comparing the coefficients of the linearly independent products $K_s(u')_{ik}h_{s-1,k}(q)$ in \eqref{complete:eq:kernel-map} gives $C_{s,ik}(u,u')=\epsilon_{s,i}\mathbf 1_{\{u-u'=\psi(p)-p\}}$ for every offset $u$ used at $p$. Two positions that use a common offset therefore have the same shift $\psi(p)-p$. Under the padding convention of Section~\ref{sec:settings}, all positions use the offset $0$ (same padding) or every offset (valid padding), so the shift is constant. It vanishes because $\psi$ maps the finite set $\Omega_s$ onto itself. Since every offset is used at some position, $C_{s,ik}(u,u')=\epsilon_{s,i}\mathbf 1_{\{u=u'\}}$. Thus each neuron $(s,p,i)$ is paired with itself, and $\widetilde{z}_s(x) = E_s z_s(x)$, that is, $\widetilde{A}_s = E_s A_s$ and $\widetilde{b}_s = E_s b_s$. For the channels with $\epsilon_{s,i}=-1$, we substitute $\sigma(-z)=\sigma(z)-(1+\alpha)z$. This cancels all terms $\sigma(z_{s,v})$ in \eqref{complete:eq:cut} and leaves $(1+\alpha)\sum_{\epsilon_v=-1}\mathsf{r}_{s,v}z_{s,v}(x)=0$, which forces $E_s=I$. Hence $\widetilde{A}_s=A_s$, $\widetilde{b}_s=b_s$, and $\widetilde{h}_s=h_s$.

Next, consider a residual block with inner layer $s-1$ and outer layer $s\in\mathcal{R}$. We treat both layers in one induction step and assume $\widetilde{h}_{s-2}=h_{s-2}$. Each of the two layers reads the preceding layer as above, and the shortcut only adds $h_{s-2}$ to the output of the outer layer: $h_s=\sigma(z_s)+h_{s-2}$ and $\widetilde h_s=\sigma(\widetilde z_s)+\widetilde I_sh_{s-2}$. The argument above therefore applies to both layers, except that~\eqref{complete:eq:cut} now contains shortcut terms. These terms are linear in $h_{s-2}$ and do not involve the kernels $K_{s-1}$ and $K_s$. At the inner layer, we let $b_s$ tend to $+\infty$ as well, so that the outer layer is linear. The comparison of the zero hyperplanes then gives $\widetilde z_{s-1}=E_{s-1}z_{s-1}$. After the substitution $\sigma(-z)=\sigma(z)-(1+\alpha)z$, the remainder $(1+\alpha)\sum_{\epsilon_v=-1}\mathsf r_{s-1,v}z_{s-1,v}$ is the only term that involves $K_{s-1}$, so it vanishes. This forces $E_{s-1}=I$, $\widetilde A_{s-1}=A_{s-1}$, $\widetilde b_{s-1}=b_{s-1}$, and $\widetilde h_{s-1}=h_{s-1}$. At the outer layer, with $b_s$ finite, the same argument gives $\widetilde z_s=E_sz_s$ with the same coefficients $\mathsf r_{s,p,j}$. The remainder $(1+\alpha)\sum_{\epsilon_v=-1}\mathsf r_{s,v}z_{s,v}$ is the only term that involves $K_s$, so $E_s=I$, $\widetilde A_s=A_s$, and $\widetilde b_s=b_s$. The remaining shortcut terms give $\sum_{p,j}\mathsf r_{s,p,j}(w)\bigl[(I-\widetilde I_s)h_{s-2}\bigr]_j(p)=\mathrm{const}$. The coordinates of $h_{s-2}$ are linearly independent, and so are the polynomials $\mathsf r_{s,p,1},\ldots,\mathsf r_{s,p,c_s}$, because each of them involves weights that read only its own channel. Hence $\widetilde I_s=I$, that is, $(P_s,D_s(w))=(P_{s-2},D_{s-2}(w))$ and $\widetilde h_s=h_s$.

This completes the induction. Since $f(\cdot;\widetilde\theta)=f(\cdot;\theta)$ and $\widetilde h_{\widetilde L}=h_{\widetilde L}$, the readouts of $\widetilde\theta$ and $\theta$ agree as well, because the pooled outputs of distinct channels and the constant $1$ are linearly independent. Thus $\widetilde{\theta}=\theta$.

\medskip
\noindent\textbf{Conclusion.}
By the definition of $\widetilde\theta$ in Step~4, the identity $\widetilde\theta=\theta$ states that $\widehat T(\theta)$ is a positive scaling of $P\cdot\theta$, where $P=(P_s)$ is the channel permutation of Step~3, which satisfies the coupling of Proposition~\ref{prop:scaling-permutation} because $\widetilde I_s=I$. By Step~2, the same holds for $T(\theta)$ for generic $\theta$. Positive scalings do not change path products, so the map $\theta\mapsto P^{-1}\cdot T(\theta)$ fixes every path product for generic $\theta$, and hence for all $\theta$ by continuity. This map has components in $\mathcal C^{\rm qa}$ and preserves $f$. By (1c), $P^{-1}\cdot T(\theta)=D(\theta)\cdot\theta$ with positive diagonal matrices $D(\theta)$ with entries in $\mathcal C^{\rm qa}$. Therefore $T(\theta)=PD(\theta)\cdot\theta$ for every $\theta\in\R^S$.
\end{proof}

\section{The Upper Bounds}\label{app:main}

\begin{proof}[Proof of Theorem~\ref{thm:covering-upper-bound} and Corollary~\ref{thm:covering-number-sp-deep}]
The proof extends the approach of \citet[Appendix~A.3]{shen2024complexity} from channel permutation to positive scaling and has three steps. Step~1 uses positive scaling to reduce the parameter cube $\Theta=[-B,B]^S$ to a subset $\Theta_0$ that consists of faces of dimension at most $S-H$ and still realizes every function in $\mathcal F$. Step~2 bounds the Lipschitz constant $L_B$ of the map $\theta\mapsto f(\cdot;\theta)$ on $\Theta$. Step~3 covers $\Theta_0$ by a grid that is invariant under channel permutations and counts the grid points up to permutation.

\medskip
\noindent\textbf{Step 1 (Effective parameter space).} By Proposition~\ref{prop:scaling-permutation}, a positive scaling assigns a factor to each scalable channel, and each residual output channel inherits the factor of the corresponding channel of its block input. We therefore group each scalable channel and all residual output channels that inherit its factor into one \emph{node} $v\in\{1,\dots,H\}$ with factor $\lambda_v>0$. The node $0$, with $\lambda_0=1$, consists of the input channels, the residual output channels that inherit their factor, and the readout. A parameter $\theta_e$ goes from node $u$ to node $v$ if it multiplies an output of $u$ and feeds a channel of $v$, and a bias goes from node $0$ to the node of its channel. The scaling acts on each parameter $\theta_e$ from node $u$ to node $v$ as
\begin{equation}\label{eq:scale}
 \theta'_e = \frac{\lambda_v}{\lambda_u}\,\theta_e .
\end{equation}
Every parameter is an incoming parameter of exactly one node. Let $k_v$ be the number of incoming parameters of node $v$. Then $k_v\ge2$ for $v\ge1$, and $k_0\ge1$ because node $0$ receives the readout parameters. Hence $S-H=k_0+\sum_{v=1}^H(k_v-1)$.

We now choose the factors for a given $\theta\in\Theta$. If a node $v\ge1$ cannot be reached from node $0$ along nonzero parameters, the outputs of its channels vanish identically, and we set its incoming and outgoing parameters to $0$ without changing the network function. For every other node $v$, we set
\[
 \log\lambda_v := \min_{\pi:\,0\to v}\ \sum_{e\in\pi}\log\frac{B}{|\theta_e|} \ge 0,
\]
where the minimum is taken over the paths of nonzero parameters from node $0$ to $v$. All terms $\log(B/|\theta_e|)$ are nonnegative. Therefore $\log\lambda_v\le\log\lambda_u+\log(B/|\theta_e|)$ for every nonzero parameter $\theta_e$ from $u$ to $v$, with equality on the last parameter of a minimizing path. By~\eqref{eq:scale}, every rescaled parameter satisfies $|\theta'_e|\le B$, and every node $v\ge1$ that is reached has an incoming parameter with $|\theta'_e|=B$.

Let $\Theta_0\subset\Theta$ be the set of parameters for which the incoming parameters of every node $v\ge1$ are all zero or include one that equals $\pm B$. By the construction above, every function in $\mathcal F$ is realized by a parameter in $\Theta_0$, and $\Theta_0$ is the union of at most $\prod_{v=1}^H(1+2k_v)$ faces of $\Theta$ of dimension at most $S-H$. Moreover, by Proposition~\ref{prop:scaling-permutation}, the channel permutations form a group $\Gamma$ of order $G$ that preserves the network function. Each $P\in\Gamma$ permutes the nodes within each scalable layer and maps the incoming parameters of each node onto those of its image. In particular, $\Gamma$ maps $\Theta_0$ onto itself.

\medskip
\noindent\textbf{Step 2 (Lipschitz constant).} We bound the change of $f(\cdot;\theta)$ under a change of $\theta\in\Theta$. We use the $\widetilde{L}=L+L_{\mathrm{res}}$ layers of~\eqref{eq:preactivation}. For each layer $j$, let $n_j$ be the number of weights that feed a single neuron of layer $j$, so that $\prod_{j=1}^{\widetilde{L}}n_j=\prod_{l=1}^L\kappa_l$ with $\kappa_l$ as in Theorem~\ref{thm:covering-upper-bound}. Since $|\sigma_j(t)|\le |t|$ and $\sigma_j$ is $1$-Lipschitz, induction on $j$ gives $\|h_j(\theta)\|_\infty\le M_j$ and $\|h_j(\theta)-h_j(\eta)\|_\infty\le E_j\|\theta-\eta\|_\infty$ for all $\theta,\eta\in\Theta$, where $M_0=R_{\mathcal{X}}$, $E_0=0$, and
\begin{align*}
 M_j&:=B(n_j M_{j-1}+1)+\mathbf{1}_{\{j\in\mathcal{R}\}}M_{j-2},\\
 E_j&:=(n_j M_{j-1}+1)+B n_j E_{j-1}+\mathbf{1}_{\{j\in\mathcal{R}\}}E_{j-2}.
\end{align*}
The three terms of $E_j$ bound the effect of changing the parameters of layer $j$, the input $h_{j-1}$ of layer $j$, and the shortcut input $h_{j-2}$. The same argument at the readout gives
\[
 \|f(\cdot;\theta)-f(\cdot;\eta)\|_\infty
 \le (1+c_L M_{\widetilde{L}}+B c_L E_{\widetilde{L}})\|\theta-\eta\|_\infty
 =: L_B\|\theta-\eta\|_\infty.
\]
For $B\ge 1$, we have $M_j\le 2B n_j M_{j-1}$, and induction on $j$ yields
\begin{gather*}
 M_{\widetilde{L}} \le R_{\mathcal{X}}(2B)^{\widetilde{L}}\prod_{l=1}^L\kappa_l,\qquad
 E_{\widetilde{L}} \le 2\widetilde{L} R_{\mathcal{X}}(2B)^{\widetilde{L}-1}\prod_{l=1}^L\kappa_l,\\
 L_B \le 2(\widetilde L+1)R_{\mathcal{X}}(2B)^{\widetilde L+1}c_L\prod_{l=1}^L\kappa_l.
\end{gather*}
Thus the constant of Theorem~\ref{thm:covering-upper-bound} satisfies $C_+^{1/(S-H)}\ge12BL_B$.

\medskip
\noindent\textbf{Step 3 (Covering up to permutation).} By Step~2, every $(\varepsilon/L_B)$-net of $\Theta_0$ in the $\ell_\infty$-norm yields an $\varepsilon$-cover of $\mathcal{F}$. We construct a net that $\Gamma$ maps onto itself, so that one center for each orbit of $\Gamma$ suffices. Fix $0<\varepsilon\le2B$, let $N:=\lceil BL_B/\varepsilon\rceil\le2BL_B/\varepsilon$, and let $\mathcal T\subset[-B,B]$ be a set of $N$ points within distance $\varepsilon/L_B$ of every point of $[-B,B]$. For $k\ge2$, let $Z_k\subset[-B,B]^k$ consist of $0$ and of the vectors with one coordinate equal to $\pm B$ and all other coordinates in $\mathcal T$, so that $|Z_k|\le1+2kN^{k-1}$. Let $\mathcal Z\subset\Theta$ be the set of parameters whose incoming parameters at each node $v\ge1$ form a vector in $Z_{k_v}$ and whose other $k_0$ coordinates lie in $\mathcal T$. For $\theta\in\Theta_0$, the incoming parameters of each node $v\ge1$ are all zero or include one that equals $\pm B$. Keeping this parameter and rounding all others to $\mathcal T$ gives a point of $\mathcal Z$ within $\ell_\infty$-distance $\varepsilon/L_B$ of $\theta$. Hence $\mathcal Z$ is an $(\varepsilon/L_B)$-net of $\Theta_0$ with $|\mathcal Z|\le\prod_{v=1}^H(1+2k_v)N^{S-H}$.

The group $\Gamma$ maps $\mathcal Z$ onto itself and preserves the network function. Therefore $\mathcal N(\mathcal F,\varepsilon,\|\cdot\|_\infty)$ is at most the number of orbits of $\Gamma$ in $\mathcal Z$. By Burnside's lemma, this number equals $G^{-1}$ times the sum over $P\in\Gamma$ of the number of points of $\mathcal Z$ that $P$ fixes. Consider a scalable layer with $q$ nodes, each with $k$ incoming parameters. If $P$ fixes a point of $\mathcal Z$, then along each cycle of $P$ on this layer the incoming parameters of one node determine those of the other nodes. Hence the layer contributes at most one factor $|Z_k|$ for each cycle, and node $0$ contributes at most $N^{k_0}$. Since $\Gamma$ is the product of the symmetric groups of the scalable layers, the sum over $\Gamma$ factorizes. The sum of $|Z_k|^{\#\text{cycles}}$ over the $q!$ permutations of one layer equals $|Z_k|(|Z_k|+1)\cdots(|Z_k|+q-1)\le(|Z_k|+q-1)^q$, by the generating function of the Stirling numbers of the first kind. For $B\ge1$, Step~2 gives $L_B\ge1+c_LM_{\widetilde L}\ge1+c_L\prod_{l=1}^L\kappa_l\ge1+q$, because $\kappa_{l+1}\ge c_l$ and $\kappa_l\ge m_l$. With $\varepsilon\le2B$, this gives $q\le2N-1$ and $|Z_k|+q-1\le2kN^{k-1}+2N\le(2k+2)N^{k-1}$. Therefore
\[
 \mathcal N(\mathcal F,\varepsilon,\|\cdot\|_\infty)\le\frac{N^{k_0}}{G}\prod_{v=1}^H(2k_v+2)N^{k_v-1}
 =\frac{1}{G}\prod_{v=1}^H(2k_v+2)\,N^{S-H}.
\]

We now derive the stated bounds. Since $2k+2\le6^{k-1}$ for $k\ge2$, the last display is at most $G^{-1}(6N)^{S-H}\le G^{-1}(12BL_B/\varepsilon)^{S-H}$, which proves~\eqref{eq:detailed-upper-bound}. Since $2k+2\le2(1+2k)$ and $H\le S-H$, it is also at most the third bound in~\eqref{eq:C-three-bounds} below. The second bound follows from the size of $\mathcal Z$ without permutations. The first bound follows in the same way from the grid $\mathcal T^S$ of $\Theta$, with $|Z_k|$ replaced by $N^k\ge q-1$, which gives $G^{-1}2^HN^S$. In summary,
\begin{equation}\label{eq:C-three-bounds}
 \mathcal N(\mathcal F,\varepsilon,\|\cdot\|_\infty)\le
 \begin{cases}
  \dfrac{1}{G}\Bigl(\dfrac{4B L_B}{\varepsilon}\Bigr)^{S},
      & \text{permutation only},\\[10pt]
  \displaystyle\prod_{v=1}^H(1+2k_v)
      \Bigl(\dfrac{4B L_B}{\varepsilon}\Bigr)^{S-H},
      & \text{scaling only},\\[10pt]
  \dfrac{1}{G}\prod_{v=1}^H(1+2k_v)
      \Bigl(\dfrac{4B L_B}{\varepsilon}\Bigr)^{S-H},
      & \text{scaling and permutation}.
 \end{cases}
\end{equation}
Finally, we express the constants of Corollary~\ref{thm:covering-number-sp-deep} through $C_+$. The first bound is at most $G^{-1}C_+^{S/(S-H)}\varepsilon^{-S}$ because $4BL_B\le C_+^{1/(S-H)}$. For the scaling-only bound, $1+2k\le6^{k-1}$ for $k\ge2$ and $k_0\ge1$ give $|\mathcal Z|\le6^{S-H-1}N^{S-H}\le e^{-1}(12BL_B/\varepsilon)^{S-H}\le e^{-1}C_+\varepsilon^{-(S-H)}$. Therefore Corollary~\ref{thm:covering-number-sp-deep} holds with
\begin{equation}\label{eq:main-prefactors}
 C_1=C_+^{S/(S-H)},\qquad C_2=e^{-1}C_+,
\end{equation}
so that $C_+=e\,C_2$ and $C_1^{1/S}=C_+^{1/(S-H)}=e^{1/(S-H)}C_2^{1/(S-H)}$.
\end{proof}

\section{The Lower Bound}
\label{app:lower-bound-proof}

This appendix proves Theorem~\ref{thm:sharp-covering-exponent}(a), and then Theorem~\ref{thm:shallow-lower-bound} and Theorem~\ref{thm:sharp-covering-exponent}(b). We use the layer indexing of Appendix~\ref{app:sec2}. The biases $b_{l,j}$ of the layers $l\notin\mathcal R$ are the biases of the $H$ scalable channels. We call them the \emph{fixed biases}, and we call the remaining $r:=S-H$ coordinates of $\theta$ the \emph{free parameters}. For a set $U$ of inputs, $U_\theta$ denotes the set of all $x\in U$ at which every neuron $v$ has a nonzero preactivation $z_v(x;\theta)$. Near such a pair $(x,\theta)$, the activation pattern is constant, so $f$ is a polynomial in $\theta$.

Part~(a) rests on a simple principle. Suppose that, at some parameter, the partial derivatives of $f$ with respect to the $r$ free parameters are linearly independent functions of $x$. Then a grid of nearby parameters yields of order $\varepsilon^{-r}$ functions that are pairwise more than $2\varepsilon$ apart. The main work is to construct such a parameter for every architecture. The construction determines the coefficients of a vanishing linear combination of these derivatives layer by layer, from its jumps across the zero sets of individual neurons. For layers that read the output of a preceding activation, we construct sufficiently many zeros along a ray in the input space. The following lemma provides them.

\begin{lemma}\label{lem:affine}
Let $\varphi_p:[0,\infty)\to\R^m$, for $p$ in a finite index set, be continuous functions that are affine on the pieces of a common finite partition of $[0,\infty)$. Let $E_p\subset\{1,\ldots,m\}$ be sets with $\bigcup_pE_p=\{1,\ldots,m\}$ such that the coordinates of $\varphi_p$ outside $E_p$ vanish identically and such that $1$ and the coordinates of $\varphi_p$ in $E_p$ are linearly independent functions. Fix $b\ne0$, and put $g_p(t;w):=w^\top\varphi_p(t)+b$ for $w\in\R^m$. We call $w$ \emph{admissible} if, for every $p$, the function $g_p(\cdot\,;w)$ is nonzero at the breakpoints of the partition, including $t=0$, has a nonzero slope on the unbounded piece, and has a zero. If an admissible $w$ exists, then there is an admissible $w$ such that the vectors $\varphi_p(t)$, over all $p$ and all zeros $t$ of $g_p(\cdot\,;w)$, span $\R^m$. Every vector sufficiently close to such a $w$ has the same properties.
\end{lemma}

\begin{proof}[Proof of Lemma~\ref{lem:affine}]
Let $w$ be admissible. Every zero of $g_p(\cdot\,;w)$ lies in the interior of a piece on which $g_p$ is affine and not constant. The signs of $g_p$ at the breakpoints and of its slope on the unbounded piece determine which pieces contain a zero. Under a small change of $w$, the zeros therefore move continuously, and no new zeros appear. This proves the last assertion.

Among all admissible vectors, choose $w$ such that the span of the vectors $\varphi_p(t)$ at the zeros $t$ of the functions $g_p(\cdot\,;w)$ has maximal dimension. Let $u\in\R^m$ be orthogonal to this span. We show that $u=0$. For every $s\in\R$, each zero $t$ of $g_p(\cdot\,;w)$ is a zero of $g_p(\cdot\,;w+su)$, because $u^\top\varphi_p(t)=0$. Moreover, $w+su$ is admissible for all but finitely many $s$, because each condition at a breakpoint or on the unbounded piece excludes at most one value of $s$. By the maximality, no admissible $w+su$ has a zero $t$ of some $g_p$ with $u^\top\varphi_p(t)\ne0$.

Fix $p$ and a piece of the partition, and suppose that $u^\top\varphi_p$ is not a constant multiple of $g_p(\cdot\,;w)$ on this piece. Then the ratio $-g_p(\cdot\,;w)/u^\top\varphi_p$ is defined and not constant on the interior of the piece, except at the zero of $u^\top\varphi_p$, so it takes infinitely many values. We can therefore choose a point $t_*$ at which the ratio equals some $s$ with $w+su$ admissible. Then $g_p(t_*;w+su)=0$ and $u^\top\varphi_p(t_*)\ne0$, a contradiction. Thus $u^\top\varphi_p=\lambda\,g_p(\cdot\,;w)$ on each piece, and the continuity at the breakpoints, where $g_p\ne0$, shows that $\lambda=\lambda_p$ is the same on all pieces. Hence $(u-\lambda_pw)^\top\varphi_p-\lambda_pb=0$ on $[0,\infty)$. Since $1$ and the coordinates of $\varphi_p$ in $E_p$ are linearly independent and $b\ne0$, we get $\lambda_p=0$ and $u_i=0$ for all $i\in E_p$. The sets $E_p$ cover $\{1,\ldots,m\}$, so $u=0$.
\end{proof}

\begin{proof}[Proof of Theorem~\ref{thm:sharp-covering-exponent}(a)]
Steps~1 and~2 make the principle above quantitative. Step~1 derives a packing of $\mathcal F$ from linearly independent derivatives, and Step~2 transfers it to the domain $\mathcal X$ and the bound $B$ by a rescaling. Step~3 reduces the linear independence to conditions on the zeros of the neurons. Step~4 constructs parameters that satisfy these conditions for networks without residual blocks, and Step~5 treats residual blocks.

Residual blocks need additional care, because the shortcut couples the input and the output of a block. We therefore write the parameters of every outer layer $l\in\mathcal R$ as $(K_l,b_l)=\eta\,(V_l,\varrho_l)$ with $\eta>0$, and we let $\eta\to0$. By positive homogeneity, the output of the residual branch is
\begin{equation}\label{eq:branch}
 \sigma_l(K_l\star h_{l-1}+b_l)=\eta\,\sigma_l(V_l\star h_{l-1}+\varrho_l),
\end{equation}
so every residual block tends to the identity. Throughout the proof, $\vartheta$ denotes the vector of the free parameters in which the entries of $(K_l,b_l)$ are replaced by those of $(V_l,\varrho_l)$ for $l\in\mathcal R$, and $\theta(\vartheta,\eta)$ denotes the corresponding parameter. The fixed biases are nonzero constants that we do not display. For a neuron $v$ of an outer layer $l$, we let $z_v$ denote the rescaled preactivation ${(V_l\star h_{l-1}+\varrho_l)}_v$, which has the sign of the preactivation and is also defined for $\eta=0$. We write $z_v(x;\vartheta)$ for the preactivations at $\eta=0$ and $U_\vartheta$ for the set of all $x\in U$ at which they are nonzero. Without residual blocks, $\vartheta$ is the vector of the free parameters, and all statements below hold with $\eta=1$.

\medskip
\noindent\textbf{Step 1 (Packing from linearly independent derivatives).}
Let $\vartheta_0$ be a value of $\vartheta$, let $U$ be a set of inputs, and let $x\in U_{\vartheta_0}$. Near $(\vartheta_0,0)$, the activation pattern at $x$ is constant, so $f(x;\theta(\vartheta,\eta))$ is a polynomial in $(\vartheta,\eta)$. By~\eqref{eq:branch}, the entries of $V_l$, of $\varrho_l$, and of the inner kernel $K_{l-1}$ of the same block enter this polynomial only through monomials that contain the factor $\eta$. The partial derivatives with respect to these entries, divided by $\eta$, are therefore continuous at $\eta=0$. We call their values at $(\vartheta_0,0)$, together with the partial derivatives with respect to the other coordinates of $\vartheta$ at $(\vartheta_0,0)$, the \emph{limiting derivatives} at $\vartheta_0$. They are $r$ functions of $x\in U_{\vartheta_0}$.

Suppose that the limiting derivatives at $\vartheta_0$ are linearly independent functions on $U_{\vartheta_0}$. Then there are inputs $x_1,\ldots,x_r\in U_{\vartheta_0}$ at which they form an invertible matrix $\widetilde J_0\in\R^{r\times r}$. Let $\Phi(\vartheta,\eta):=\bigl(f(x_a;\theta(\vartheta,\eta))\bigr)_{a=1}^r$, and let $\widetilde J(\vartheta,\eta)$ be the Jacobian of $\Phi$ with respect to $\vartheta$ in which the columns of the entries of $V_l$, $\varrho_l$, and the inner kernels are divided by $\eta$. Then $\widetilde J$ is continuous near $(\vartheta_0,0)$ and equals $\widetilde J_0$ there. Hence there are $\ell>0$, $\omega\in(0,1)$, and $\eta_0\in(0,1]$ such that the activation patterns at $x_1,\ldots,x_r$ are constant and $\|\widetilde J_0^{-1}(\widetilde J(\vartheta,\eta)-\widetilde J_0)\|_\infty\le\omega$ in the $\ell_\infty$ operator norm for all $0\le\eta\le\eta_0$ and all $\vartheta$ in the cube $Q:=\{\vartheta:\|\vartheta-\vartheta_0\|_\infty\le\ell\}$. Fix $\eta\in(0,\eta_0]$. The Jacobian of $\Phi$ with respect to $\vartheta$ is $\widetilde J$ with some columns multiplied by $\eta\le1$. The fundamental theorem of calculus therefore gives, for all $\vartheta,\vartheta'\in Q$,
\begin{align*}
 \|\widetilde J_0^{-1}\|_\infty\max_{1\le a\le r}\bigl|f(x_a;\theta(\vartheta,\eta))-f(x_a;\theta(\vartheta',\eta))\bigr|
 &\ge\bigl\|\widetilde J_0^{-1}\bigl(\Phi(\vartheta,\eta)-\Phi(\vartheta',\eta)\bigr)\bigr\|_\infty\\
 &\ge(1-\omega)\,\eta\,\|\vartheta-\vartheta'\|_\infty .
\end{align*}
For every $\varepsilon>0$, a grid in $Q$ with step $3\|\widetilde J_0^{-1}\|_\infty\varepsilon/((1-\omega)\eta)$ therefore contains at least $(\gamma_0\eta/\varepsilon)^r$ points, where $\gamma_0:=2\ell(1-\omega)/(3\|\widetilde J_0^{-1}\|_\infty)$, and the functions $f(\cdot;\theta(\vartheta,\eta))$ of two distinct grid points differ by at least $3\varepsilon$ at one of the inputs $x_1,\ldots,x_r$. Moreover, $\widetilde J(\vartheta,\eta)$ is invertible for $\vartheta\in Q$, so the $r$ free partial derivatives of $f$ at $\theta(\vartheta,\eta)$ are linearly independent.

\medskip
\noindent\textbf{Step 2 (Rescaling to the domain $\mathcal X$ and the bound $B$).}
Steps~3--5 provide a parameter $\vartheta_0$ and a bounded open set $U\subset\mathcal P:=(0,\infty)^{\Omega_0\times c_0}$ that satisfy the assumption of Step~1 and depend only on the architecture and the fixed activation slopes. We fix the resulting quantities of Step~1 and a constant $A\ge1$ that exceeds the moduli of all coordinates of $x_1,\ldots,x_r$, of all fixed biases, and of all coordinates of all $\vartheta\in Q$. For $\vartheta\in Q$ and $0<\eta\le\eta_0$, all entries of $\theta(\vartheta,\eta)$ then have modulus less than $A$, and the entries of the outer layers have modulus less than $\eta A$.

We move these parameters into $[-B,B]^S$ by the following linear map. For $0<c\le1$ and $0<\rho\le1$, let $s_l$ be $c$ raised to the number of non-residual blocks among the blocks $1,\ldots,l$. The map $\Psi_{c,\rho}$ multiplies the kernel or weight matrix of a non-residual block $l$ by $c$ and its bias by $\rho s_l$. In a residual block $l$, it multiplies the inner kernel by $c$, the inner bias by $\rho cs_l$, the second kernel by $c^{-1}$, and the outer bias by $\rho s_l$. It multiplies $a$ by $c$ and $\beta$ by $\rho cs_L$. Positive homogeneity and induction over the blocks show that the output of block $l$ in~\eqref{eq:block-network} satisfies $h_l(x;\Psi_{c,\rho}\theta)=\rho s_l\,h_l(x/\rho;\theta)$, because in a residual block the factor $c^{-1}$ of the second kernel compensates the factor $c$ of the inner kernel. Hence
\begin{equation}\label{eq:transfer}
 f(x;\Psi_{c,\rho}\theta)=\rho\,c^{L-L_{\mathrm{res}}+1}\,f(x/\rho;\theta).
\end{equation}

Let $d_0:=|\Omega_0|c_0$ and $\delta_0:=\delta/\sqrt{d_0}$. In both cases of the theorem, $\mathcal X$ contains a cube $y_0+[0,\delta_0]^{\Omega_0\times c_0}$ with $\|y_0\|_\infty\le R_{\mathcal X}$. In the first case $y_0=0$, and in the second case the cube lies in the ball. In the second case, the first block is fully connected, and replacing its bias $b_1$ by $b_1-W_1y_0$ replaces $f(x;\theta)$ by $f(x-y_0;\theta)$. We choose
\[
 \rho:=\frac{\delta_0}{A},\qquad c:=\frac{\min\{1,B\}}{2Ad_0R_{\mathcal X}},\qquad\eta:=\eta_0c^2 .
\]
For $\vartheta\in Q$, let $\theta'$ be obtained from $\Psi_{c,\rho}\theta(\vartheta,\eta)$ by the replacement of the first bias if $y_0\ne0$. Every entry of $\theta'$ has modulus less than $B$. Indeed, the weights and inner kernels are bounded by $cA$ and the second kernels by $c^{-1}\eta A=\eta_0cA$. Every bias is bounded by $c$, because its factor is at most $\rho c$ unless it is an outer bias, which is bounded by $\eta A\rho\le c$. Finally, the replacement changes the first bias by at most $cAd_0R_{\mathcal X}=\min\{1,B\}/2$. Thus $f(\cdot;\theta')\in\mathcal F$. By~\eqref{eq:transfer}, $f(y_0+\rho x;\theta')=\rho\,c^{L-L_{\mathrm{res}}+1}f(x;\theta(\vartheta,\eta))$, and the inputs $y_0+\rho x_a$ lie in $\mathcal X$. We use the grid of Step~1 with $\varepsilon/(\rho\,c^{L-L_{\mathrm{res}}+1})$ in place of $\varepsilon$. The resulting functions $f(\cdot;\theta')$ are pairwise at least $3\varepsilon$ apart in $\|\cdot\|_\infty$, so no ball of radius $\varepsilon$ contains two of them. Hence, for every $\varepsilon>0$,
\[
 \mathcal N(\mathcal F,\varepsilon,\|\cdot\|_\infty)\ge\Bigl(\frac{\gamma_0\,\eta\,\rho\,c^{L-L_{\mathrm{res}}+1}}{\varepsilon}\Bigr)^{S-H},
 \quad
 \gamma_0\,\eta\,\rho\,c^{L-L_{\mathrm{res}}+1}=\frac{\gamma_0\eta_0\,\delta}{A\sqrt{d_0}}\Bigl(\frac{\min\{1,B\}}{2Ad_0R_{\mathcal X}}\Bigr)^{L-L_{\mathrm{res}}+3}.
\]
The same bound holds for every class of functions on $\mathcal X$ that contains $\mathcal F$, because such a class contains the same separated functions. Without residual blocks, the factors $\eta$ and $\eta_0$ are absent and the exponent is $L+1$. In both situations, the exponent is at most $L+L_{\mathrm{res}}+2$. This proves~\eqref{eq:main-result} with $\gamma_-:=\gamma_0\eta_0A^{-1}d_0^{-1/2}(2Ad_0R_{\mathcal X})^{-(L+3)}$, which depends only on the architecture, the fixed activation slopes, and $R_{\mathcal X}$.

We record a consequence for the proof of part~(b). By Step~1, the $r$ free partial derivatives of $f$ at $\theta(\vartheta,\eta)$ are linearly independent, so the partial derivatives with respect to all parameters have the maximal rank $S-H$. The map $\Psi_{c,\rho}$ and the replacement of the first bias are invertible linear maps of the parameter space, which preserve this rank. Hence $\theta'\in(-B,B)^S$ is non-degenerate in the sense of Definition~\ref{def:generic-identifiability}.

\medskip
\noindent\textbf{Step 3 (Linear independence from the zeros of the neurons).}
We now give conditions on $\vartheta_0$ and $U$ under which the limiting derivatives are linearly independent. All quantities in this step are evaluated at $\vartheta=\vartheta_0$ and $\eta=0$, where every residual block is the identity, $h_l=h_{l-2}$ for $l\in\mathcal R$.

For a layer $l$ and a channel $j$, let $w_{l,j}$ and $b_{l,j}$ be the row and the bias of the channel. The row is the $j$th row of the weight matrix or the flattened filter $(K_l(u)_{jk})_{u,k}$, and for $l\in\mathcal R$ the row and the bias are taken from $V_l$ and $\varrho_l$. Let $n_l$ be the length of $w_{l,j}$, and let $\varphi_{l,p}(x)\in\R^{n_l}$ be the patch read at the position $p\in\Omega_l$. The patch is the flattened input of a fully connected layer, or $(h_{l-1}(p+u))_{u\in\Lambda_l}$ for a convolution, with zeros for $p+u\notin\Omega_{l-1}$. Thus $z_v=w_{l,j}^\top\varphi_{l,p}+b_{l,j}$ for the neuron $v=(l,p,j)$. Let $\bar h_j:=|\Omega_{\widetilde L}|^{-1}\sum_{p\in\Omega_{\widetilde L}}h_{\widetilde L,j}(p)$ be the pooled output of channel $j$. Finally, let $T_v(x)$ be the sensitivity of the output to the neuron $v$. If $v$ does not belong to an inner layer, $T_v$ is the partial derivative of $f$ with respect to the coordinate of $h_l$ at the position and channel of $v$. If $v$ belongs to an inner layer $l$, then $T_v:=\sum_uT_u\,\sigma'(z_u)V_{uv}$, where $u$ runs over the neurons of the outer layer $l+1$ and $V_{uv}$ is the entry of $V_{l+1}$ that connects $v$ to $u$. By~\eqref{eq:branch} and the chain rule, a linear combination of the limiting derivatives with coefficients $(\dot{w},\dot{b},\dot{a},\dot{\beta})$, where $\dot{b}_{l,j}=0$ for $l\notin\mathcal{R}$, equals
\begin{equation}\label{eq:limit}
 \tau(x) = \sum_{v=(l,p,j)}T_v(x)\,\sigma'(z_v(x))\bigl(\dot{w}_{l,j}^\top\varphi_{l,p}(x)+\dot{b}_{l,j}\bigr)
   +\sum_{j=1}^{c_{\widetilde L}}\dot{a}_j\bar{h}_{j}(x)+\dot{\beta}
\end{equation}
on $U_{\vartheta_0}$. We call $x\in U$ a \emph{simple zero} of a neuron $v$ if $z_v(x)=0$, $\nabla_xz_v(x)\ne0$, and $z_{v'}(x)\ne0$ for all neurons $v'\ne v$. We claim that $\tau\equiv0$ on $U_{\vartheta_0}$ implies $(\dot{w},\dot{b},\dot{a},\dot{\beta})=0$ under the following two conditions.
\begin{itemize}
 \item \emph{Condition~(i).} Every channel $(l,j)$ has simple zeros $x_1,\ldots,x_{n_l}\in U$ of neurons $v_a=(l,p_a,j)$ of this channel such that $T_{v_a}(x_a)\ne0$ and the patches $\varphi_{l,p_a}(x_a)$ span $\R^{n_l}$.
 \item \emph{Condition~(ii).} The functions $1,\bar{h}_{1},\ldots,\bar{h}_{c_{\widetilde L}}$ are linearly independent on $U_{\vartheta_0}$.
\end{itemize}
To prove the claim, we show by induction on $l$ that $\dot w_{l,j}=0$ and $\dot b_{l,j}=0$. Suppose that this holds for all layers before $l$, and let $x_a$ be a simple zero of $v_a$ as in Condition~(i). The terms of the earlier layers in~\eqref{eq:limit} vanish. No other neuron changes sign near $x_a$, so all factors of the remaining terms are continuous near $x_a$ except $\sigma'(z_{v_a})$, which jumps by $1-\alpha_l$ across $\{z_{v_a}=0\}$. Hence $\tau\equiv0$ gives
\begin{equation}\label{eq:jump}
 T_{v_a}(x_a)\bigl(\dot w_{l,j}^\top\varphi_{l,p_a}(x_a)+\dot b_{l,j}\bigr)=0 .
\end{equation}
If $l\notin\mathcal R$, then $\dot b_{l,j}=0$, and \eqref{eq:jump} gives $\dot w_{l,j}=0$ because the patches span $\R^{n_l}$. If $l\in\mathcal R$, then \eqref{eq:jump} and $z_{v_a}(x_a)=0$ show that $(\dot w_{l,j},\dot b_{l,j})$ and $(w_{l,j},b_{l,j})$ both solve the equations $y^\top\varphi_{l,p_a}(x_a)+\chi=0$, $a=1,\ldots,n_l$. The patches span $\R^{n_l}$, so the solutions $(y,\chi)$ form a line, and $(\dot w_{l,j},\dot b_{l,j})=\mu\,(w_{l,j},b_{l,j})$ for some $\mu\in\R$. Near $x_1$, the term of $v_1$ in \eqref{eq:limit} is then $\mu\,T_{v_1}\sigma(z_{v_1})$. Every other term is affine near $x_1$, because no neuron reads the outer layer at $\eta=0$. The gradient of $\tau$ therefore jumps by $(1-\alpha_l)\,\mu\,T_{v_1}(x_1)\nabla_x z_{v_1}(x_1)$ across $\{z_{v_1}=0\}$, so $\mu=0$. This completes the induction. Condition~(ii) then gives $\dot a=0$ and $\dot\beta=0$.

\medskip
\noindent\textbf{Step 4 (Construction without residual blocks).}
Suppose that no block is residual, so that layers and blocks coincide. We construct a parameter $\vartheta_0$, with nonzero fixed biases, and a bounded open set $U\subset\mathcal P$ that satisfy Conditions~(i) and~(ii). We work along a ray $\{te:t\ge0\}$ with $e\in\mathcal P$ and regard the preactivations $z_v$, the outputs $h_v=\sigma_l(z_v)$, and the patches $\varphi_{l,p}$ as functions of $t$, which are continuous and piecewise affine. The construction has three ingredients. Lemma~\ref{lem:affine} chooses the row of a channel such that the zeros of its neurons on the ray have spanning patches. Two channels of every block receive opposite rows, so that one of them is active at every input and position. This makes the sensitivities $T_v$ nonzero, also for the ReLU, and lets every neuron of the next block change sign. Generic choices keep the zeros of distinct neurons apart.

We choose the blocks $l=1,\ldots,L$ in this order such that the following properties hold.
\begin{itemize}
 \item[(F1)] Every $z_v$ is nonzero at $t=0$, has a nonzero slope on its unbounded piece, and has at least one zero. Its zeros have nonzero slope and differ from the zeros of all other neurons.
 \item[(F2)] $w_{l,2}=-w_{l,1}$ and $b_{l,2}=\nu_l-b_{l,1}$ with $\nu_l>0$. Thus $z_{l,p,1}+z_{l,p,2}=\nu_l$, and $h_{l,p,1}(t)>0$ or $h_{l,p,2}(t)>0$ for every $t$ and $p$.
 \item[(F3)] For positions $p\ne q$ of block $l$, the vectors $(h_{l,p,j}(t))_j$ and $(h_{l,q,j}(t))_j$ differ for all but finitely many $t$.
 \item[(F4)] For $l\ge2$ and every channel $j$, the patches $\varphi_{l,p}(t)$ at the zeros $t$ of $z_{l,p,j}$, over all $p\in\Omega_l$, span $\mathbb R^{n_l}$.
\end{itemize}
We note two consequences. First, by (F1), the slope of $h_v$ jumps at each zero of $z_v$, while the outputs of all other neurons of the same and of earlier blocks are affine nearby. Hence $1$ and the outputs $h_v$ of the neurons of block $l$ are linearly independent functions on $[0,\infty)$, and so are $1$ and the non-padding coordinates of each patch $\varphi_{l+1,p}$. Second, for positions $p\ne q$ of block $l$, the difference $\varphi_{l,p}-\varphi_{l,q}$ vanishes at only finitely many $t$. For $l=1$, this holds if the patches $\varphi_{1,p}(e)$ are pairwise distinct. For $l\ge2$, the two positions share an offset that lies in the grid for both, by the padding convention of Section~\ref{sec:settings}, so this follows from (F3) for block $l-1$.

\emph{Property (F3).} Given (F2), property (F3) holds for a generic row $w_{l,1}$. Indeed, suppose that $(h_{l,p,j})_j=(h_{l,q,j})_j$ on an interval. By (F2), there are $i\in\{1,2\}$ and a subinterval on which $z_{l,p,i}>0$. There $h_{l,q,i}=h_{l,p,i}>0$, which forces $z_{l,q,i}=z_{l,p,i}$, that is, $w_{l,1}^\top(\varphi_{l,p}-\varphi_{l,q})=0$. Hence $w_{l,1}$ is orthogonal to the values of $\varphi_{l,p}-\varphi_{l,q}$ on a piece on which this difference does not vanish identically. A generic row avoids these finitely many proper subspaces.

\emph{Block~$1$.} Choose $e$ with pairwise distinct coordinates. Then the patches $\varphi_{1,p}(e)$, $p\in\Omega_1$, are pairwise distinct, because any two positions share an offset that lies in the grid for both. Choose a generic row $w_{1,1}$ with negative entries and $b_{1,1}>0$. Then $z_{1,p,1}(t)=b_{1,1}+t\,w_{1,1}^\top\varphi_{1,p}(e)$ has exactly one zero, and the zeros of distinct positions differ. Define channel~$2$ by (F2) with a generic $\nu_1\in(0,b_{1,1})$, and choose every channel $j\ge3$ like channel~$1$. Then (F1)--(F3) hold.

\emph{Block~$l\ge2$.} Suppose that (F1)--(F4) hold for the blocks before $l$. For channel~$1$, we apply Lemma~\ref{lem:affine} to the patches $\varphi_{l,p}$, $p\in\Omega_l$, with the partition by the zeros of all neurons chosen so far and with $E_p$ the set of non-padding coordinates. Some position has a full patch, so the sets $E_p$ cover all coordinates, and the linear independence was noted above. An admissible row exists. Indeed, let $w$ have the entries $-1$ at the channels $1$ and $2$ of block $l-1$ and small negative entries elsewhere. On the unbounded piece, every earlier neuron has the sign of its slope. By (F2), the two coordinates of the channels $1$ and $2$ at a position therefore contribute $-(1-\alpha_{l-1})|s|<0$ to the slope of $w^\top\varphi_{l,p}$, where $s\ne0$ is the slope of the neuron of channel $1$. Hence every $w^\top\varphi_{l,p}+b$ has a negative slope on the unbounded piece and, for a large $b>0$, is positive at $t=0$, so it has a zero. A small perturbation of $w$ makes these functions nonzero at the breakpoints. With $b_{l,1}:=b$, the lemma provides an open set of rows $w_{l,1}$ that satisfy (F4) and (F1), except possibly for the distinctness of the zeros of different positions. A generic row in this set also satisfies (F3) and has distinct zeros. To see the latter, note that a common zero $t$ of $z_{l,p,1}$ and $z_{l,q,1}$, $p\ne q$, imposes on the row the equations $w^\top\varphi_{l,p}(t)=-b$ and $w^\top(\varphi_{l,p}(t)-\varphi_{l,q}(t))=0$. Since $b\ne0$, they are independent or inconsistent whenever $\varphi_{l,p}(t)\ne\varphi_{l,q}(t)$, which holds for all but finitely many $t$. As $t$ varies, the rows that satisfy both equations therefore form a closed null set. Next, define channel~$2$ by (F2) with a small $\nu_l>0$ outside a finite set. Then $z_{l,p,2}=\nu_l-z_{l,p,1}$ has exactly one zero near each zero of $z_{l,p,1}$, with a nearby patch, and no other zeros, so (F1), (F3), and (F4) remain valid. Every channel $j\ge3$ is chosen like channel~$1$, with the zeros of the channels $i<j$ added to the partition.

\emph{Verification of Conditions~(i) and~(ii).} Choose $0<t_0<t_1$ such that $(t_0,t_1)$ contains all zeros, and let $U\subset\mathcal P$ be a bounded open neighborhood of the segment $\{te:t_0\le t\le t_1\}$. At a zero of a neuron of channel $k$ of block $L$, only $\bar h_{k}$ among the pooled outputs has a jump of its slope. Hence $1,\bar h_1,\ldots,\bar h_{c_L}$ are linearly independent on the segment. A linear relation among them on $U_{\vartheta_0}$ holds at all points $te$ at which no neuron vanishes, and by continuity on the whole segment. This proves Condition~(ii). For $l\ge2$, every zero $t$ of $z_{l,p,j}$ gives a simple zero $x=te$ of the neuron $(l,p,j)$, because the derivative of $z_{l,p,j}$ along $e$ is nonzero and all other neurons are nonzero at $x$ by (F1). The patches at these zeros span $\R^{n_l}$ by (F4). For $l=1$, take a position $p$ with a full patch and the zero $t$ of $z_{1,p,j}$. All other neurons are nonzero on a ball in $U$ around $te$, so every point $x$ of this ball with $z_{1,p,j}(x)=0$ is a simple zero. The patches $\varphi_{1,p}(x)$ of these points fill an open subset of the hyperplane $\{y:w_{1,j}^\top y=-b_{1,j}\}$ of $\mathbb R^{n_1}$. This hyperplane does not contain $0$, so $n_1$ of these patches are linearly independent.

It remains to obtain $T_{v_a}(x_a)\ne0$ at the finitely many simple zeros $x_a$ selected above. All properties used so far are stable under small perturbations of the weights and of $a$, and so is the activation pattern at each $x_a$. Near the constructed parameter, $T_{v_a}(x_a)$ is therefore a polynomial in $a$ and the weights of the blocks after $v_a$. It is the sum, over the paths from $v_a$ to the output, of the products of the weights along the path, multiplied by $|\Omega_L|^{-1}$ and by the slopes of the neurons on the path. By (F2), one of the channels $1$ and $2$ is active at $x_a$ at every position of every block, and every position lies in the patch of a position of the next block. Hence some path from $v_a$ to the output passes only through active neurons. Its monomial has a positive coefficient, so the polynomial does not vanish identically. A generic small perturbation of the weights and of $a$ therefore makes all $T_{v_a}(x_a)$ nonzero and preserves all other properties.

\medskip
\noindent\textbf{Step 5 (Construction with residual blocks).}
In general, we choose $\vartheta_0$ by the procedure of Step~4, applied to the layers in their order. At $\eta=0$, every residual block is the identity. Consider a residual block with inner layer $l-1$ and outer layer $l$. The inner layer reads $h_{l-2}$, which is the output of the last fully connected or convolutional layer before it, or the input $x$. The outer layer, with the kernel $V_l$ and the bias $\varrho_l$, reads the inner layer, and no layer reads the outer layer. Whenever a layer reads $x$ at $\eta=0$, use the first-layer construction of Step~4, excluding the finitely many bias values that would repeat an earlier zero on the ray. For a layer that reads the output of an activation, the later-layer construction of Step~4 gives (F4) with all previous zeros included in the partition. For each channel $(l,j)$ reading $x$, choose a position with a full patch. Near its simple zero on the ray, all other preactivations remain nonzero, and the patches at its zeros fill a relatively open subset of $\{y:w_{l,j}^{\top}y=-b_{l,j}\}$. Since $b_{l,j}\ne0$, these patches span $\R^{n_l}$.

For Condition~(ii), apply Step~4 to the last non-residual block if one exists. Otherwise the pooled features are the input channel averages, which vary independently on an open subset of $U_{\vartheta_0}$. For the sensitivities, recall that $T_v$ is the sensitivity of $f$ to a coordinate of $h_l=h_{l-2}$ if $v$ belongs to an outer layer $l$, and $T_v=\sum_uT_u\sigma'(z_u)V_{uv}$ if $v$ belongs to an inner layer. In both cases, $T_v$ is again a polynomial with nonnegative coefficients in $a$, in the entries of the kernels $V_l$, and in the weights of the later layers. Its monomials correspond to the paths from $v$ to the output, which may now pass through shortcuts with weight $1$. The paired preactivations in (F2) sum to $\nu_l>0$ throughout $U$. Hence a path through active neurons exists, so a generic small perturbation makes all $T_{v_a}(x_a)$ nonzero. Each selected simple zero persists on a transverse line under this perturbation, so the selected patches retain their rank.
\end{proof}

\begin{proof}[Proof of Theorem~\ref{thm:shallow-lower-bound} and Theorem~\ref{thm:sharp-covering-exponent}(b)]
We count, for almost every $g\in\mathcal F$, the parameters in the set $\Theta_0$ of Step~1 of Appendix~\ref{app:main} that realize $g$, and we convert this count into a volume by an area formula. Steps~1--3 derive~\eqref{eq:exact-G} from generic identifiability, provided that the volume $V$ defined below is positive. Step~4 proves $V>0$. Step~5 proves that networks with one hidden layer are generically identifiable, which yields Theorem~\ref{thm:shallow-lower-bound}.

Write $\Phi(\theta):=f(\cdot;\theta)\in C(\mathcal X)$, and recall the nodes $u=1,\ldots,H$ and the group $\Gamma$ of Appendix~\ref{app:main}. A face $Q$ of $\Theta_0$ of dimension $r=S-H$ is determined by one incoming parameter $e_u$ of every node $u$, which we call the \emph{fixed parameter} of $u$, and a sign. It consists of the $\theta\in[-B,B]^S$ with $\theta_{e_u}=\pm B$ for all $u$, and we identify it with a cube in $\mathbb R^{r}$. The faces of lower dimension and the intersections of faces are null sets, which we ignore. For $\theta$ in the interior of a face $Q$ and a tangent vector $\zeta$ of $Q$, let $\|\partial_{\zeta}f(\cdot;\theta)\|_\infty$ be the supremum of $|\partial_{\zeta}f(x;\theta)|$ over $x\in\mathcal X_\theta$. We define
\begin{equation}\label{eq:area-volume}
 J_\theta:=\frac{\omega_r}{\vol_r\bigl\{\zeta:\|\partial_{\zeta}f(\cdot;\theta)\|_\infty\le1\bigr\}},\qquad
 V:=\int_{\Theta_0}J_\theta\,d\theta,\qquad C_-:=\frac{V}{\omega_r},
\end{equation}
where $\omega_{r}$ is the volume of the unit ball of $\mathbb R^{r}$, $\vol_{r}$ is the $r$-dimensional Hausdorff measure, and $J_\theta:=0$ if the set in the denominator is unbounded. ``Almost every'' refers to the Lebesgue measure on $\mathbb R^S$ or on the faces, and to $\vol_{r}$ on $\mathcal F$.

\medskip
\noindent\textbf{Step 1 (Area formula).} By Step~2 of Appendix~\ref{app:main}, $\Phi$ is Lipschitz on each face $Q$. Hence $\Phi$ has a metric differential $\mathrm{md}\,\Phi_\theta$, which is a seminorm on $\mathbb R^{r}$, at almost every $\theta\in Q$. The area formula of \citet{kirchheim1994rectifiable} states that, for every measurable $A\subset Q$,
\begin{equation}\label{eq:area-formula}
 \int_AJ(\mathrm{md}\,\Phi_\theta)\,d\theta=\int_{C(\mathcal X)}\#\bigl(A\cap\Phi^{-1}(g)\bigr)\,d\vol_{r}(g),
 \qquad J(s):=\frac{\omega_{r}}{\vol_{r}\{\zeta:s(\zeta)\le1\}} .
\end{equation}
Moreover, $J(\mathrm{md}\,\Phi_\theta)=J_\theta$ for almost every $\theta$. Indeed, for a countable dense set $\mathcal D\subset\mathcal X$, the map $g\mapsto(g(x))_{x\in\mathcal D}$ embeds $C(\mathcal X)$ isometrically into $\ell^\infty(\mathcal D)$, the dual of the separable space $\ell^1(\mathcal D)$. For Lipschitz maps into such a dual, the metric differential is almost everywhere the norm of the weak$^*$ differential \citep[Section~3]{ambrosio2000rectifiable}. For almost every $\theta$, no point of $\mathcal D$ lies on the zero set of a neuron. The weak$^*$ differential along $\zeta$ is then $(\partial_{\zeta}f(x;\theta))_{x\in\mathcal D}$, and its norm is $\|\partial_{\zeta}f(\cdot;\theta)\|_\infty$ because $\partial_{\zeta}f(\cdot;\theta)$ is continuous on $\mathcal X_\theta$. Summing \eqref{eq:area-formula} over the faces yields
\begin{equation}\label{eq:area-count}
 V=\int_{\mathcal F}n(g)\,d\vol_{r}(g),\qquad n(g):=\#\bigl(\Theta_0\cap\Phi^{-1}(g)\bigr).
\end{equation}
Applied to $A=\{J_\theta=0\}$ and to null sets $A$, \eqref{eq:area-formula} shows that for almost every $g\in\mathcal F$, no preimage of $g$ lies in such a set.

\medskip
\noindent\textbf{Step 2 (Almost every function has exactly $G$ preimages).} We call $\theta\in\Theta_0$ \emph{typical} if it lies in the interior of a face $Q$ of dimension $r$, $J_\theta>0$, all biases of the scalable channels are nonzero, $|\theta_e|<B$ for every parameter other than the fixed parameters of $Q$, and $P\theta\ne\theta$ for every $P\in\Gamma$ other than the identity. The last condition fails only if two nodes of the same scalable layer have incoming parameters that agree up to a permutation of coordinates, which is a proper linear condition on each face. By Step~1, for almost every $g\in\mathcal F$, all preimages of $g$ in $\Theta_0$ are therefore typical.

Let $\theta$ be typical. We write $D(\lambda)$, $\lambda\in(0,\infty)^H$, for the scaling~\eqref{eq:scale} with the factors $\lambda_u$ of the nodes. For a node $u$, let $g_u\in\mathbb R^S$ be the derivative of $D(\lambda)\theta$ with respect to $\log\lambda_u$ at $\lambda=\mathbf 1$. Its entry is $\theta_e$ at every parameter $\theta_e$ that feeds $u$, $-\theta_e$ at every parameter $\theta_e$ that reads from $u$, and $0$ elsewhere. By Proposition~\ref{prop:scaling-permutation}, the derivative of $f$ along $g_u$ vanishes.

\emph{(2a) Transversality.} Let $M$ be the $H\times H$ matrix with the entries $M_{u'u}:=(g_u)_{e_{u'}}$. Suppose that $Mc=0$ for some $c\ne0$. Then $\sum_uc_ug_u$ is tangent to $Q$. This vector is nonzero, because its entry at the bias of a node $u$ with $c_u\ne0$ is $c_u$ times this bias. The derivative of $f$ along it vanishes, which contradicts $J_\theta>0$. Hence $M$ is invertible, and the vectors $g_u$ span a complement of the tangent space of $Q$. The partial derivatives of $f(\cdot;\theta)$ therefore span the same space as its derivatives along $Q$, which has dimension $r$ because $J_\theta>0$. Thus $\theta$ is non-degenerate. Moreover, ${(Mc)}_{u'}=\theta_{e_{u'}}(c_{u'}-c_u)$ if the fixed parameter $e_{u'}$ reads from a node $u\ge1$, and ${(Mc)}_{u'}=\theta_{e_{u'}}c_{u'}$ otherwise. If the fixed parameters formed a directed cycle of nodes, the indicator vector $c$ of the nodes that can be reached from this cycle along fixed parameters would satisfy $Mc=0$. Hence the fixed parameters of $Q$ form no cycle. Conversely, if the fixed parameters form no cycle, then $Mc=0$ forces $c_{u'}=0$ node by node along the fixed parameters, so $M$ is invertible. Step~4 uses this converse.

\emph{(2b) Unique normalization.} We show that $D(\lambda)\theta\in\Theta_0$ implies $\lambda=\mathbf 1$. Put $\lambda_0:=1$ for node $0$ and $\ell_e:=\log(B/|\theta_e|)\ge0$ for $\theta_e\ne0$, so that $\ell_e=0$ exactly for the fixed parameters. The condition $D(\lambda)\theta\in\Theta_0$ states that, for every node $u'\ge1$, the largest rescaled incoming parameter has modulus $B$. Equivalently, $\log\lambda_{u'}$ is the minimum of $\log\lambda_u+\ell_e$ over the nonzero parameters $\theta_e$ from a node $u$ to $u'$. Suppose that $\min_{u'}\log\lambda_{u'}<0$ is attained at $u'$. Then the minimizing parameter has $\ell_e=0$ and reads from a node $u\ne0$ that attains the same minimum. Iterating this argument produces a cycle of fixed parameters, which (2a) excludes. Suppose that $\max_{u'}\log\lambda_{u'}>0$ is attained at $u'$. Then $\log\lambda_{u'}\le\log\lambda_u+\ell_{e_{u'}}=\log\lambda_u$ for the node $u$ from which $e_{u'}$ reads, so $u\ne0$ attains the maximum, and iterating again produces a cycle. Hence $\lambda=\mathbf 1$.

\emph{(2c) Counting.} The group $\Gamma$ maps $\Theta_0$ onto itself and preserves $\Phi$, and $\Gamma\theta$ consists of $G$ points. Hence $n(g)\ge G$ for almost every $g\in\mathcal F$. Now assume generic identifiability, and let $E\subset\mathbb R^S$ be the null set of the non-degenerate parameters at which it fails. Scalings and permutations preserve non-degeneracy and the fibers of $\Phi$, so they preserve $E$. By (2a), the map $(\vartheta,\lambda)\mapsto D(\lambda)\vartheta$ is a local diffeomorphism from $Q\times(0,\infty)^H$ into $\mathbb R^S$ near $(\theta,\mathbf 1)$. Since $D(\lambda)\vartheta\in E$ if and only if $\vartheta\in E$, Fubini's theorem shows that $E\cap Q$ is a null set near $\theta$. By Step~1, for almost every $g\in\mathcal F$, all preimages of $g$ in $\Theta_0$ are therefore typical and lie outside $E$. Let $\theta$ and $\theta'$ be two such preimages. By (2a), both are non-degenerate, so $\theta'=PD(\lambda)\theta$. Then $D(\lambda)\theta=P^{-1}\theta'\in\Theta_0$, and (2b) gives $\theta'=P\theta$. Thus $n(g)=G$ for almost every $g\in\mathcal F$, and \eqref{eq:area-count} yields $\vol_{r}(\mathcal F)=V/G$.

\medskip
\noindent\textbf{Step 3 (From volume to covering numbers).} Let $\vol_{r,\delta}$ denote the Hausdorff premeasure at scale $\delta$. Thus $\vol_{r,\delta}(A)\le\sum_i\omega_r(\operatorname{diam}A_i/2)^r$ for every cover of $A$ by sets $A_i$ of diameter at most $\delta$, and $\vol_{r,\delta}(A)\uparrow\vol_r(A)$ as $\delta\downarrow0$. A cover of $\mathcal F$ by $\mathcal N(\mathcal F,\varepsilon,\|\cdot\|_\infty)$ closed balls of radius $\varepsilon$ gives $\vol_{r,2\varepsilon}(\mathcal F)\le\omega_r\varepsilon^r\,\mathcal N(\mathcal F,\varepsilon,\|\cdot\|_\infty)$. Letting $\varepsilon\to0$ and using Step~2, we obtain $\liminf_{\varepsilon\to0}\varepsilon^r\mathcal N(\mathcal F,\varepsilon,\|\cdot\|_\infty)\ge\vol_r(\mathcal F)/\omega_r=C_-/G$. This is~\eqref{eq:exact-G} and, for $L=1$, \eqref{eq:shallow-lower-bound}, provided that $V>0$.

\medskip
\noindent\textbf{Step 4 ($V>0$).} Let $\theta_*\in(-B,B)^S$ be non-degenerate. Under the assumptions of Theorem~\ref{thm:sharp-covering-exponent}, Step~2 of the proof of part~(a) provides such a parameter, and for one hidden layer Step~5(i) below provides it. Non-degeneracy is an open condition, because a basis of the span of the partial derivatives is detected at finitely many inputs. After a small perturbation, all parameters of $\theta_*$ are therefore nonzero, and its normalization $\theta=D(\lambda)\theta_*\in\Theta_0$ from Step~1 of Appendix~\ref{app:main} lies in the interior of a face $Q$ of dimension $r$. Every cycle of parameters of $\theta_*$ has a positive length $\sum_e\ell_e$, because all parameters have modulus less than $B$, and this length is invariant under scalings. Hence the fixed parameters of $\theta$, which have $\ell_e=0$, form no cycle, and the matrix $M$ of (2a) is invertible. Since $\theta$ is non-degenerate, the kernel of the derivative of $f$ at $\theta$ has dimension $H$. It is spanned by the vectors $g_u$, which are linearly independent and span a complement of the tangent space of $Q$. Thus $\|\partial_{\zeta}f(\cdot;\theta)\|_\infty>0$ for every nonzero tangent vector $\zeta$ of $Q$, so $J_\theta>0$. The same holds near $\theta$ in $Q$, and hence $V>0$.

\medskip
\noindent\textbf{Step 5 (One hidden layer is generically identifiable).} Let $L=1$, write $t_j(x):=w_j^\top x+b_j$ and $H_j:=\{t_j=0\}$, and let $\mathcal X^\circ$ be the connected interior of $\mathcal X$.

\emph{(i) Non-degeneracy.} We show that $\theta$ is non-degenerate if and only if $a_j\ne0$ for all $j$ and the hyperplanes $H_j$ are pairwise distinct and meet $\mathcal X^\circ$. Such parameters exist in $(-B,B)^S$. First, let $\theta$ satisfy these conditions. Since $\partial_{a_j}f=\sigma(t_j)=\sigma'(t_j)\,t_j$ and $a_j\ne0$, the partial derivatives span the same space as the $d_1(d_0+1)+1=S-H$ functions $\sigma'(t_j)x_i$, $\sigma'(t_j)$, and $1$. Suppose that $\sum_j\sigma'(t_j)\ell_j+c=0$ on $\mathcal X^\circ$ for affine functions $\ell_j$ and a constant $c$. The jump across $H_j\cap\mathcal X^\circ$ shows that $\ell_j$ vanishes on $H_j$, so $\ell_j=\mu_jt_j$ and $\sum_j\mu_j\sigma(t_j)+c=0$. The jumps $(1-\alpha)\mu_jw_j$ of the gradient then give $\mu_j=0$ and $c=0$. Hence $\theta$ is non-degenerate. Conversely, if $a_j=0$, the derivatives with respect to $w_j$ and $b_j$ vanish. If $H_j$ does not meet $\mathcal X^\circ$, then $\sigma'(t_j)$ is constant on $\mathcal X^\circ$, so the derivatives of neuron $j$ and of $\beta$ lie in the span of $x_1,\dots,x_{d_0}$ and $1$. If $H_j=H_{j'}$ for some $j'\ne j$, then, by $\sigma'(-t)=1+\alpha-\sigma'(t)$, the derivatives of the neurons $j$ and $j'$ and of $\beta$ lie in the span of $\sigma'(t_j)x_i$, $\sigma'(t_j)$, $x_i$, and $1$. In each case, the partial derivatives span a space of dimension less than $S-H$.

\emph{(ii) Identifiability.} Let $\theta$ be non-degenerate with $\sum_{j\in I}a_jw_j\ne0$ for every nonempty $I\subset\{1,\dots,d_1\}$, which excludes a null set. Let $\theta'\in\mathbb R^S$ satisfy $f(\cdot;\theta')=f(\cdot;\theta)$ on $\mathcal X$. At the points of $H_j\cap\mathcal X^\circ$ outside the other hyperplanes, the gradient of $f(\cdot;\theta)$ jumps by $(1-\alpha)a_jw_j\ne0$. These points form a nonempty relatively open subset of $H_j$. The function $f(\cdot;\theta')$ is differentiable outside the hyperplanes $H'_{j'}$ of its neurons with $w'_{j'}\ne0$, so $H_j=H'_{j'}$ for some $j'$. The $d_1$ hyperplanes $H_j$ are distinct. After a permutation of the neurons of $\theta'$, we therefore have $H'_j=H_j$, that is, $(w'_j,b'_j)=c_j(w_j,b_j)$ with $c_j\ne0$, and equal jumps of the gradient give $a'_j|c_j|=a_j$. Using $\sigma(-t)=\sigma(t)-(1+\alpha)t$, we obtain
\[
 f(x;\theta')-f(x;\theta)=-(1+\alpha)\sum_{j:\,c_j<0}a_j\,t_j(x)+\beta'-\beta\qquad(x\in\mathcal X).
\]
The linear part vanishes on the open set $\mathcal X^\circ$, so $\sum_{j:\,c_j<0}a_jw_j=0$. Hence $c_j>0$ for all $j$ and $\beta'=\beta$. Thus $\theta'$ is obtained from $\theta$ by a permutation and the positive scalings $c_j$. This proves generic identifiability for $L=1$, and Theorem~\ref{thm:shallow-lower-bound} follows from Steps~1--4.
\end{proof}

\section{Continuous Piecewise-Linear Activations}\label{app:cpwl}

In this appendix, we prove the results of Section~\ref{sec:cpwl-extension}.

\begin{proof}[Proof of Theorem~\ref{thm:covering-number-cpwl}]
The proof follows Steps~2 and~3 of Appendix~\ref{app:main} without positive scaling. Permuting the $d_l$ hidden neurons of a layer, together with their incoming and outgoing parameters, preserves the network function for every activation. These permutations form a group $\Gamma$ of order $G=\prod_{l=1}^Ld_l!$.

We first bound the Lipschitz constant of $\theta\mapsto f(\cdot;\theta)$ on $\Theta=[-B,B]^S$. Since $|\rho_l(t)|\le|\rho_l(0)|+J_{\rho_l}|t|$, induction on $l$ gives $\|h^{(l)}(\theta)\|_\infty\le M_l$ and $\|h^{(l)}(\theta)-h^{(l)}(\eta)\|_\infty\le E_l\|\theta-\eta\|_\infty$ for all $\theta,\eta\in\Theta$, where $M_0=R_{\mathcal{X}}$, $E_0=0$, and
\[
 M_l := |\rho_l(0)| + J_{\rho_l} B(d_{l-1}M_{l-1}+1),\qquad
 E_l := J_{\rho_l}(d_{l-1}M_{l-1}+1) + J_{\rho_l} B d_{l-1}E_{l-1}.
\]
At the readout, this yields
\[
 \|f(\cdot;\theta)-f(\cdot;\eta)\|_\infty
 \le (1 + d_L M_L + B d_L E_L)\|\theta-\eta\|_\infty
 =: L_{\boldsymbol\rho}\|\theta-\eta\|_\infty.
\]
For $B\ge 1$, induction on $l$ gives $M_l\le(R_{\mathcal{X}}+\rho_0)(2B)^l\prod_{j=1}^l(\bar{J}_{\rho_j}d_{j-1})$ and
\[
 L_{\boldsymbol\rho}\le 4(L+1)(R_{\mathcal{X}}+\rho_0)(2B)^{L+1}\Bigl(\prod_{l=1}^L\bar{J}_{\rho_l}\Bigr)\Bigl(\prod_{l=0}^L d_l\Bigr).
\]

We now count as in the permutation-only case of Step~3 of Appendix~\ref{app:main}, with $L_B$ replaced by $\overline L_{\boldsymbol\rho}:=\max\{L_{\boldsymbol\rho},1+\max_ld_l\}$, so that $d_l\le2N-1$ for $N:=\lceil B\overline L_{\boldsymbol\rho}/\varepsilon\rceil$. The number of orbits of $\Gamma$ in the grid $\mathcal T^S$ is at most $G^{-1}2^HN^S$, and hence
\[
 \mathcal{N}(\mathcal{F}_{\boldsymbol\rho},\varepsilon,\|\cdot\|_\infty)
 \le \frac{(2N)^S}{G}
 \le \frac{1}{G}\Bigl(\frac{4B\overline L_{\boldsymbol\rho}}{\varepsilon}\Bigr)^S,
 \qquad C_{\boldsymbol\rho} := 4B\overline L_{\boldsymbol\rho}.
\]
This bound holds for every $B>0$ and $0<\varepsilon\le2B$. For $B\ge1$, the bound on $L_{\boldsymbol\rho}$ above also exceeds $1+\max_ld_l$, which gives the constant in~\eqref{eq:cpwl-covering}.
\end{proof}

\begin{proof}[Proof of Corollary~\ref{cor:cpwl-lower-bound}]
We embed the class of ReLU networks with widths $\tilde{d}_l:=\lfloor d_l/2\rfloor$ and parameters in $[-1,1]$ into $\mathcal F_{\boldsymbol\rho}$ and apply Theorem~\ref{thm:sharp-covering-exponent}(a) to this class. The embedding represents each ReLU unit by two $\rho_l$-units, as in \citet[Proposition~1(b)]{yarotsky2017}, with parameter bounds that are uniform over the class.

For each layer $l$, choose a breakpoint $a_l$ of $\rho_l$ and $\eta_l>0$ such that $\rho_l$ is affine on $[a_l-\eta_l,a_l]$ and on $[a_l,a_l+\eta_l]$, with slopes $s_{l,-}\ne s_{l,+}$. Put
\[
 a_{l,+}:=a_l,\qquad a_{l,-}:=a_l-\frac{\eta_l}{2},\qquad
 \Delta_l:=s_{l,+}-s_{l,-}\ne 0,\qquad \chi_l:=\rho_l(a_{l,+})-\rho_l(a_{l,-}).
\]
In a ReLU network of the class above, the preactivations of layer $l$ are bounded by $U_l$, where $U_0:=R_{\mathcal{X}}$ and $U_l:=\tilde{d}_{l-1}U_{l-1}+1$. Let $\lambda:=\min\{1,\min_l\eta_l/(2\max_lU_l)\}$. For every $|t|\le\max_lU_l$, both $a_{l,+}+\lambda t$ and $a_{l,-}+\lambda t$ lie in $[a_l-\eta_l,a_l+\eta_l]$, and the second point lies to the left of $a_l$. Hence
\[
 \max\{t,0\} = \frac{\rho_l(a_{l,+}+\lambda t) - \rho_l(a_{l,-}+\lambda t) - \chi_l}{\Delta_l\lambda}.
\]
We replace each ReLU unit $i\le\tilde{d}_l$ of layer $l$, with preactivation $z_i^{(l)}$, by the two units $\rho_l(a_{l,\pm}+\lambda z_i^{(l)})$, and we give the remaining units of $\mathcal F_{\boldsymbol\rho}$ zero outgoing weights. By the identity above, the resulting network in $\mathcal F_{\boldsymbol\rho}$ realizes the same function. Its first-layer weights are $\lambda W_{i,j}^{(1)}$, its first-layer biases are $a_{1,\pm}+\lambda b_i^{(1)}$, and for $2\le l\le L$ the unit $(i,\pm)$ has the weights and the bias
\[
 W_{(i,\pm),(j,+)}^{\prime(l)}=\frac{W_{i,j}^{(l)}}{\Delta_{l-1}}
 =-W_{(i,\pm),(j,-)}^{\prime(l)},\qquad
 b_{i,\pm}^{\prime(l)}=a_{l,\pm}+\lambda b_i^{(l)}
   -\frac{\chi_{l-1}}{\Delta_{l-1}}\sum_{j=1}^{\tilde d_{l-1}} W_{i,j}^{(l)}.
\]
The readout weights are $\pm a_i/(\Delta_L\lambda)$, and the readout bias is $\beta-\chi_L\sum_ia_i/(\Delta_L\lambda)$. All these parameters have modulus at most
\begin{equation}\label{eq:cpwl-embedding-bound}
 B_{\boldsymbol\rho}:=1+\max_{1\le l\le L}\max\{|a_{l,+}|,|a_{l,-}|\}+\max_{1\le l\le L}\frac{1+\tilde d_{\max}|\chi_l|}{|\Delta_l|\lambda},
 \qquad \tilde d_{\max}:=\max_{1\le l\le L}\tilde d_l .
\end{equation}
Thus, for $B\ge B_{\boldsymbol\rho}$, the class $\mathcal F_{\boldsymbol\rho}$ contains the ReLU class above. By the remark at the end of Step~2 of the proof of Theorem~\ref{thm:sharp-covering-exponent}(a), the bound~\eqref{eq:main-result} for the ReLU class, with $B=1$ and $\alpha_l=0$ for every $l$, also holds for $\mathcal F_{\boldsymbol\rho}$. This gives $\mathcal{N}(\mathcal{F}_{\boldsymbol\rho},\varepsilon,\|\cdot\|_\infty)\ge C_{\boldsymbol\rho,-}\,\varepsilon^{-\tilde{r}}$ with $\tilde{r}:=d_0\tilde{d}_1+\sum_{l=2}^L\tilde{d}_{l-1}\tilde{d}_l+\tilde{d}_L+1$ and $C_{\boldsymbol\rho,-}:=(\gamma_-\delta)^{\tilde r}$, where $\gamma_-$ is the constant of the ReLU architecture and $\delta\le1$ is the radius of a ball in $\mathcal X$. Finally, $\lfloor d/2\rfloor\ge2d/5$ for $d\ge 4$, with $\lfloor d/2\rfloor=d/2$ for even $d$. Hence $\tilde{r}\ge(S-H)/7$ in general and $\tilde{r}\ge(S-H)/4$ for even widths.
\end{proof}

\begin{proof}[Proof of Corollary~\ref{cor:cpwl-local-range}]
Let $a_l$ be the midpoint of $I_l=[a_l-\eta_{0,l},a_l+\eta_{0,l}]$, where $0<\eta_{0,l}\le 1$. On $I_l$, we have $\rho_l(t)=\rho_l(a_l)+\psi_l(t-a_l)$, where $\psi_l$ is positively homogeneous with the slopes $s_{l,-}\ne s_{l,+}$ of $\rho_l$ on the two sides of $a_l$.

\medskip
\noindent\textbf{Step 1 (Covering bound \eqref{eq:cpwl-local-cover}).}
We rewrite every network in $\mathcal{F}_{\boldsymbol\rho,\mathbf{I}}(B)$ as a network with positively homogeneous activations and apply Appendix~\ref{app:main}. Let $J_{0,l}:=\max\{|s_{l,-}|,|s_{l,+}|\}$ and $\sigma_l:=\psi_l/J_{0,l}$, which is positively homogeneous and $1$-Lipschitz. For $f(\cdot;\theta)\in\mathcal{F}_{\boldsymbol\rho,\mathbf{I}}(B)$, the output of layer $l$ is $\rho_l(a_l)+J_{0,l}\sigma_l(z_l-a_l)$ on $\mathcal X$. Hence $f(\cdot;\theta)$ is realized on $\mathcal X$ by a network with the activations $\sigma_l$, the weights $J_{0,l-1}W_l$, and the biases $b_l-a_l+\rho_{l-1}(a_{l-1})W_l\mathbf 1$, with a corresponding change of the readout. With $J_0:=\max_lJ_{0,l}$, $a_{\max}:=\max_l|a_l|$, $\rho_{\max}:=\max_l|\rho_l(a_l)|$, and $d_{\max}:=\max_{0\le l\le L}d_l$, these parameters have modulus at most
\[
 \overline{B} := \max\bigl\{1,\,\max\{1,J_0\}\bigl[B(1+\rho_{\max}d_{\max})+a_{\max}\bigr]\bigr\} \ge B.
\]
The proof in Appendix~\ref{app:main} uses only that the activations are positively homogeneous and $1$-Lipschitz, so the third bound in~\eqref{eq:C-three-bounds} holds for the class of these networks with parameters in $[-\overline B,\overline B]^S$. Every node of layer $l$ has $k_v=d_{l-1}+1$ incoming parameters. We apply this bound with radius $\varepsilon/2\le2\overline B$ and replace the center of each ball that meets $\mathcal{F}_{\boldsymbol\rho,\mathbf{I}}(B)$ by an element of $\mathcal{F}_{\boldsymbol\rho,\mathbf{I}}(B)$ in the ball. This proves \eqref{eq:cpwl-local-cover} with
\[
 C_{\boldsymbol\rho,\mathbf{I}} := \prod_{l=1}^L(2d_{l-1}+3)^{d_l}\bigl(8\overline{B}L_{\overline{B}}\bigr)^{S-H}.
\]

\medskip
\noindent\textbf{Step 2 (Approximation with a parameter bound independent of $\delta$).}
Let $\rho_l=\rho$ and $I_l=I=[a-\eta_0,a+\eta_0]$ for all layers. We convert a ReLU approximant into a network in $\mathcal{F}_{\rho,I}(B_*)$ and keep all parameters bounded independently of $\delta$. Put $a_+:=a$, $a_-:=a-\eta_0/2$, $\lambda_0:=\eta_0/2$, $\Delta:=s_+-s_-$, and $\chi:=\rho(a_+)-\rho(a_-)$. Choose a nonzero slope $s\in\{s_-,s_+\}$, and let $t_0=a\pm\eta_0/2$ be the midpoint of the corresponding piece of $\rho$ in $I$. Set $M:=R_H+1$ and $\tau_0:=\eta_0/(4M)$, and define
\begin{align*}
 B_* := \max\biggl\{&1,\,\lambda_0,\,\frac{1}{|\Delta|},\,
 \max\{|a_+|,|a_-|\}+\lambda_0+\Bigl|\frac{\chi}{\Delta}\Bigr|,\,
 \frac{\tau_0}{|\Delta|\lambda_0},\\
 &|t_0|+\tau_0\Bigl(1+\Bigl|\frac{\chi}{\Delta\lambda_0}\Bigr|\Bigr),\,
 \frac{1}{2|s|},\,\frac{1}{2|s|\tau_0}\biggr\}.
\end{align*}
By Theorem~1 of \citet{kong2025posterior}, for every $f_0\in\mathcal{H}_{d_0}^s(R_H)$ and $0<\delta\le 1$, there is a ReLU network $f_{\mathrm{ReLU}}$ with depth $L_0=O(\log(2/\delta))$, widths $d_l=O(\delta^{-d_0/(2s)})$, and a parameter bound $B_0\ge 1$ such that $\|f_{\mathrm{ReLU}}\|_\infty\le M$ and $\|f_{\mathrm{ReLU}}-f_0\|_\infty\le\delta$. Its preactivations in layer $l$ are bounded by $U_l$, where $U_0:=R_{\mathcal{X}}$ and $U_l:=B_0(d_{l-1}U_{l-1}+1)$. The conversion has two parts. First, we divide the preactivations of layer $l$ by $U_l$ and replace each normalized ReLU unit, with preactivation $\tilde z_i^{(l)}\in[-1,1]$, by the two units $\rho(a_\pm+\lambda_0\tilde{z}_i^{(l)})$, as in the proof of Corollary~\ref{cor:cpwl-lower-bound}. This produces $f_{\mathrm{ReLU}}/U_{L_0+1}$ with parameters in $[-B_*,B_*]$ and preactivations in $I$. Second, we multiply the output by $U_{L_0+1}$. Let $Q:=\lceil\max\{2,d_0,\ldots,d_{L_0},B_0,U_0\}\rceil$, and write $U_{L_0+1}=\prod_{j=1}^mq_j$ with $m=\lceil\log_Q U_{L_0+1}\rceil=O(L_0)$ factors $q_j\in[1,Q]$. We append $m$ layers of width $2Q$. Each of them uses the pairs $\rho(t_0\pm\tau_0 u)$, which are affine in $u$, with the weights $\pm q_j/(2sQ)$ and multiplies its input by $q_j$. The readout weights are $\pm 1/(2s\tau_0)$. The resulting network equals $f_{\mathrm{ReLU}}$ and belongs to $\mathcal{F}_{\rho,I}(B_*)$, with depth $O(\log(2/\delta))$ and width $O(\delta^{-d_0/(2s)})$.
\end{proof}

\section{Applications and Implications}\label{app:sec4}

\begin{proof}[Proof of Proposition~\ref{prop:risk-bounds}]
Steps~1 and~2 derive the upper bounds of parts~(a) and~(b) from the covering-number bound. Step~3 proves the lower bounds of parts~(a) and~(b) by a two-point argument, and Step~4 proves part~(c) by the local asymptotic minimax theorem.

Let $\mathcal{G}=\{\ell(f(\cdot),\cdot):f\in\mathcal{F}\}$ denote the loss class. Since $\ell$ is $L_\ell$-Lipschitz in its first argument on the range of $\mathcal F\times\mathcal Y$, Theorem~\ref{thm:covering-upper-bound} gives, for every $0<\varepsilon\le 2B$,
\begin{equation}\label{eq:sp-cover}
  \mathcal{N}(\mathcal{G},L_\ell\varepsilon,\|\cdot\|_\infty)
  \le \mathcal{N}(\mathcal{F},\varepsilon,\|\cdot\|_\infty)
  \le \frac{C_+}{G}\,\varepsilon^{-(S-H)}.
\end{equation}

\medskip
\noindent\textbf{Step 1 (Upper bound in part~(a)).}
For every empirical risk minimizer $\widehat{f}$, the standard risk decomposition gives
\[
  \mathbb{E}\bigl[R(\widehat{f})\bigr]-\min_{f\in\mathcal{F}}R(f)
  \le 2\,\mathbb{E}\sup_{f\in\mathcal{F}}\bigl|R_n(f)-R(f)\bigr|.
\]
We bound the right-hand side by the entropy integral of $\mathcal G$. Let $\varepsilon_*=\min\{M_\ell/L_\ell,2B\}$. By \eqref{eq:sp-cover}, for $0<u\le L_\ell\varepsilon_*$,
\[
  \log\mathcal{N}(\mathcal{G},u,\|\cdot\|_\infty)
  \le \Bigl((S-H)\log\frac{1}{\varepsilon_*}+\log C_+-\log G\Bigr)
      + (S-H)\log\frac{L_\ell\varepsilon_*}{u}.
\]
For $L_\ell\varepsilon_*<u\le M_\ell$, the monotonicity of covering numbers shows that the first term alone is an upper bound. Using $\sqrt{a+b}\le\sqrt{a}+\sqrt{b}$ and $\int_0^{L_\ell\varepsilon_*}\sqrt{\log(L_\ell\varepsilon_*/u)}\,du=\frac{\sqrt{\pi}}{2}L_\ell\varepsilon_*$, we obtain
\[
  \int_0^{M_\ell}\sqrt{\log\mathcal{N}(\mathcal{G},u,\|\cdot\|_\infty)}\,du
  \le M_\ell\sqrt{(S-H)\log\frac{1}{\varepsilon_*}+\log C_+-\log G}
      + \frac{\sqrt{\pi}}{2}L_\ell\varepsilon_*\sqrt{S-H}.
\]
Symmetrization and Dudley's entropy integral \citep{wainwright2019high} then yield the upper bound in part~(a) with $C_\ell = 48\max\bigl\{M_\ell,\frac{\sqrt{\pi}}{2}L_\ell\varepsilon_*\bigr\}$.

\medskip
\noindent\textbf{Step 2 (Upper bound in part~(b)).}
Under the Bernstein condition, a union bound over a finite cover replaces the entropy integral. Let $f^\star\in\arg\min_{f\in\mathcal{F}}R(f)$, and let $\{f_1,\ldots,f_m\}\subset\mathcal{F}$ be a minimal $\varepsilon$-cover of $\mathcal{F}$ with $\varepsilon=1/(L_\ell n)\le 2B$. We apply Bernstein's inequality and Young's inequality to $\ell(f_j(X),Y)-\ell(f^\star(X),Y)$ for $j=1,\ldots,m$ \citep{bartlett2005local}, and we choose $f_j$ with $\|\widehat{f}-f_j\|_\infty\le\varepsilon$. For every $u>0$, with probability at least $1-e^{-u}$,
\[
  R(\widehat{f})-\min_{f\in\mathcal{F}}R(f)
  \le \frac{3}{n} + \frac{2(v+2M_\ell/3)}{n}\bigl((S-H)\log(L_\ell n)+\log C_+-\log G+u\bigr).
\]
Integrating this tail bound over $u>0$ gives the upper bound in part~(b).

\medskip
\noindent\textbf{Step 3 (Lower bounds in parts~(a) and~(b)).}
Both lower bounds compare two distributions $P_+$ and $P_-$ of $(X,Y)$ whose risk minimizers are different constant functions in $\mathcal F$. Let $s:=\min\{1,B\}$, so that the constant functions with values in $[0,s]$ belong to $\mathcal F$. Under $P_\pm$, let $X\sim P_X$, and let $Y$ be independent of $X$. We write $R_\pm$ for the risk under $P_\pm$ and $e_\pm(a):=\mathbb E_\pm\ell(a,Y)-\min_{a'\in\R}\mathbb E_\pm\ell(a',Y)$ for the excess loss of a prediction $a\in\R$. In both constructions below, the minimum over $a'$ is attained at a value in $[0,s]$. Hence $R_\pm(f)-\min_{g\in\mathcal F}R_\pm(g)=\mathbb E\,e_\pm(f(X))$ for every function $f$. If $e_+(a)+e_-(a)\ge\Delta$ for all $a\in\R$, then every estimator $\widetilde f$ based on $n$ observations satisfies
\begin{equation}\label{eq:two-point}
 \max_{\pm}\Bigl\{\mathbb E_\pm\bigl[R_\pm(\widetilde f)\bigr]-\min_{f\in\mathcal F}R_\pm(f)\Bigr\}
 \ge\frac{\Delta}{2}\bigl(1-\mathrm{TV}(P_+^n,P_-^n)\bigr)
 \ge\frac{\Delta}{2}\Bigl(1-\sqrt{\tfrac n2\,\mathrm{KL}(P_+\|P_-)}\Bigr),
\end{equation}
where the last step is Pinsker's inequality.

For part~(a), consider the absolute loss $\ell(a,y)=|a-y|$ and $Y\in\{0,s\}$ with $P_\pm(Y=s)=\frac12\pm\gamma$, where $0<\gamma\le\frac14$. The expected loss is minimized at $a'=s$ under $P_+$ and at $a'=0$ under $P_-$, and a direct computation gives $e_+(a)+e_-(a)\ge2\gamma s$ for all $a\in\R$. Moreover, $\mathrm{KL}(P_+\|P_-)=2\gamma\log\frac{1+2\gamma}{1-2\gamma}\le16\gamma^2$. With $\gamma=1/(6\sqrt n)$, \eqref{eq:two-point} gives the lower bound $s/(12\sqrt n)$.

For part~(b), consider the squared loss $\ell(a,y)=(a-y)^2$ and $Y\in\{0,1\}$ with $P_\pm(Y=1)=p_\pm:=s(\frac12\pm\gamma)$, where $0<\gamma\le\frac14$. The expected loss is minimized at $a'=p_\pm$, and $e_\pm(a)=(a-p_\pm)^2$, so $e_+(a)+e_-(a)\ge(p_+-p_-)^2/2=2\gamma^2s^2$. The Bernstein condition holds, because the loss difference $\ell(f(X),Y)-\ell(p_\pm,Y)$ is the product of $f(X)-p_\pm$ and $f(X)+p_\pm-2Y$, and the second factor is bounded on the range of $\mathcal F$. Moreover, $\mathrm{KL}(P_+\|P_-)\le(p_+-p_-)^2/(p_-(1-p_-))\le32\gamma^2$. With $\gamma=1/(8\sqrt n)$, \eqref{eq:two-point} gives the lower bound $s^2/(128n)$.

Both lower bounds of the proposition therefore hold with $c_-:=\min\{s/12,\,s^2/128\}$.

\medskip
\noindent\textbf{Step 4 (Lower bound in part~(c)).}
We apply the local asymptotic minimax theorem to a family in $\mathcal F$ of dimension $r:=S-H$. Steps~1 and~2 of the proof of Theorem~\ref{thm:sharp-covering-exponent}(a) provide a cube $Q\subset\R^r$, an affine map $\vartheta\mapsto\theta'(\vartheta)$ from $Q$ into $(-B,B)^S$, and inputs $x_1',\ldots,x_r'$ in the interior of $\mathcal X$ with the following property. For every $\vartheta\in Q$, all neurons of $\theta'(\vartheta)$ have nonzero preactivations at these inputs, and the matrix of the partial derivatives of $f(x_a';\theta'(\vartheta))$ with respect to $\vartheta$ is invertible. Write $f_\vartheta:=f(\cdot;\theta'(\vartheta))$. For almost every $\vartheta\in Q$, the zero set of every neuron of $\theta'(\vartheta)$ is a Lebesgue null set. Indeed, for fixed parameters of the earlier layers, a preactivation $w^\top\varphi+b$ vanishes identically on one of the finitely many linear regions only for $(w,b)$ in a proper affine subspace, because the fixed biases are nonzero. We fix such a $\vartheta_0$ in the interior of $Q$.

The partial derivatives $g_i:=\partial f_\vartheta/\partial\vartheta_i$ at $\vartheta_0$ then exist $P_X$-almost everywhere and are continuous near the inputs $x_a'$. Since $P_X$ has a positive density near these inputs, $g_1,\ldots,g_r$ are linearly independent in $L^2(P_X)$. Let $e_1,\ldots,e_r$ be an orthonormal basis of their span, and put $\psi(\vartheta):=(\langle f_\vartheta,e_k\rangle)_{k=1}^r$, where $\langle\cdot,\cdot\rangle$ is the inner product of $L^2(P_X)$. By Step~2 of Appendix~\ref{app:main}, $f_\vartheta(x)$ is Lipschitz in $\vartheta$ uniformly in $x$, and it is differentiable at $\vartheta_0$ for $P_X$-almost every $x$. Hence the model of the observations $(X,Y)$ with $Y=f_\vartheta(X)+\xi$ is differentiable in quadratic mean at $\vartheta_0$, with the Fisher information matrix $\varsigma^{-2}\Gamma$, where $\Gamma_{ij}:=\langle g_i,g_j\rangle$ is nonsingular. Moreover, $\psi$ is differentiable at $\vartheta_0$ with the derivative $A_{ki}:=\langle g_i,e_k\rangle$, and $A^\top A=\Gamma$.

For an estimator $\widetilde f_n$ based on $n$ observations, let $T_n:=(\langle\widetilde f_n,e_k\rangle)_{k=1}^r$, so that $\|\widetilde f_n-f_\vartheta\|_{L^2(P_X)}^2\ge|T_n-\psi(\vartheta)|^2$. The local asymptotic minimax theorem \citep[Theorem~8.11]{vanderVaart1998asymptotic}, applied with the loss $\min\{|u|^2,M\}$ and $M\to\infty$, gives
\[
 \liminf_{n\to\infty}\ n\sup_{\vartheta\in Q}\mathbb E_\vartheta\bigl|T_n-\psi(\vartheta)\bigr|^2\ \ge\ \mathbb E|Z|^2,
 \qquad Z\sim\mathcal N\bigl(0,\,A(\varsigma^{-2}\Gamma)^{-1}A^\top\bigr)=\mathcal N(0,\varsigma^2I_r).
\]
Since $\mathbb E|Z|^2=\varsigma^2r$ and $f_\vartheta\in\mathcal F$ for $\vartheta\in Q$, this proves part~(c).

\end{proof}

\begin{proof}[Proof of Corollary~\ref{cor:oracle-regression}]
We apply the oracle inequality of \citet[Lemma~4]{schmidthieber2020nonparametric}. It concerns the regression model with standard normal noise and a class $\mathcal G$ of functions such that $\mathcal G$ and $f_0$ are bounded by some $F\ge1$. For a least-squares estimator $\widehat g$ over $\mathcal G$ and every $\delta\in(0,1]$ such that at least three $\|\cdot\|_\infty$-balls of radius $\delta$ are needed to cover $\mathcal G$, it gives
\[
 \mathbb E\|\widehat g-f_0\|_{L^2(P_X)}^2
 \le4\Bigl[\inf_{g\in\mathcal G}\|g-f_0\|_{L^2(P_X)}^2+F^2\,\frac{18\log\mathcal N_\delta+72}{n}+32\,\delta F\Bigr],
\]
where $\mathcal N_\delta$ is the number of balls of radius $\delta$, with arbitrary centers, that cover $\mathcal G$. The lemma is stated for inputs in the unit cube, and an affine change of the input variable reduces the bounded domain $\mathcal X$ to this case.

We apply the lemma to the rescaled model $Y/\varsigma=f_0(X)/\varsigma+\xi/\varsigma$ with $\mathcal G=\{f/\varsigma:f\in\mathcal F_M\}$, $F=\max\{1,M/\varsigma\}$, and $\delta=1/n$. A cover of $\mathcal G$ by balls of radius $1/n$ corresponds to a cover of $\mathcal F_M$ by balls of radius $\varsigma/n$. Since $n\ge3\varsigma$, we have $\varsigma/n\le1/3\le2B$, and every cover of $\mathcal F$ also covers $\mathcal F_M$. Theorem~\ref{thm:covering-upper-bound} therefore gives $\log\mathcal N_\delta\le(S-H)\log(n/\varsigma)+\log C_+-\log G$. Moreover, $\mathcal F_M$ contains the constant functions with values in $[-1,1]$, and a ball of radius $\varsigma/n$ contains such constants only from an interval of length $2\varsigma/n$. Hence at least $n/\varsigma\ge3$ balls are needed. Multiplying the inequality of the lemma by $\varsigma^2$ and using $\varsigma F=\max\{\varsigma,M\}$ gives
\[
 \mathbb E\|\widehat f-f_0\|_{L^2(P_X)}^2
 \le4\inf_{f\in\mathcal F_M}\|f-f_0\|_{L^2(P_X)}^2
 +\frac{\max\{\varsigma,M\}^2}{n}\bigl(72\log\mathcal N_\delta+288+128\bigr),
\]
where the last term uses $128\,\varsigma\max\{\varsigma,M\}\le128\max\{\varsigma,M\}^2$. Since $288+128\le72\cdot6$, this proves the corollary.

\medskip
\noindent\textbf{Rate over H\"older balls.}
Let $f_0\in\mathcal{H}_{d_0}^s(R_H)$ and $M=R_H+1$. For $0<\delta\le 1$, Theorem~1 of \citet{kong2025posterior} provides an architecture with depth $L=O(\log(2/\delta))$, width $d=O(\delta^{-d_0/(2s)})$, and a fixed bound $B=B_*\ge 1$ such that $\mathcal F$ contains a function $f_\delta$ with $\|f_\delta-f_0\|_\infty\le\delta$. Then $\|f_\delta\|_\infty\le M$, so $f_\delta\in\mathcal F_M$. For this architecture, $S-H=O(Ld^2)=O(\delta^{-d_0/s}\log(e/\delta))$ and $\log C_+=O((S-H)\log^2(e/\delta))$. The choice $\delta=(\log^3(en)/n)^{s/(2s+d_0)}$ in Corollary~\ref{cor:oracle-regression} yields $\mathbb{E}\|\widehat{f}-f_0\|_{L^2(P_X)}^2=O\bigl([(\log^3 n)/n]^{2s/(2s+d_0)}\bigr)$.
\end{proof}

\end{document}